\documentclass{article}

\usepackage{microtype}
\usepackage{graphicx}
\usepackage{subfigure}
\usepackage{booktabs} 

\usepackage{hyperref}

\newcommand{\ours}{Video-LaVIT}
\newcommand{\web}{\url{https://video-lavit.github.io}}

\usepackage[accepted]{icml2024}

\usepackage{amsmath}
\usepackage{amssymb}
\usepackage{mathtools}
\usepackage{amsthm}
\usepackage{enumitem}
\usepackage{bm}

\usepackage[capitalize,noabbrev]{cleveref}

\theoremstyle{plain}

\theoremstyle{definition}

\theoremstyle{remark}

\usepackage[textsize=tiny]{todonotes}
\usepackage{multirow}

\icmltitlerunning{Video-LaVIT: Unified Video-Language Pre-training with Decoupled Visual-Motional Tokenization}

\begin{document}

\twocolumn[
\icmltitle{Video-LaVIT: Unified Video-Language Pre-training with Decoupled Visual-Motional Tokenization}



\icmlsetsymbol{equal}{*}

\begin{icmlauthorlist}
\icmlauthor{Yang Jin}{pku}
\icmlauthor{Zhicheng Sun}{pku}
\icmlauthor{Kun Xu}{kuaishou}
\icmlauthor{Kun Xu}{kuaishou}
\icmlauthor{Liwei Chen}{kuaishou}
\icmlauthor{Hao Jiang}{pku}
\icmlauthor{Quzhe Huang}{pku}
\icmlauthor{Chengru Song}{kuaishou}
\icmlauthor{Yuliang Liu}{kuaishou}
\icmlauthor{Di Zhang}{kuaishou}
\icmlauthor{Yang Song}{kuaishou}
\icmlauthor{Kun Gai}{kuaishou}
\icmlauthor{Yadong Mu}{pku}
\end{icmlauthorlist}

\icmlaffiliation{pku}{Peking University, China}
\icmlaffiliation{kuaishou}{Kuaishou Technology, China}
\icmlcorrespondingauthor{Yadong Mu}{myd@pku.edu.cn}

\icmlkeywords{Machine Learning, ICML}

\vskip 0.3in
]



\printAffiliationsAndNotice{}  

\begin{abstract}
In light of recent advances in multimodal Large Language Models (LLMs), there is increasing attention to scaling them from image-text data to more informative real-world videos. Compared to static images, video poses unique challenges for effective large-scale pre-training due to the modeling of its spatiotemporal dynamics. In this paper, we address such limitations in video-language pre-training with an efficient video decomposition that represents each video as keyframes and temporal motions. These are then adapted to an LLM using well-designed tokenizers that discretize visual and temporal information as a few tokens, thus enabling unified generative pre-training of videos, images, and text. At inference, the generated tokens from the LLM are carefully recovered to the original continuous pixel space to create various video content. Our proposed framework is both capable of comprehending and generating image and video content, as demonstrated by its competitive performance across 13 multimodal benchmarks in image and video understanding and generation. Our code and models are available at \web.
\end{abstract}

\section{Introduction}
\label{sec:intro}
Recently, the significant breakthrough of Large Language Models (LLMs)~\cite{brown2020language,touvron2023llama} has brought a surge in building general-purpose multimodal AI assistants~\cite{openai2023gptv,gemini2023gemini} that can follow both textual and visual instructions. Drawing on the remarkable reasoning abilities of LLMs and knowledge in massive alignment corpus (e.g., image-text pairs), they showcase the great potential of accurately comprehending and generating visual content~\cite{sun2024emu, jin2024unified, dong2023dreamllm}. Despite their success, these multimodal LLMs~\cite{alayrac2022flamingo,liu2023llava} predominantly concentrate on the image-text data, leaving the adaptation for video modality less explored. In contrast to static images, video serves as a dynamic media form that is more in line with human visual perception. Learning effectively from video is particularly essential for enhancing machine intelligence to comprehend the real world.

\begin{figure}[t]
\begin{center}
\centerline{\includegraphics[width=\linewidth]{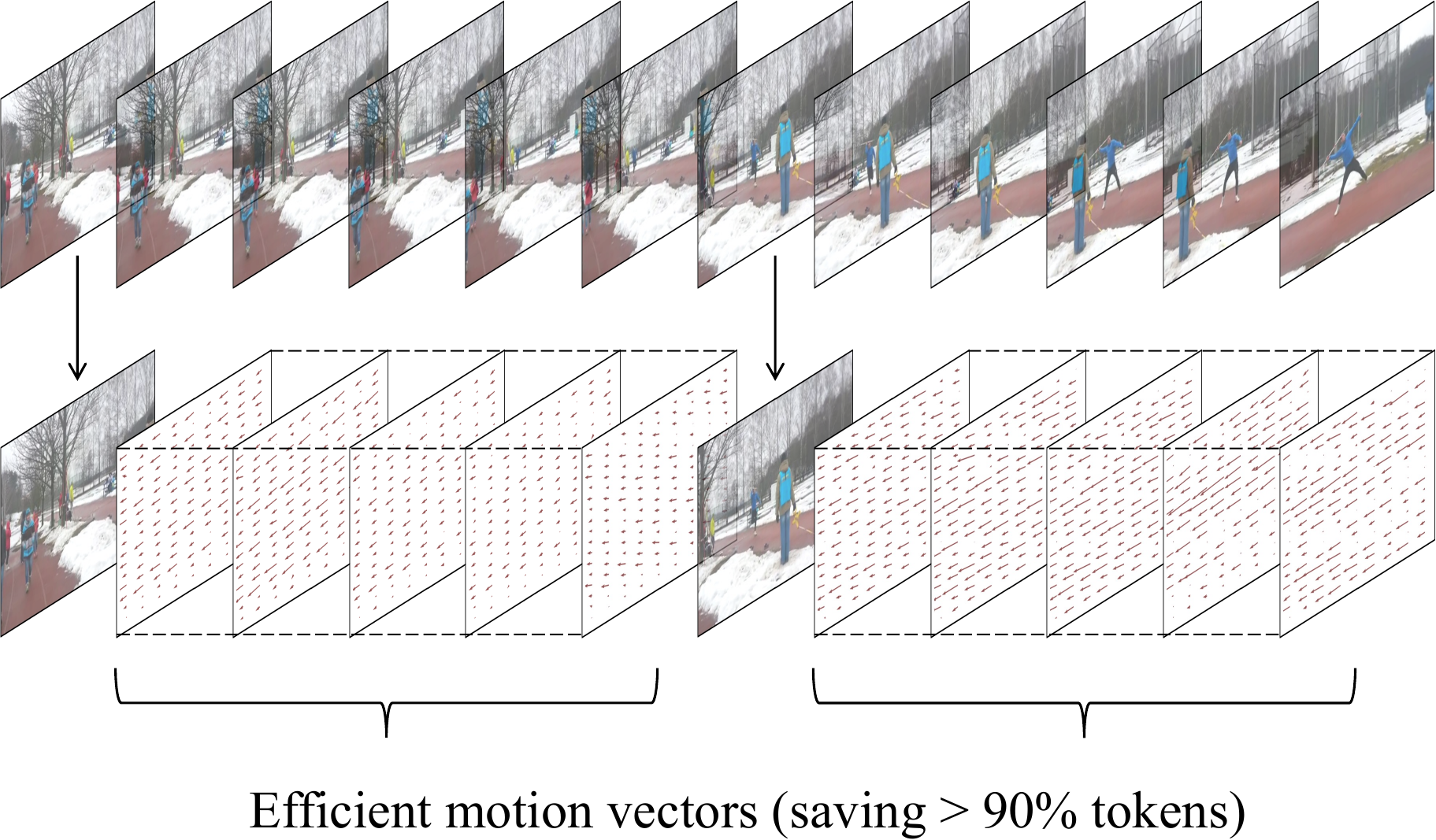}}
\caption{The key observation in this work is: most video parts have a high degree of temporal redundancy that may be described by motion vectors. By exploiting these motion vectors, the video can be efficiently tokenized for pre-training of multimodal LLMs.}
\label{fig:intro}
\end{center}
\vskip -0.3in
\end{figure}

To this end, several approaches have made attempts at harnessing the generative capabilities of LLMs for handling video data. Inheriting the successful paradigm from the image domain, they represent video as a sequence of visual tokens that aligns with LLMs' semantic space by utilizing a pre-trained 2D image model~\cite{li2023videochat, zhang2023video} or a 3D video backbone~\cite{kondratyuk2023videopoet}. Nevertheless, the existing designs are still not competent for effectively encoding videos. Compared to images, videos pose unique challenges associated with higher demands for learning complex spatiotemporal clues, such as time-varying actions and scene changes. In this regard, encoding individual video frames separately by the 2D visual encoder falls short of capturing the temporal motion information, which plays a vital role in identifying distinct behaviors and events within the video content. Although the recent concurrent work VideoPoet~\cite{kondratyuk2023videopoet} crafts a 3D video tokenizer for video generation with LLM, its applicability is constrained to short video clips due to the use of long token sequences (e.g., 1280 tokens for a 2.2s clip).\linebreak When it comes to understanding or generating long videos, inputting excessive numbers of tokens into LLMs is deemed unacceptable in terms of computational resources.

This work addresses the limitation in video-language pre-training by exploring an efficient video representation that decomposes video into keyframes and temporal motions.
Our motivation is built upon the natural characteristics of video data itself. As illustrated in~\cref{fig:intro}, a video is typically divided into several shots, where video frames within each shot often exhibit substantial information redundancy. It is superfluous to encode all of these frames as tokens and incorporate them into the generative pre-training of LLMs. This fact strongly spurs us to decompose each video into alternating keyframes and motion vectors, where the former encapsulate the primary visual semantics and the latter depict the dynamic evolution of its corresponding keyframe over time. There are several benefits to such decomposed representation: (1) Compared to processing consecutive video frames utilizing 3D encoders, the combination of a single keyframe and motion vectors requires fewer tokens to represent video temporal dynamics, which is more efficient for large-scale pre-training. (2) The model can inherit the acquired visual knowledge from an off-the-shelf image-only LLM and focus solely on modeling temporal information without learning from scratch.

Based on the above motivations, we present \textbf{\ours} (\textbf{La}nguage-\textbf{VI}sion \textbf{T}ransformer), a new multimodal pre-training approach that effectively empowers LLMs to comprehend and generate video content in a unified framework. Specifically, \ours~incorporates two core components: a \textit{tokenizer} and a \textit{detokenizer} to handle video modality. The video tokenizer aims to transform the continuous video data into a sequence of compact discrete tokens akin to a foreign language, where the keyframes are processed by utilizing an established image tokenizer~\cite{jin2024unified}. For converting the temporal motions into the compatible discrete format, a spatiotemporal motion encoder is devised. It can capture the time-varying contextual information contained in extracted motion vectors, thereby significantly enhancing LLMs' ability to comprehend the intricate actions in video. The video detokenizer is responsible for mapping the discretized video token generated by LLMs back into its original continuous pixel space. During training, video is represented as an alternating discrete visual-motion token sequence, and thus can be optimized under the same next-token prediction objective together with different modalities. Since video is inherently a time series, this joint autoregressive pre-training contributes to learning the sequential relationships of different video clips. We found that \ours, is capable of serving as a multimodal generalist to achieve promising results in both understanding and generation tasks without further fine-tuning. The key contributions of this work are summarized as: 

\begin{itemize}[leftmargin=*]
\item We introduce \ours, a multimodal pre-training method that pushes the limit of LLMs' unified understanding and generation capability towards video.
\item To efficiently model visual and temporal information in video, \ours~incorporates a novel video tokenizer and detokenizer that operates on the decomposed representations of keyframes and motion vectors.
\item Experiments on 13 multimodal benchmarks demonstrate that \ours~achieves very competitive performance, ranging from image and video comprehension to zero-shot text-to-image and text-to-video generation.
\end{itemize}






\section{Related Work}

\textbf{Vision-language pre-training}. Following the success of using large-scale image-text pairs for contrastive learning of vision-language models~\cite{radford2021learning}, a similar idea has been exploited in generative pre-training, where visual and language data are jointly modeled under an autoregressive process. In practice, this is typically achieved by adapting visual image inputs to pre-trained LLMs~\cite{raffel2020exploring, brown2020language, touvron2023llama} via an intermediate module like cross-attention~\cite{alayrac2022flamingo}, Q-Former~\cite{li2023blip2}, or linear projection~\cite{liu2023llava}. More recent approaches such as CM3Leon~\cite{yu2023scaling} and LaVIT~\cite{jin2024unified} advocate the use of discrete visual tokenizers~\cite{van2017neural, esser2021taming} to form a unified next token prediction objective. However, these methods are primarily focused on image-text data and cannot be directly extended to videos due to the significantly higher computational cost.

\begin{figure*}[t]
\begin{center}
\includegraphics[width=0.96\linewidth]{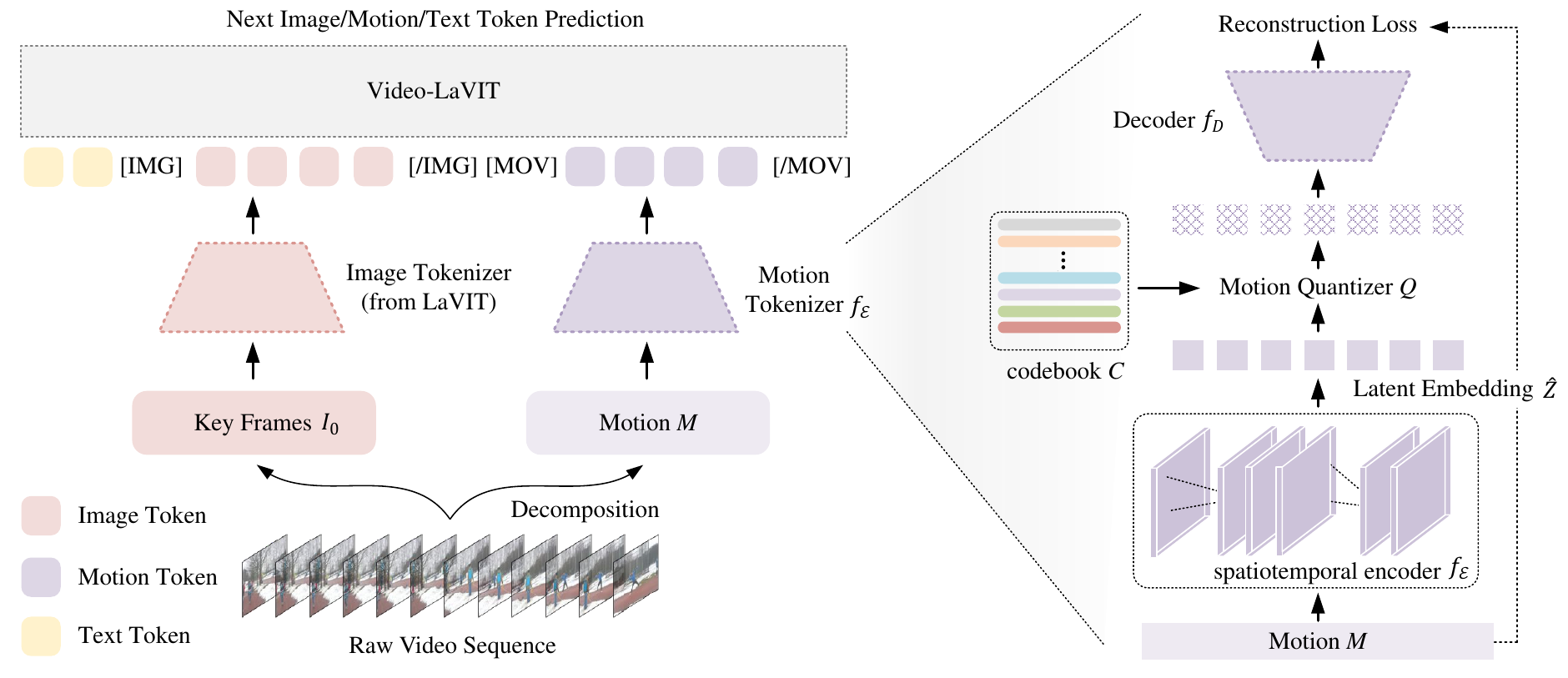}
\end{center}
\vspace{-0.15in}
\caption{For each video-text pair, \ours~decomposes the video into keyframes and motion vectors for efficient tokenization. The tokenizers are learned by maximally reconstructing original inputs (e.g., the motion tokenizer is shown on the right). Finally, the encoded tokens are concatenated with text tokens to form a multimodal sequence, allowing for unified generative pre-training of the LLM (left).}
\label{fig:pipeline}
\end{figure*}

\textbf{Video understanding and generation}. By unifying videos in the above pre-training framework, remarkable progress has been made in video comprehension with masked~\cite{yang2022zero} and autoregressive language models~\cite{li2023videochat, zhang2023video, maaz2023video}. However, for video generation, the mainstream approaches are still based on diffusion models~\cite{sohl2015deep, song2019generative, ho2020denoising}, which enhance existing image pre-trained models with better temporal consistency~\cite{ho2022video, singer2023make, blattmann2023align, esser2023structure, blattmann2023stable}. Language model based counterparts~\cite{yan2021videogpt, hong2023cogvideo, kondratyuk2023videopoet}, on the other hand, face the critical challenge of efficiently encoding video temporal dynamics with limited context windows and computational resources. In response, our work leverages motion vectors, a classic and effective cue in video modeling~\cite{zhang2016real, wang2023videocomposer, shen2024decouple}, for improving the efficacy of LLM-based video comprehension and generation.

\section{Method}
This work aims to present an effective pre-training framework that harnesses the exceptional modeling capability of Large Language Models (LLMs) to facilitate the learning of video modality. In pursuit of this goal, we highlight two core designs: a video \textit{tokenizer} (\cref{sec:tokenization}) which allows for the representation of all modalities in a unified discrete form, and a video \textit{detokenizer} (\cref{sec:detokenization}) to map the generated discrete tokens back to the continuous pixel space. Coped with these two main components, \ours~can be optimized through a unified autoregressive training paradigm (\cref{sec:pre_training}), enabling it to simultaneously comprehend and generate various multimodal content.

\subsection{Video Tokenization}
\label{sec:tokenization}


To encode an untrimmed video as inputs to LLMs, the prevailing approaches~\cite{lin2023video,li2023videochat} mainly uniformly downsample the original video into a series of frames. Then, a pre-trained ViT encoder~\citep{radford2021learning,fang2023eva} is employed to separately encode these frames and produce a sequence of frame-level embeddings as the video representation. This straightforward way disregards the modeling of temporal dynamics between frames, thus impeding the capacity to understand the actions and camera transitions occurring in the video. While the utilization of 3D video encoders in very recent~\cite{kondratyuk2023videopoet} enables the encoding of temporal information, it only applies to short video clips and inevitably yields a substantial proliferation of tokens (e.g., 1280 tokens for one 2.2s clip), resulting in a heavy computational overhead.


\textbf{Motion-aware Video Decomposition}. Given the above concerns, our proposed video tokenizer seeks to integrate temporal dynamics into the video representations efficiently. We observe that a video clip captured in the same shot can convey its primary semantics through a single keyframe, while the subsequent frames only illustrate the temporal evolvement based on that keyframe. This property empowers the decomposed video tokenization for keyframe and temporal motion. For the keyframe, we employ an off-the-shelf image tokenizer from LaVIT~\cite{jin2024unified} to inherit the learned visual codebook and prior knowledge without training from scratch. For encoding temporal motion information, a common alternative is to calculate hand-crafted dense optical flow between adjacent frames~\cite{beauchemin1995computation}. Despite providing a fine-grained depiction of object motions in videos, the expensive computations render it unsuitable for scaling to large-scale video data during pre-training. Hence, we resort to motion vectors, which can be directly extracted at high speed on the CPU~\cite{wu2018compressed} during the compressed video decoding process. 

\begin{figure*}[t]
\begin{center}
\includegraphics[width=0.95\linewidth]{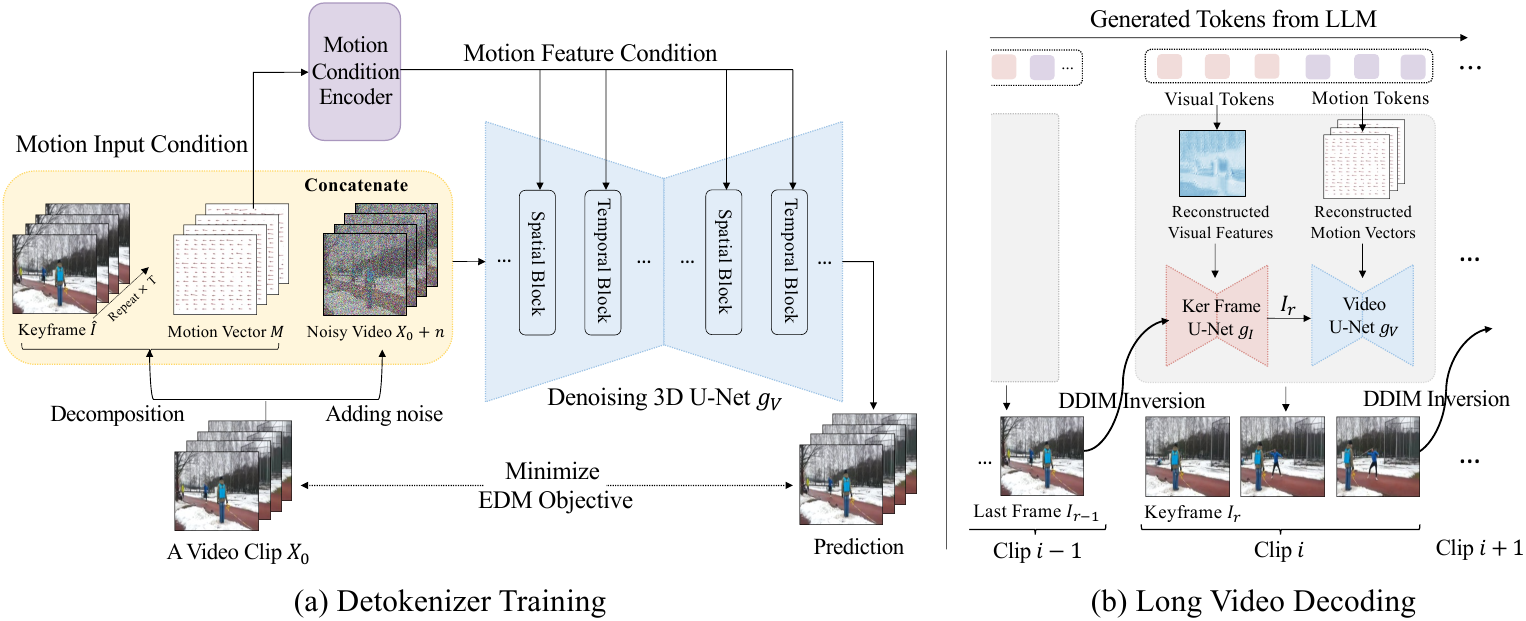}
\end{center}
\vspace{-0.15in}
\caption{Illustrations for video detokenization in \ours. (a) Training pipeline for the video detokenizer, which aims to reconstruct the original video clip using one keyframe and the subsequent motion vectors. (b) Autoregressive inference for long video decoding.}
\refstepcounter{subfigure}\label{fig:detokeize_a}
\refstepcounter{subfigure}\label{fig:detokeize_b}
\label{fig:detokeize}
\vspace{-0.05in}
\end{figure*}

As illustrated in~\cref{fig:pipeline}, we employ the MPEG-4~\cite{le1991mpeg} compression technique to extract keyframe and motion information. For simplicity, the I-frames in MPEG-4 are considered as the keyframes requiring tokenization. More sophisticated (but expensive) keyframe selection schemes can also be considered, but are not the main focus of this work. Formally, each video frame is partitioned into $16 \times 16$ non-overlapping macroblocks. Motion vectors $\vec{m}$ of the $t$-th frame are estimated by finding the best macroblock correspondence between adjacent frames $I_t$ and $I_{t-1}$ :
\begin{equation}
    \vec{m}(p,q) = \arg \min_{i, j} \| I_t(p,q) - I_{t-1}(p-i, q-j) \|,
\end{equation}
where $I(p,q)$ indicates the pixel values of the macroblock at location $(p,q)$, and $(i,j)$ is the coordinate offset between the center of two macroblocks. Then, a video clip can be decomposed into a keyframe $I_0 \in \mathbb{R}^{H \times W \times 3}$ and the motion vectors $M \in \mathbb{R}^{T \times \frac{H}{16} \times \frac{W}{16} \times 2}$ of its subsequent $T$ frames. 


\textbf{Motion Vector Tokenization}. To transform the motion vectors into a sequence of discrete tokens like a foreign language, we develop a motion-specific tokenizer based on the VQ-VAE architecture~\cite{van2017neural}. It includes a spatiotemporal encoder $f_{\mathcal{E}}$, a learnable codebook $\mathcal{C}=\{c_k\}_{k=1}^{K}$, and a decoder $f_{\mathcal{D}}$. The encoder $f_{\mathcal{E}}$ has $L$ stacked transformer blocks consisting of spatial and temporal attention layers to fuse the contextual motion information among the $T$ frames. It maps the motion vectors  $M \in \mathbb{R}^{T \times \frac{H}{16} \times \frac{W}{16} \times 2}$ into a 1D latent embedding sequence $\hat{Z} \in {\mathbb{R}^{N \times d}}$. Each embedding vector $\hat{z} \in \mathbb{R}^{d}$ is then tokenized by a vector quantizer $Q$, which assigns it to the closest code in $\mathcal{C}$: 
\begin{equation}
    z_i = \arg \min_{j} \| l_{2}(\hat{z_i}) - l_{2}(c_j)\|_2,
\end{equation}
where $l_2$ indicates the $L_2$ normalization. The decoder $f_{\mathcal{D}}$ has a similar structure to the encoder and is obliged to map the discrete motion codes $\{z_i\}_{i=1}^N$ back to the original motion vectors. The whole motion tokenizer can be updated by optimizing the reconstruction quality. To prevent codebook collapse during training, we follow Yu et al. (\citeyear{yu2022vector}) to project the motion embeddings $\hat{Z}$ into a low-dimensional space before quantization and use exponential moving average (EMA) updates. More details about the motion tokenizer can be found in~\cref{app:model}. Finally, a video is tokenized into alternating $\langle visual, motion, ... \rangle$ codes that serve as the supervision signals in LLMs during generative pre-training. Such a factorized tokenization significantly reduces the inter-frame redundancy in one video shot while efficiently capturing the temporal motion information.

\subsection{Video Detokenization}
\label{sec:detokenization}
The video detokenizer of \ours~is in charge of converting them back into the original continuous pixel space for video generation. Considering the challenge in learning a direct mapping from discrete tokens to the high-dimensional video space, we take a sequential decoding strategy, wherein the keyframe is initially recovered based on the visual token. The subsequent frames are then decoded by taking both the keyframe and motion tokens as the conditions. The efficacy of this strategy in enhancing video generation quality has also been validated by recent work~\cite{girdhar2023emu}. 

Specifically, the keyframe and video detokenizers both use conditional denoising U-Net~\cite{rombach2022high}. Similar to LaVIT~\cite{jin2024unified}, the keyframe U-Net $g_I$ takes the reconstructed visual features that contain image semantics as conditions to infill visual details from a Gaussian noise. Here, we primarily focus on the newly proposed video detokenizer $g_V$. As illustrated in~\cref{fig:detokeize_a}, it is a 3D variant of the original 2D U-Net architecture by inserting temporal convolution and attention layers after the spatial modules, following Blattmann et al. (\citeyear{blattmann2023align,blattmann2023stable}).

\textbf{Enhanced Motion Conditioning}. The objective of the video detokenizer $g_V$ is to rigorously adhere to the guidance of the motion vectors, thereby facilitating the recovery of $T$ frames following the keyframe. To this end, we highlight two different forms of motion conditions in $g_V$. Given the motion vectors $M \in \mathbb{R}^{T \times \frac{H}{16} \times \frac{W}{16} \times 2}$ of a sampled video clip, we adopt the nearest neighbor interpolation to ensure that it matches the spatial shape of the U-Net input. Also, the latent state $\hat{I}$ of the keyframe from the VAE is repeated $T$ times along the temporal axis to form visual conditioning. The motion vector $M$, the keyframe latent $\hat{I}$, and the noisy video frames are concatenated channel-wise as the input condition to $g_V$. Except for direct input conditioning, we also enhance conditioning with motion feature embedding via the spatial and temporal cross-attention layers in the 3D U-Net blocks. Here, the motion features are from a conditioning encoder that has a similar architecture to $f_{\mathcal{E}}$ excluding the downsample layer to reduce potential information loss. The parameters of the video detokenizer $g_V$ are updated by minimizing the following EDM training objective~\cite{karras2022elucidating} on a video training dataset $\mathcal{D}$:
\begin{equation}
\mathbb{E}_{(X_0, \hat{I}, \hat{M}) \sim \mathcal{D}, \sigma, n} \left[\lambda_{\sigma} || g_{V}(X_0 + n, \sigma, \hat{I}, M) - X_0 || \right],
\end{equation}
where $\sigma \sim p(\sigma)$ is the noise level during training, $n \sim \mathcal{N}(n;0,\sigma^2)$ is a random noise added to video sample $X_0$, and $\lambda_\sigma$ is loss weighting function. At inference, the $\langle visual, motion \rangle$ tokens yielded by LLM are first mapped into visual features and motion vectors by their corresponding tokenizers. The reconstructed visual features are fed into~$g_I$ to generate a keyframe, which is subsequently combined with reconstructed motion vectors to serve as conditions for~$g_V$ to decode the video clip (See~\cref{fig:detokeize_b}). 


\textbf{Long Video Decoding}. Since a video is expressed as multiple alternating $\langle visual, motion \rangle$ sequences, the interdependencies among different video fragments can be effectively learned by autoregressive pre-training of LLMs. Hence, \ours~naturally supports the generation of longer videos by progressive decoding multiple clips. However, separate decoding will bring inconsistencies in some fine-grained visual details among different clips (See~\cref{fig:long}). To mitigate this, we incorporate an explicit noise constraint when decoding the keyframe $I_{r}$ of a video clip. As illustrated in~\cref{fig:detokeize_b}, we reverse its last frame $I_{r-1}$ from the previously generated clip into an intermediate noisy state $x_{\Delta T}$ by reversing the DDIM sampling~\cite{song2021denoising} process $\Delta T$ times. Each inversion step is formulated by:
\vspace{-0.03in}
\begin{equation}
\label{eq:inverse}
\small
x_{t+1} = \sqrt{\frac{\alpha_{t+1}}{\alpha_t}} x_{t} + \left ( \sqrt{\frac{1}{\alpha_{t+1}} - 1} - \sqrt{\frac{1}{\alpha_t} - 1}\right) g_I(x_{t}, t, \hat{I}),
\end{equation}
where $\alpha_t$ is the noise level, $\hat{I}$ is the visual feature condition. The reversed noisy state $x_{\Delta T}$ is then considered as the initial noise in the denoising loop for keyframe $I_{r}$. As illustrated in~\cref{fig:long}, adding this noise constraint can improve the temporal consistency between video clips.


\subsection{Unified Generative Modeling}
\label{sec:pre_training}
Based on the developed decomposed video tokenization strategy, it is feasible to indiscriminately treat all the modalities (video, image, and text) as 1D discrete tokens fed into LLMs. Following LaVIT~\cite{jin2024unified}, special tokens (e.g., [MOV] and [/MOV] for motion modality) are inserted at the beginning and end of the visual and motion token sequence for differentiating modalities in the input data. During pre-training, we also exchange the order of multimodal data pairs to form both $[\text{video}(\text{image}), \text{text}]$ and $[\text{text},\text{video}(\text{image})]$ as input sequences. Formally, given a multimodal sequence $y=(y_1, y_2, .., y_S)$, \ours~inherits the successful generative language modeling paradigm from LLM to directly maximize the likelihood of each token $y_i$ in an autoregressive manner:
\begin{equation}
    p(y) = \sum_{y \in \mathcal{D}} \sum_{i=1}^{S} \log P_\theta(y_i | y_{< i}).
\end{equation}
After pre-training, \ours~is capable of serving as a multimodal generalist to achieve both multimodal comprehension and generation of data in any modality.


\textbf{Model Training.}
\ours~undergoes a three-stage training procedure on the large-scale multimodal corpora. The purpose of each stage can be summarized as follows: i) Tokenizer and Detokenizer Training. This stage requires only pure video data without corresponding textual captions. It aims to produce compact video tokens that serve as supervision signals to guide the subsequent generative pre-training, as well as to facilitate an accurate reconstruction of the original videos. ii) Generative Pre-training. Stage-2 empowers the model to learn the inter-correlation among the data of different modalities via unified generative modeling within the LLM. iii) Instruction Tuning. To fully unleash the acquired knowledge, the last stage further improves the instruction-following ability to accomplish various multimodal tasks. More details about the model architectures and training data for each stage are provided in~\cref{app:model}.

\section{Experiments}

\subsection{Multimodal Understanding}

With the decomposed video representation, \ours~is naturally capable of understanding both videos and images. Here, we demonstrate its multimodal understanding capability on 11 commonly used image and video benchmarks.

\begin{table*}[t]
\vskip -0.1in
\caption{Image understanding performance ($\uparrow$) on 8 benchmarks. \ours~achieves state-of-the-art results on most of the benchmarks. For convenience, SQA\textsuperscript{I} denotes ScienceQA-IMG~\cite{lu2022learn}, and MMB denotes MMBench~\cite{liu2023mmbench}. *~indicates that there is some overlap with the training data. Note that only LLaVA-1.5~\cite{liu2023improved} is reported with a higher image resolution of 336. The Video-LLaVA, LLaMA-VID and LLaVA-1.5 use Vicuna-1.5~\cite{chiang2023vicuna} as the language model.}
\label{tab:image_qa}
\vskip 0.1in
\begin{center}
\begin{small}
\begin{tabular}{lccccccccc}
\toprule
\multirow{2}{*}{Method} & \multirow{2}{*}{LLM size} & \multicolumn{4}{c}{Image Question Answering} & \multicolumn{4}{c}{Multimodal} \\
\cmidrule(lr){3-6} \cmidrule(l){7-10} & & VQA\textsuperscript{v2} & GQA & VizWiz & SQA\textsuperscript{I} & MME & MMB & SEED & MM-Vet \\
\midrule
Flamingo~\cite{alayrac2022flamingo} & 9B & 51.8 & - & 28.8 & - & - & - & - & - \\
BLIP-2~\cite{li2023blip} & 13B & 41.0 & 41.0 &  19.6 & 61.0 & 1293.8 & - & 46.4 & 22.4 \\
InstructBLIP~\cite{dai2023instructblip} & 13B & - & 49.5 & 34.3 & 63.1 & 1212.8 & 44.0 & - & 25.6 \\
CM3Leon~\cite{yu2023scaling} & 7B & 47.6 & - & 37.6 & - & - & - & - & - \\
Emu~\cite{sun2024emu} & 13B & 52.0 & - & 34.2 & - & - & - & - & 36.3 \\
DreamLLM~\cite{dong2023dreamllm} & 7B & 72.9* & - & 49.3 & - & - & 58.2 & - & \textbf{36.6} \\
Video-LLaVA~\cite{lin2023video} & 7B & 74.7* & 60.3* & 48.1 & 66.4 & - & 60.9 & - & 32.0 \\
LLaMA-VID~\cite{li2023llama} & 7B & 78.3* & 63.0* & 52.5 & 67.7 & 1405.6 & 65.3 & 59.7 & - \\
LLaVA-1.5~\cite{liu2023improved} & 7B & 78.5* & 62.0* & 50.0 & 66.8 & 1510.7 & 64.3 & 58.6 & 30.5 \\
\cmidrule{1-10} \ours & 7B & \textbf{80.3}* & \textbf{64.4}* & \textbf{56.0} & \textbf{70.0} & \textbf{1551.8} & \textbf{67.3} & \textbf{64.0} & 33.2 \\
\bottomrule
\end{tabular}
\end{small}
\end{center}
\vskip -0.1in
\end{table*}

\begin{table*}[t]
\vskip -0.1in
\caption{Zero-shot video question answering accuracy ($\uparrow$). \ours~demonstrates state-of-the-art accuracy on all three benchmarks. The evaluation uses a GPT assistant~\cite{maaz2023video}, with ``Score'' denoting a relative score from 0 to 5 assigned by the GPT model. The Video-LLaVA and LLaMA-VID both use Vicuna-1.5~\cite{chiang2023vicuna} as the language model.}
\label{tab:video_qa}
\vskip 0.1in
\begin{center}
\begin{small}
\setlength{\tabcolsep}{10.25pt}
\begin{tabular}{@{\hspace{5pt}}l@{}ccccccc@{\hspace{5pt}}}
\toprule
\multirow{2}{*}{Method} & \multirow{2}{*}{LLM size} & \multicolumn{2}{c}{MSVD-QA} & \multicolumn{2}{c}{MSRVTT-QA} & \multicolumn{2}{c}{ActivityNet-QA} \\
\cmidrule(lr){3-4} \cmidrule(lr){5-6} \cmidrule(l){7-8} & & Accuracy & Score & Accuracy & Score & Accuracy & Score \\
\midrule
FrozenBiLM~\cite{yang2022zero} & 1B & 32.2 & - & 16.8 & - & 24.7 & - \\
Video-LLaMA~\cite{zhang2023video} & 7B & 51.6 & 2.5 & 29.6 & 1.8 & 12.4 & 1.1 \\
VideoChat~\cite{li2023videochat} & 7B & 56.3 & 2.8 & 45.0 & 2.5 & 26.5 & 2.2 \\
Video-ChatGPT~\cite{maaz2023video} & 7B & 64.9 & 3.3 & 49.3 & 2.8 & 35.2 & 2.7 \\
LLaMA-VID~\cite{li2023llama} & 7B & 69.7 & 3.7 & 57.7 & 3.2 & 47.4 & 3.3 \\
Video-LLaVA~\cite{lin2023video} & 7B & 70.7 & 3.9 & 59.2 & \textbf{3.5 }& 45.3 & 3.3 \\
\cmidrule{1-8} \ours & 7B & \textbf{73.2} & \textbf{3.9} & \textbf{59.3} & 3.3 & \textbf{50.1} & \textbf{3.3} \\
\bottomrule
\end{tabular}
\end{small}
\end{center}
\vskip -0.1in
\end{table*}

\textbf{Image Understanding}. \Cref{tab:image_qa} presents an extensive comparison across eight widely used image question answering and multimodal benchmarks: VQA v2~\cite{goyal2017making}, GQA~\cite{hudson2019gqa}, VizWiz~\cite{gurari2018vizwiz}, ScienceQA-IMG~\cite{lu2022learn}, MME~\cite{fu2023mme}, MMBench~\cite{liu2023mmbench}, SEED~\cite{li2023seed}, MM-Vet~\cite{yu2023mm}. Our model successfully generalizes the pre-training knowledge to image comprehension tasks and provides the best overall performance. Specifically, with the same instruction dataset and the base model as LLaVA-1.5~\cite{liu2023improved}, our method consistently yields the best results on all image question answering datasets. For example on SQA\textsuperscript{I}, it surpasses LLaVA-1.5 which has a higher input resolution by 3.2\%, while consistently outperforming the other video-language models. The same advantages are further validated on more comprehensive multimodal benchmarks, where our model leads on three out of four benchmarks. 

\begin{table*}[t]
\vskip -0.1in
\caption{Zero-shot understanding ($\uparrow$) on the test set of Perception Test~\cite{patraucean2024perception} and EgoSchema~\cite{mangalam2024egoschema}.}
\label{tab:perception}
\begin{center}
\begin{small}
\setlength{\tabcolsep}{6.5pt}
\resizebox{0.95\linewidth}{!}{
\begin{tabular}{lcccc}
\toprule
Method & Flamingo~\cite{alayrac2022flamingo} & BLIP-2~\cite{li2023blip2} & VideoChat2~\cite{li2023mvbench} & \ours \\
\midrule
Accuracy & 33.5  & 39.2  & 47.3 & \textbf{47.9} \\
\bottomrule
\end{tabular}
}

\vskip 0.02in
\label{tab:ego}
\resizebox{0.95\linewidth}{!}{
\begin{tabular}{lcccc}
\toprule
Method & FrozenBiLM~\cite{yang2022zero} & mPLUG-Owl~\cite{ye2023mplug} & InternVideo~\cite{wang2022internvideo} & \ours \\
\midrule
Accuracy & 26.9  & 28.7  & 32.1 & \textbf{37.3} \\
\bottomrule
\end{tabular}
}
\end{small}
\end{center}
\vskip -0.1in
\end{table*}

\begin{table*}[!ht]
\vskip -0.1in
\caption{Zero-shot text-to-video generation performance. \ours~delivers competitive results against state-of-the-art models trained on more proprietary data, with data size reported in terms of the number of training video clips. The next best results are underlined.}
\label{tab:t2v}
\vskip 0.1in
\begin{center}
\begin{small}
\begin{tabular}{lccccccc}
\toprule
\multirow{2}{*}{Method} & \multirow{2}{*}{Data size} & \multirow{2}{*}{Public data} & \multicolumn{3}{c}{MSR-VTT} & \multicolumn{2}{c}{UCF-101} \\
\cmidrule(lr){4-6} \cmidrule(l){7-8} & & & CLIPSIM ($\uparrow$) & FVD ($\downarrow$) & FID ($\downarrow$) & IS ($\uparrow$) & FVD ($\downarrow$) \\
\midrule
CogVideo~\cite{hong2023cogvideo} & 5.4M & \checkmark & 0.2631 & 1294 & 23.59 & 25.27 & 701.59 \\
Video LDM~\cite{blattmann2023align} & 10M & \checkmark & 0.2929 & - & - & 33.45 & 550.61 \\
VideoComposer~\cite{wang2023videocomposer} & 10M & \checkmark & 0.2932 & 580 & - & - & - \\
InternVid~\cite{wang2024internvid} & 28M & \checkmark & 0.2951 & - & - & 21.04 & 616.51 \\
Make-A-Video~\cite{singer2023make} & 20M & \checkmark & \textbf{0.3049} & - & 13.17 & 33.00 & 367.23 \\
VideoPoet~\cite{kondratyuk2023videopoet} & 270M & $\times$ & \textbf{0.3049} & \underline{213} & - & 38.44 & 355.00 \\
PYoCo~\cite{ge2023preserve} & 22.5M & $\times$ & - & - & \textbf{9.73} & \textbf{47.76} & 355.19 \\
SVD~\cite{blattmann2023stable} & 152M & $\times$ & - & - & - & - & \textbf{242.02} \\
\cmidrule{1-8} \ours & 10M & \checkmark & \underline{0.3012} & \textbf{188.36} & \underline{11.27} & \underline{44.26} & \underline{280.57} \\
\bottomrule
\end{tabular}
\end{small}
\end{center}
\vskip -0.1in
\end{table*}

\begin{figure*}[t]
\begin{center}
\begin{small}
\begin{tabular}{lc@{}c@{}c@{}cc@{}c@{}c@{}c}
& \multicolumn{4}{c}{\it A majestic eagle soars gracefully over a breathtaking} & \multicolumn{4}{c}{\it A steam train moving on a mountainside by} \\[-2pt]  
& \multicolumn{4}{c}{\it mountain range.} & \multicolumn{4}{c}{\it Vincent van Gogh.} \\
\rotatebox[origin=l]{90}{\makebox[.105\linewidth]{Gen-2}} & \includegraphics[width=0.112\linewidth]{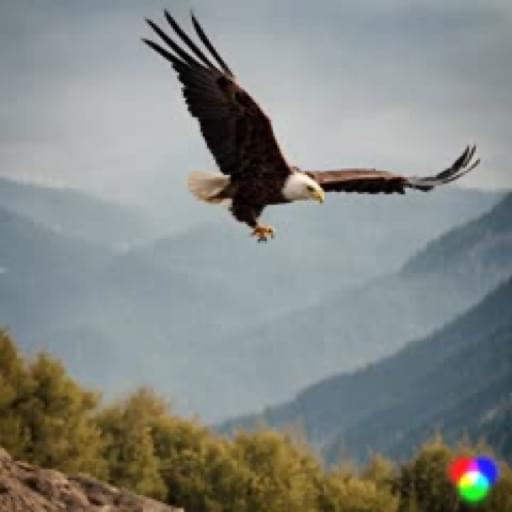} & \includegraphics[width=0.112\linewidth]{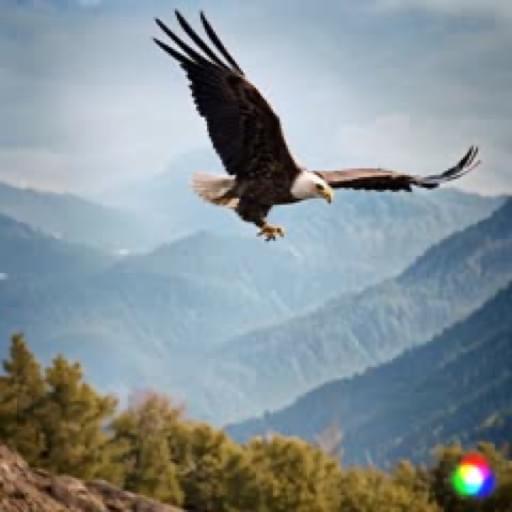} & \includegraphics[width=0.112\linewidth]{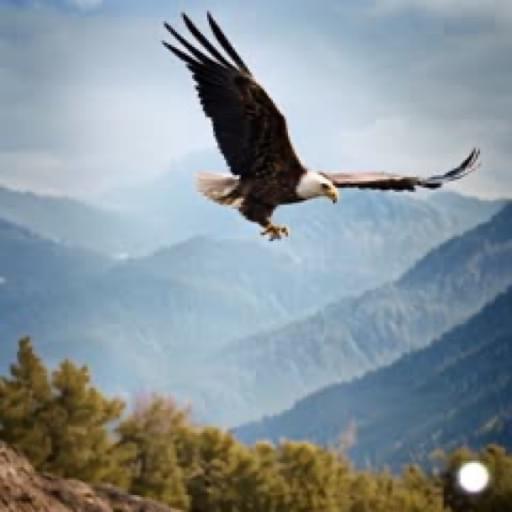} & \includegraphics[width=0.112\linewidth]{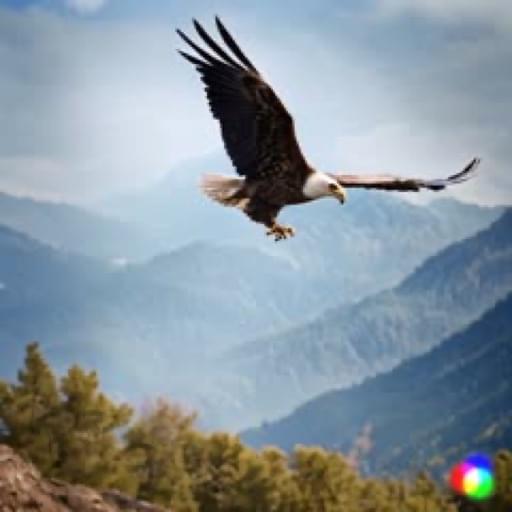} & \includegraphics[width=0.112\linewidth]{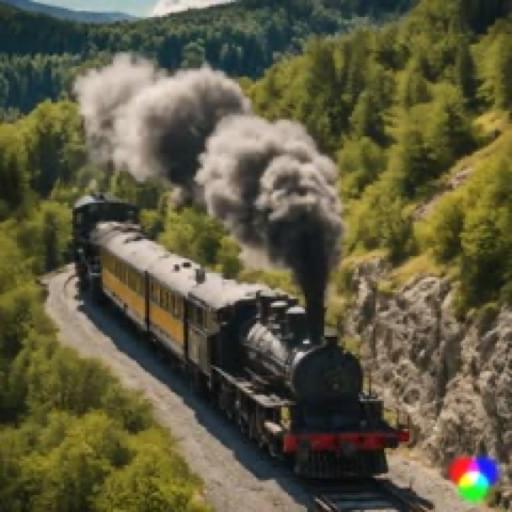} & \includegraphics[width=0.112\linewidth]{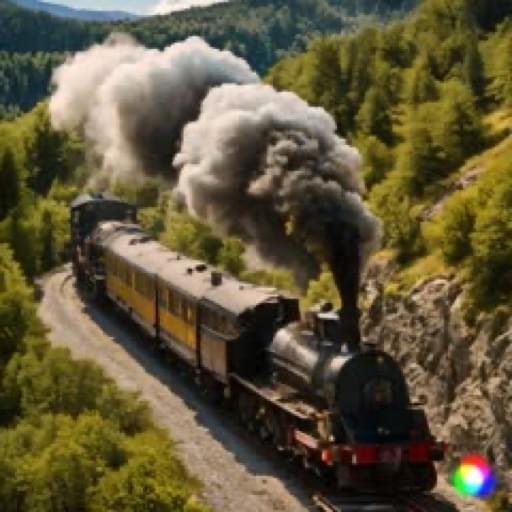} & \includegraphics[width=0.112\linewidth]{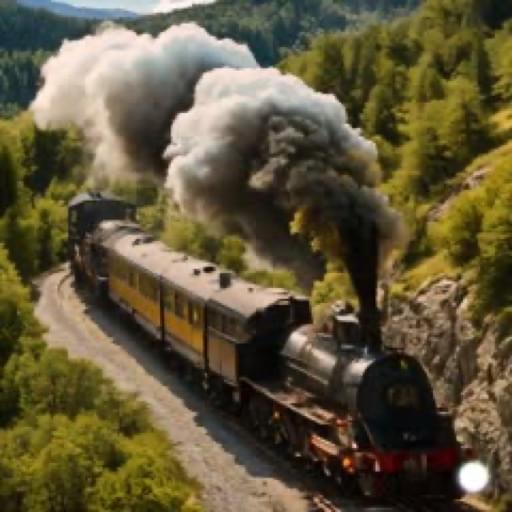} & \includegraphics[width=0.112\linewidth]{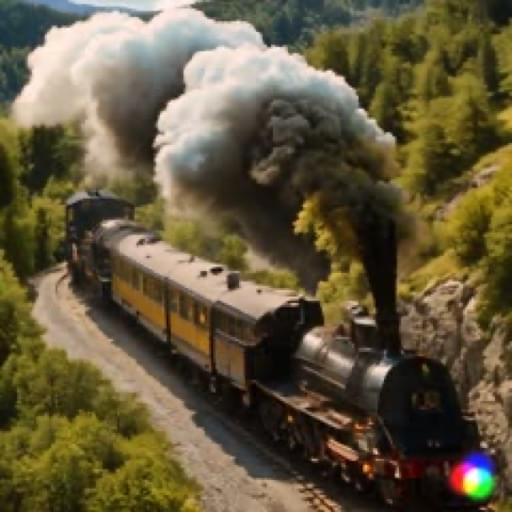} \\
\rotatebox[origin=l]{90}{\makebox[.1\linewidth]{\ours}} & \includegraphics[width=0.112\linewidth]{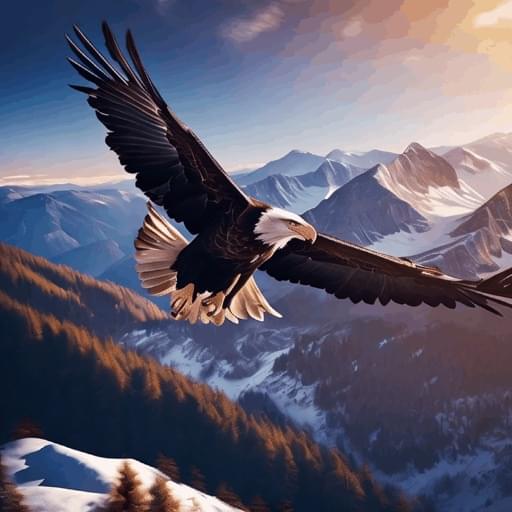} & \includegraphics[width=0.112\linewidth]{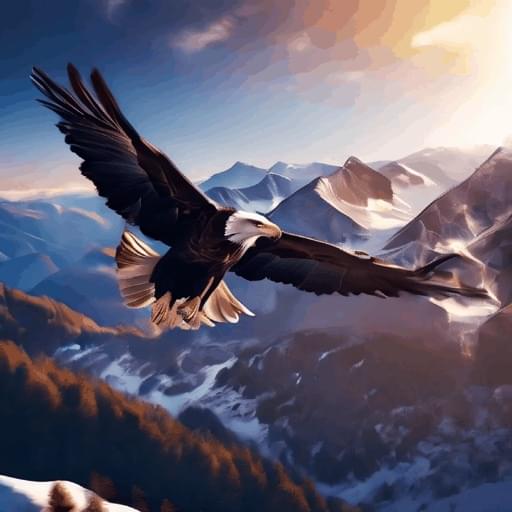} & \includegraphics[width=0.112\linewidth]{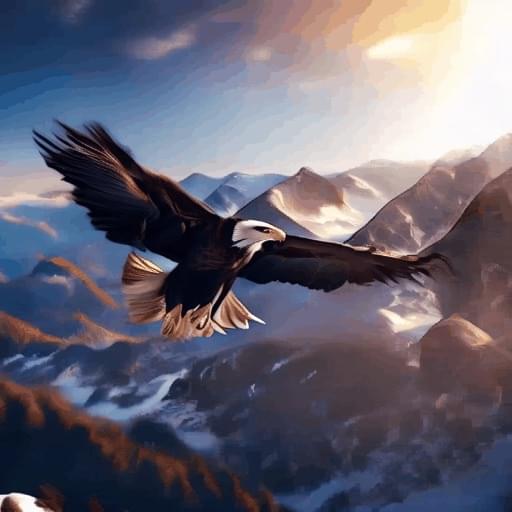} & \includegraphics[width=0.112\linewidth]{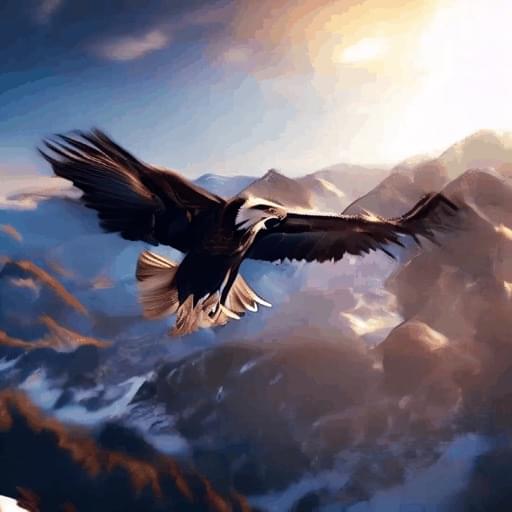} & \includegraphics[width=0.112\linewidth]{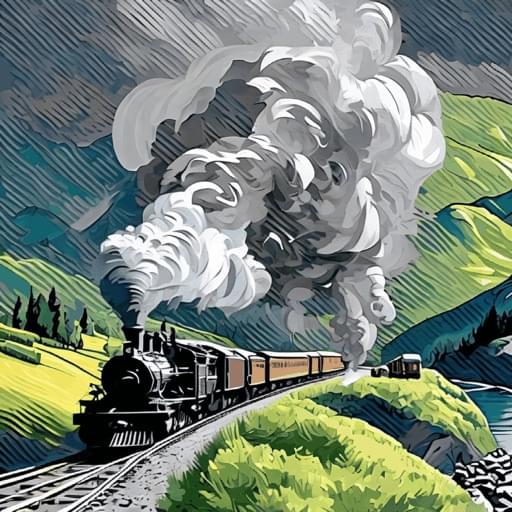} & \includegraphics[width=0.112\linewidth]{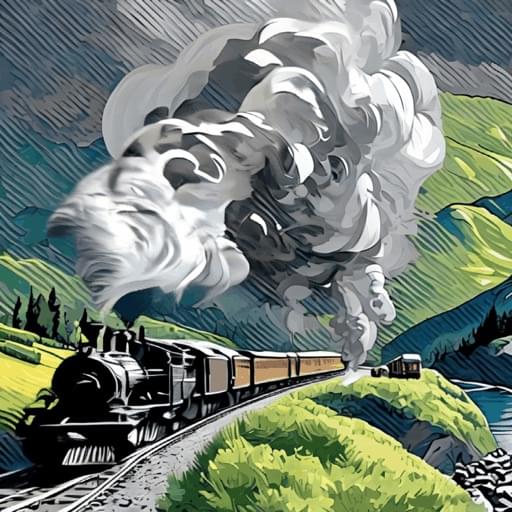} & \includegraphics[width=0.112\linewidth]{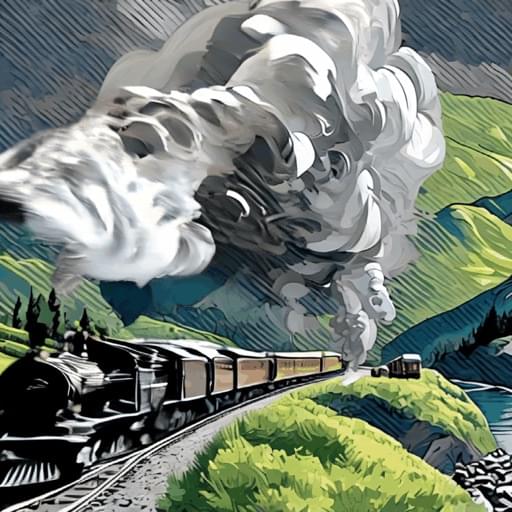} & \includegraphics[width=0.112\linewidth]{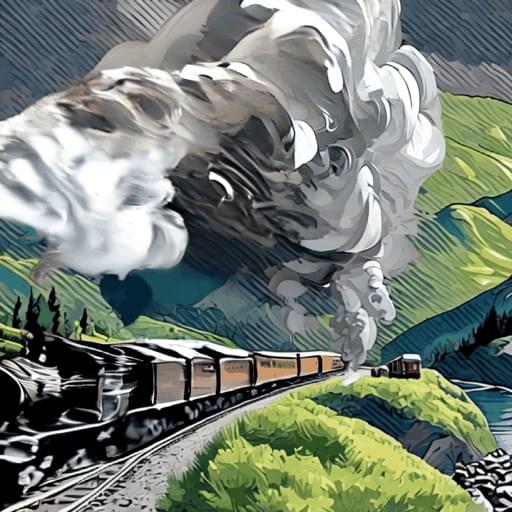} \\[-1pt]
\midrule
\rotatebox[origin=l]{90}{\makebox[.105\linewidth]{SVD}} & \includegraphics[width=0.112\linewidth]{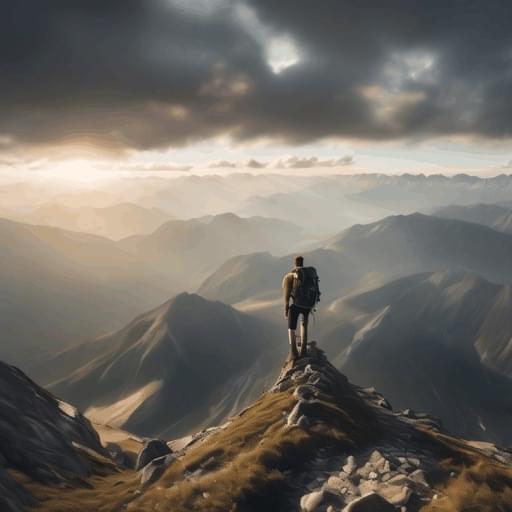} & \includegraphics[width=0.112\linewidth]{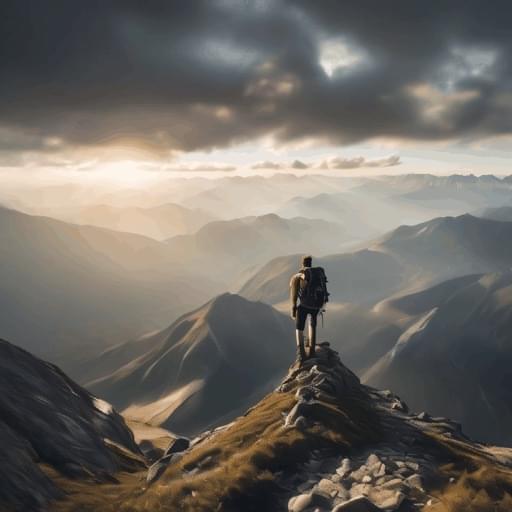} & \includegraphics[width=0.112\linewidth]{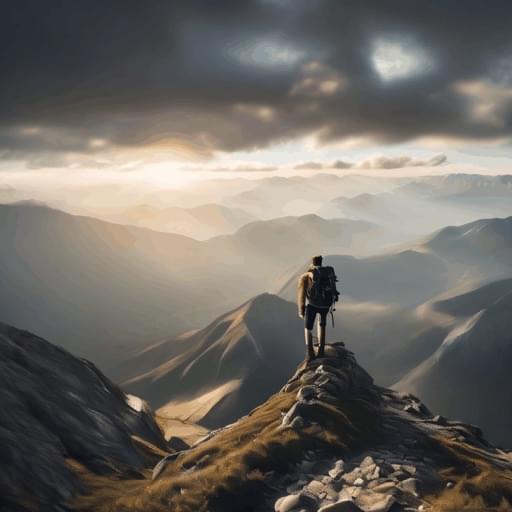} & \includegraphics[width=0.112\linewidth]{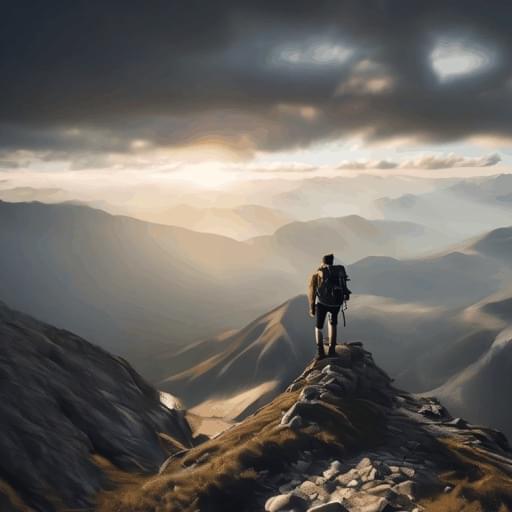} & \includegraphics[width=0.112\linewidth]{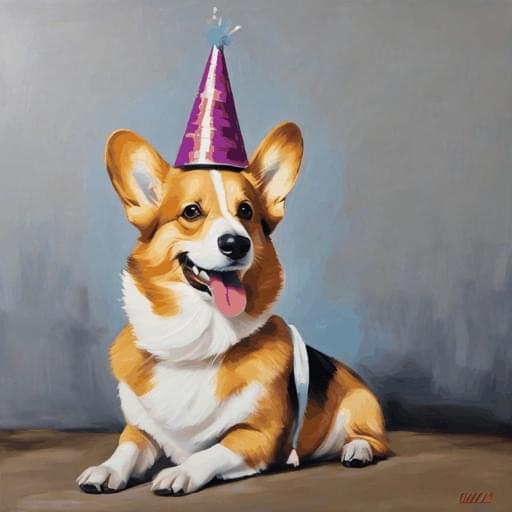} & \includegraphics[width=0.112\linewidth]{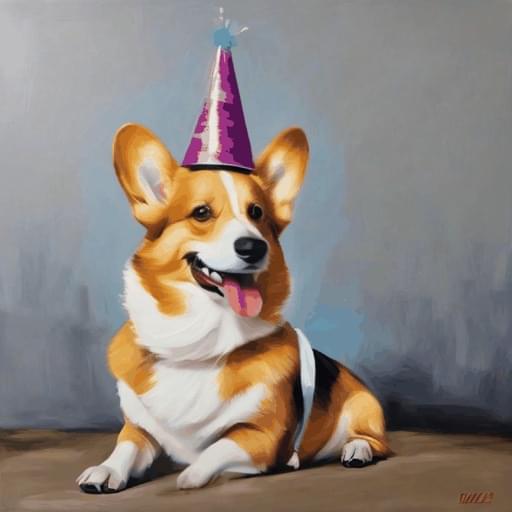} & \includegraphics[width=0.112\linewidth]{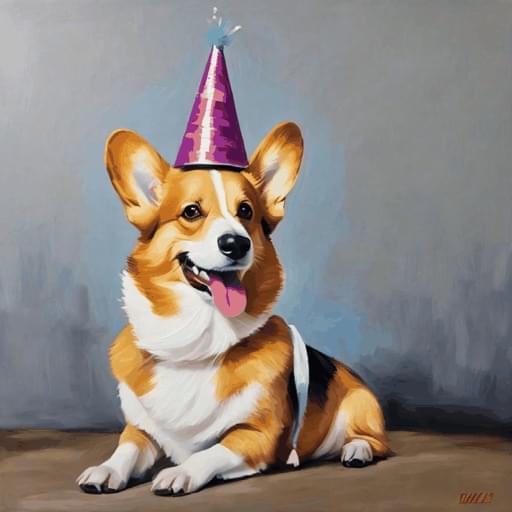} & \includegraphics[width=0.112\linewidth]{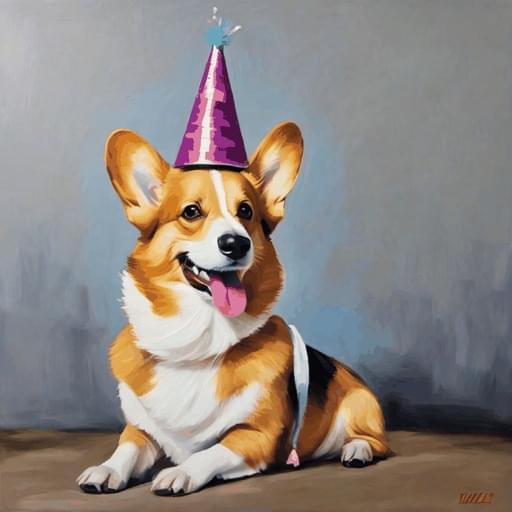} \\
\rotatebox[origin=l]{90}{\makebox[.1\linewidth]{\ours}} & \includegraphics[width=0.112\linewidth]{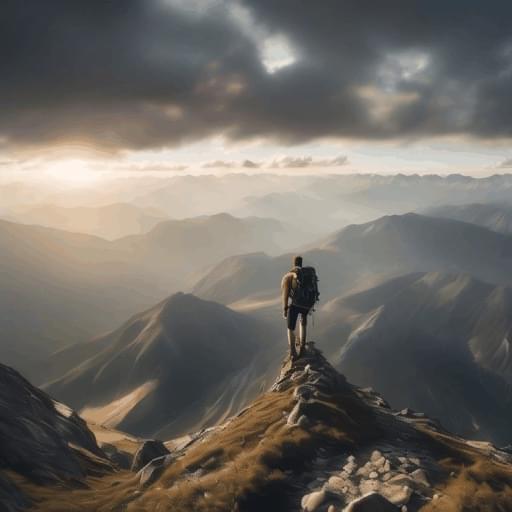} & \includegraphics[width=0.112\linewidth]{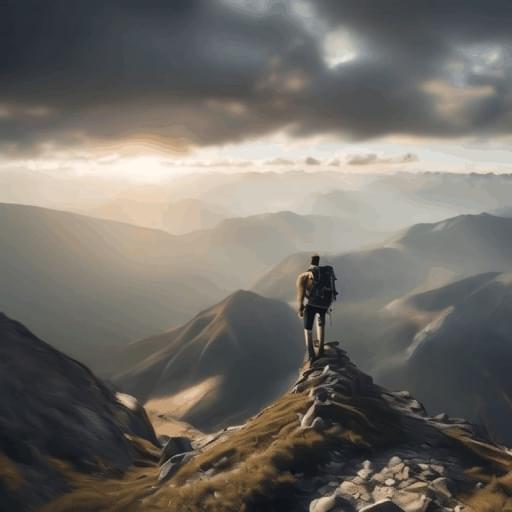} & \includegraphics[width=0.112\linewidth]{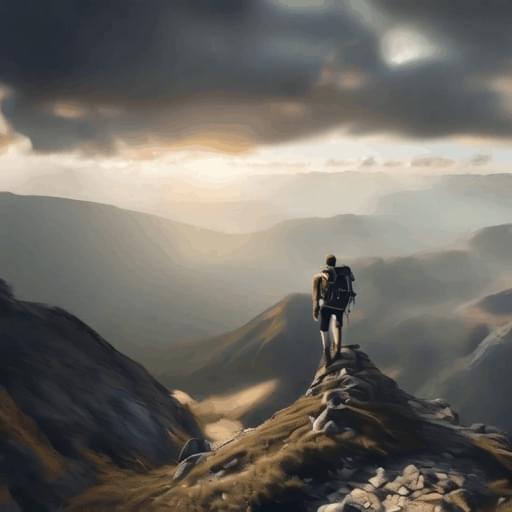} & \includegraphics[width=0.112\linewidth]{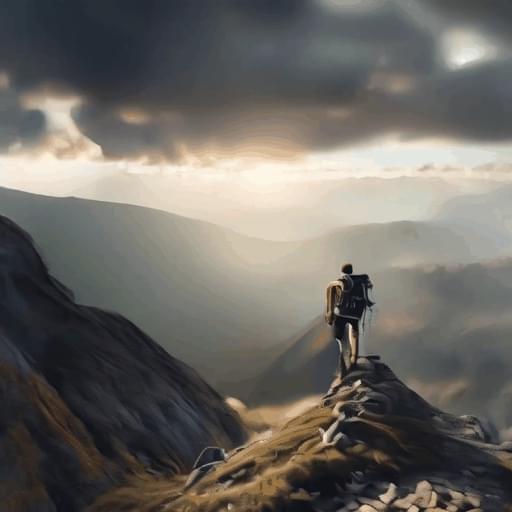} & \includegraphics[width=0.112\linewidth]{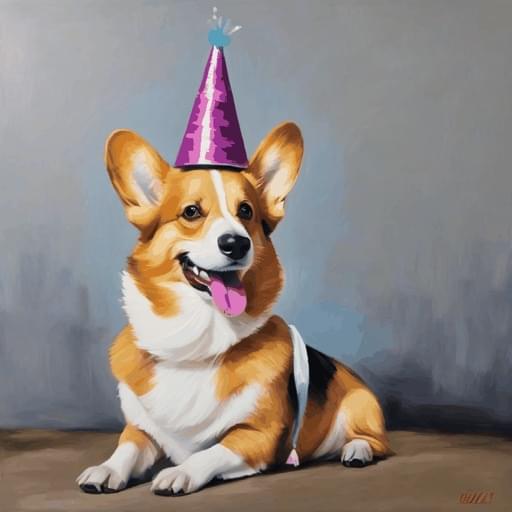} & \includegraphics[width=0.112\linewidth]{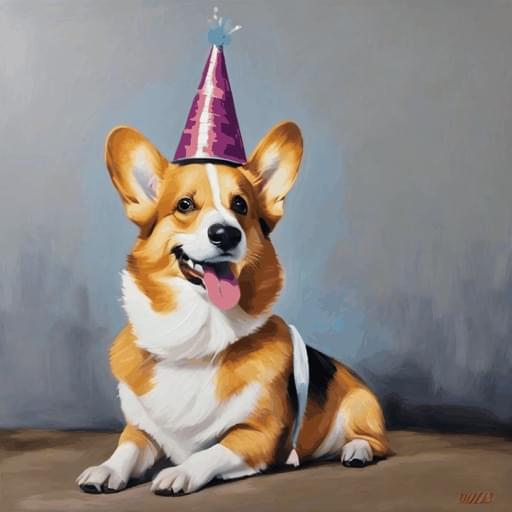} & \includegraphics[width=0.112\linewidth]{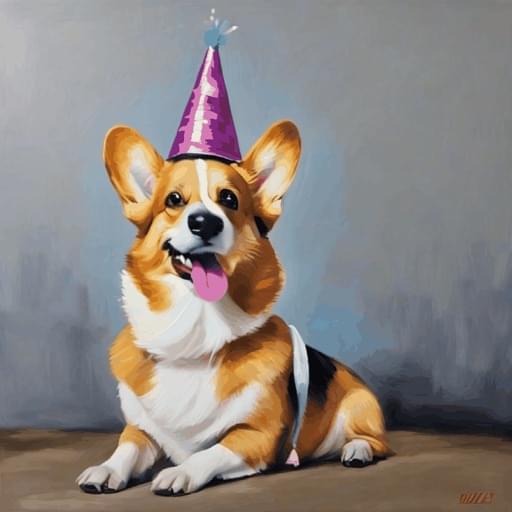} & \includegraphics[width=0.112\linewidth]{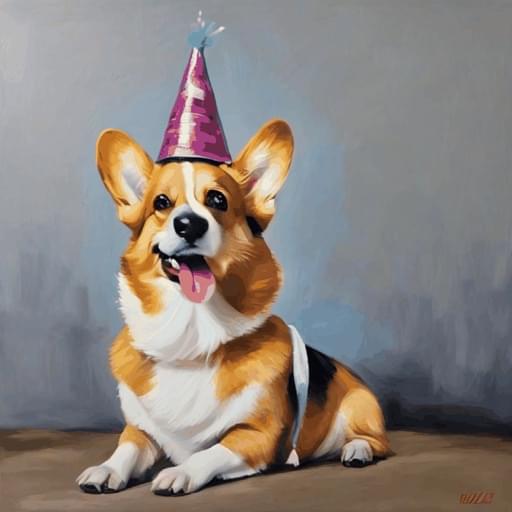} \\[-5pt]
\end{tabular}
\end{small}
\caption{Text-to-video (top) and image-to-video (bottom) generation comparison with Gen-2~\cite{runaway2023gen} and SVD-XT~\cite{blattmann2023stable}. Text prompts are from Emu Video~\cite{girdhar2023emu} and SVD. The I2V generation is conditioned on the leftmost frame.}
\label{fig:t2v_vis}
\end{center}
\vskip -0.1in
\end{figure*}

\textbf{Zero-Shot Video Question Answering}. \Cref{tab:video_qa} compares our proposed \ours~with multiple recent video-language models on three common video benchmarks: MSVD-QA~\cite{chen2011collecting}, MSRVTT-QA~\cite{xu2016msr} and ActivityNet-QA~\cite{yu2019activitynet}, in terms of accuracy and relative score measured by a GPT-3.5 assistant~\cite{maaz2023video}. We achieve state-of-the-art accuracies and very competitive relative scores on the three benchmarks, such as surpassing the previous leading model Video-LLaVA~\cite{lin2023video} by 2.5\% on MSVD-QA. Using the same 100k video-text instruction dataset from Video-ChatGPT~\cite{maaz2023video} which is also adopted by Video-LLaVA, our method outperforms these alternatives by explicitly modeling temporal dynamics with motion tokens. Especially for the ActivityNet-QA benchmark, which contains various human behaviors, incorporating motion information contributes to the recognition of different actions. For the only metric where our performance is not the best, namely the relative score on MSRVTT-QA, we deliver a high score only second to Video-LLaVA (by a margin of 0.2), again confirming the effectiveness of our method.

\textbf{Zero-Shot Video Understanding}. Besides the widely-used video question answering 
 datasets, we also evaluated \ours~on Perception Test~\cite{patraucean2024perception} or EgoSchema~\cite{mangalam2024egoschema}. These two benchmarks aim to evaluate the understanding and reasoning capability of long-term videos, rather than exploiting the hallucination capabilities of LLMs. The detailed evaluation results are shown in \Cref{tab:perception}. On the Perception Test, \ours~delivers the highest zero-shot performance. Notably, it outperforms VideoChat2~\cite{li2023mvbench}, which uses 1.9M additional instruction tuning data (including videos) to improve video understanding. In comparison, our advantageous performance is achieved with the standard instructions from LLaVA-1.5~\cite{liu2023improved} and Video-ChatGPT~\cite{maaz2023video} (amounting to 765K), demonstrating the effectiveness of our proposed method. As for EgoSchema, which focuses on long video understanding, \ours~is able to analyze 16 keyframes (with the motion vectors in between) spanning 64 seconds, thereby deliver better results. For example, it outperforms InternVideo~\cite{wang2022internvideo}, which uses up to 90 frames, by a significant 5.2\% in zero-shot QA accuracy. This validates the efficacy of the visual-motional decomposition for modeling long-term temporal information.

\subsection{Multimodal Generation}

By unified generative pre-training, \ours~can flexibly generate both video and images. Due to page limitations, we present here its text-to-video generation results, while the text-to-image evaluation is discussed in~\cref{app:gen}.

\textbf{Zero-Shot Text-to-Video Generation}. \Cref{tab:t2v} summarizes the model performance on MSR-VTT~\cite{xu2016msr} and UCF-101~\cite{soomro2012dataset}, in terms of CLIP similarity (CLIPSIM)~\cite{wu2021godiva}, Fr{\'e}chet video distance (FVD)~\cite{unterthiner2018towards}, Fr{\'e}chet Inception distance (FID)~\cite{heusel2017gans}, and Inception score (IS)~\cite{saito2020train}. Overall, our model significantly outperforms most baselines using similar public datasets, and is highly competitive against models trained on much larger proprietary data, for example leading the FVD on MSR-VTT. In particular, when compared to language model-based text-to-video generators, our method consistently outscores CogVideo~\cite{hong2023cogvideo}, while surpassing the recent concurrent work VideoPoet~\cite{kondratyuk2023videopoet}, which uses a 3D video tokenizer trained on the much larger data. This clearly validates the superiority of our tokenizer design.

\textbf{Zero-Shot Long Video Generation}. We also conducted quantitative evaluation experiments for long video generation, following the setting from FreeNoise~\cite{qiu2023freenoise}. Specifically, it is evaluated on 2048 long videos (64 frames) generated using the prompts from EvalCrafter~\cite{liu2023evalcrafter}. As shown in the table \Cref{tab:long_video_quant}, our approach yields highly competitive performance among the specialists curated for long video generation. In particular, it surpasses all baselines on the KVD metric, which measures the discrepancy between short videos (first 16 frames) and subsets of long videos (last 16 frames). These results confirm the effectiveness of our proposed long video decoding strategy with explicit noise constraint.

\subsection{Qualitative Results}
This section compares videos created by \ours~with state-of-the-art results under both text and image conditions. It also presents our special ability to generate long videos. More visualization examples are provided in~\cref{app:gen}.

\begin{figure*}[t]
\begin{center}
\begin{small}
\begin{tabular}{ll@{\hskip 5pt}c@{}c@{}c@{}c@{}c@{}c@{}c@{}c@{}c@{}c}
\multirow{2}{*}[30pt]{\rotatebox{90}{\makebox[.2\linewidth]{\ours}}} & \rotatebox[origin=l]{90}{\makebox[.1\linewidth]{\scriptsize 0--48 frames}} & \includegraphics[width=0.112\linewidth]{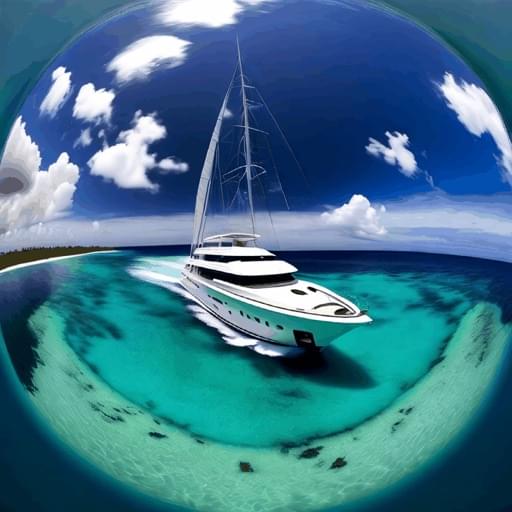} & \includegraphics[width=0.112\linewidth]{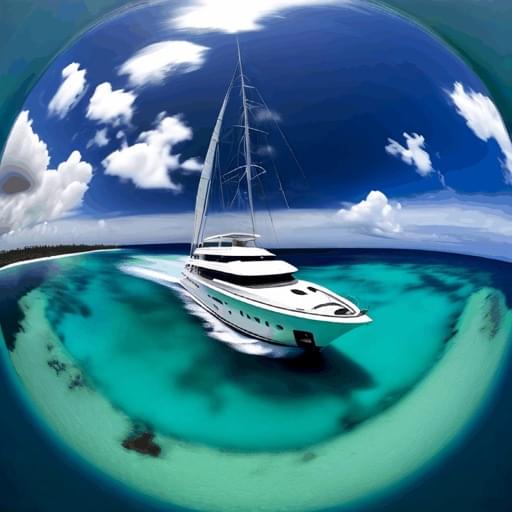} & \includegraphics[width=0.112\linewidth]{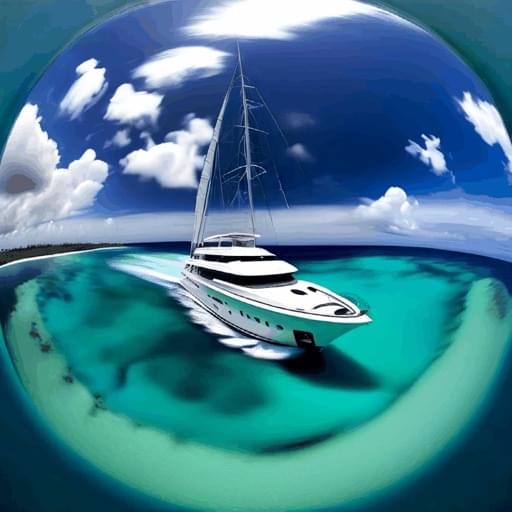} & \includegraphics[width=0.112\linewidth]{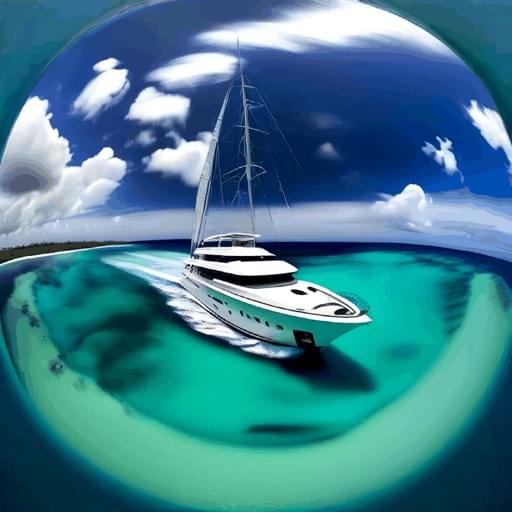} & \includegraphics[width=0.112\linewidth]{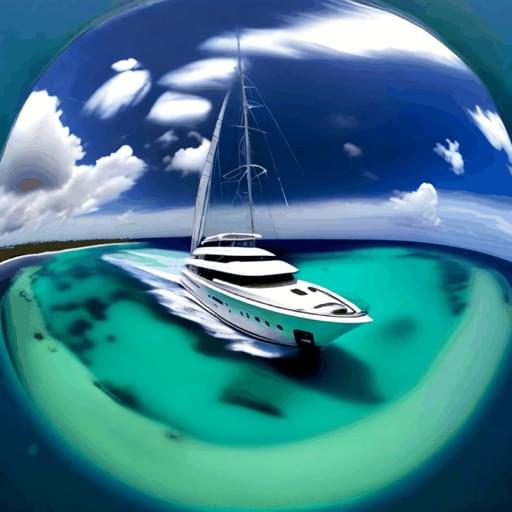} & \includegraphics[width=0.112\linewidth]{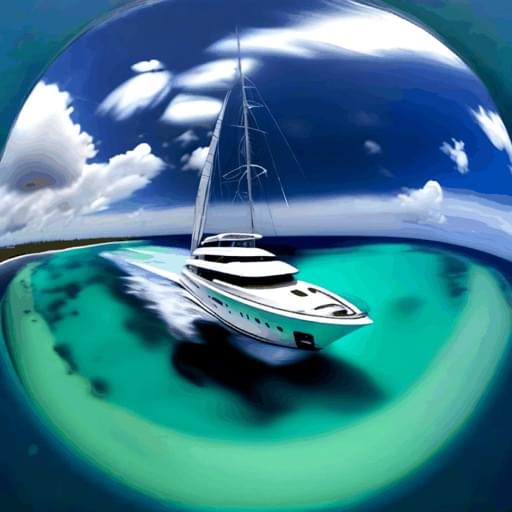} & \includegraphics[width=0.112\linewidth]{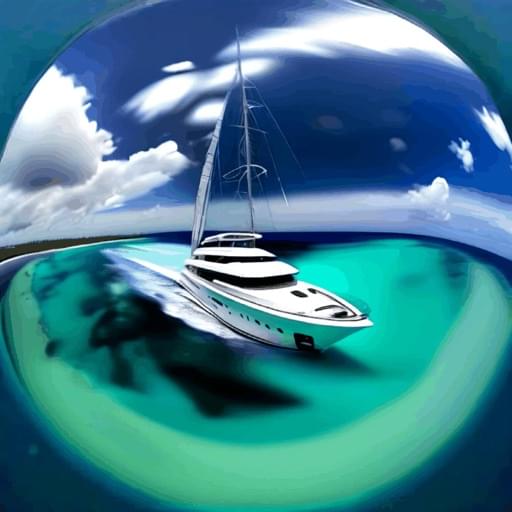} & \includegraphics[width=0.112\linewidth]{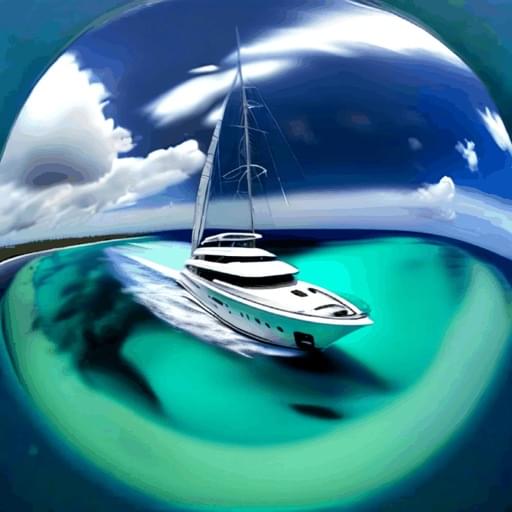} \\
& \rotatebox[origin=l]{90}{\makebox[.1\linewidth]{\scriptsize 48--96 frames}} & \includegraphics[width=0.112\linewidth]{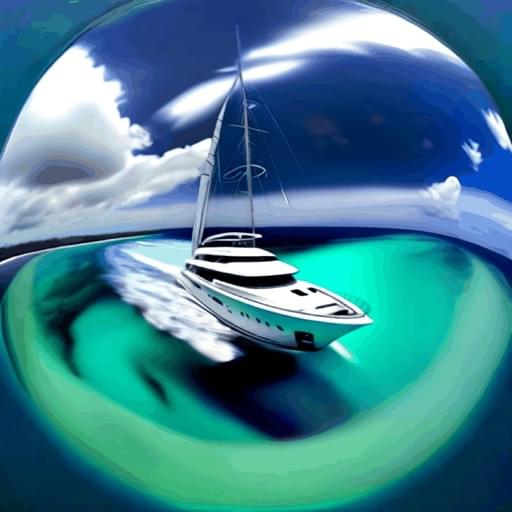} & \includegraphics[width=0.112\linewidth]{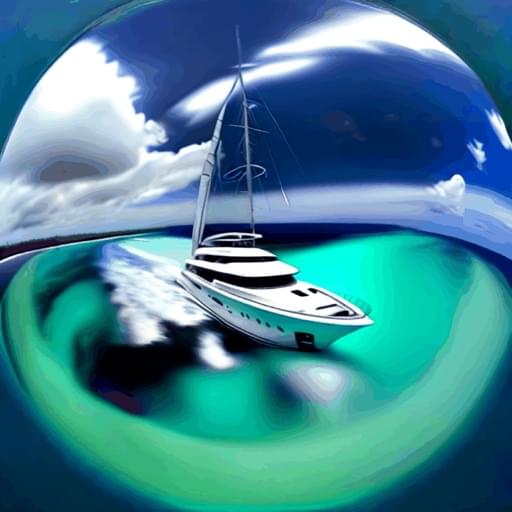} & \includegraphics[width=0.112\linewidth]{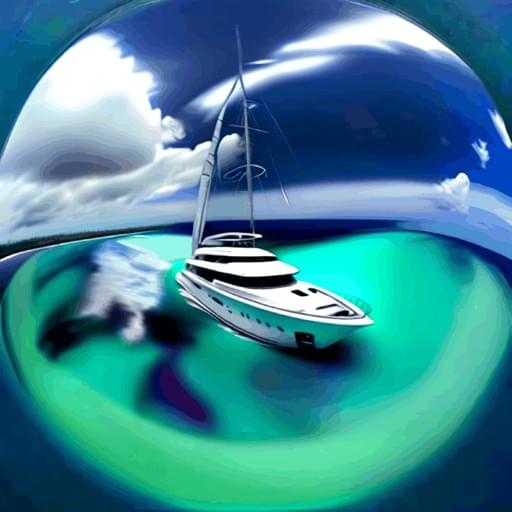} & \includegraphics[width=0.112\linewidth]{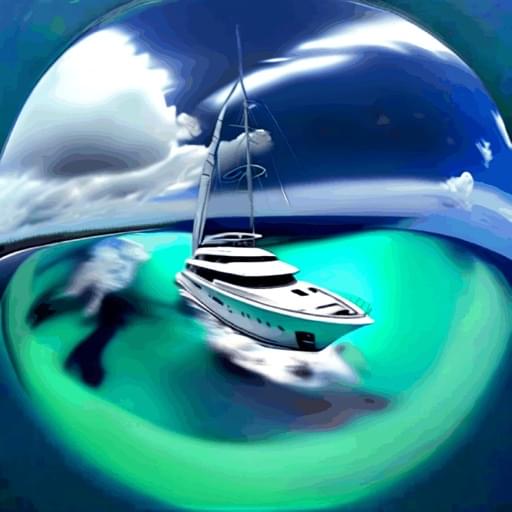} & \includegraphics[width=0.112\linewidth]{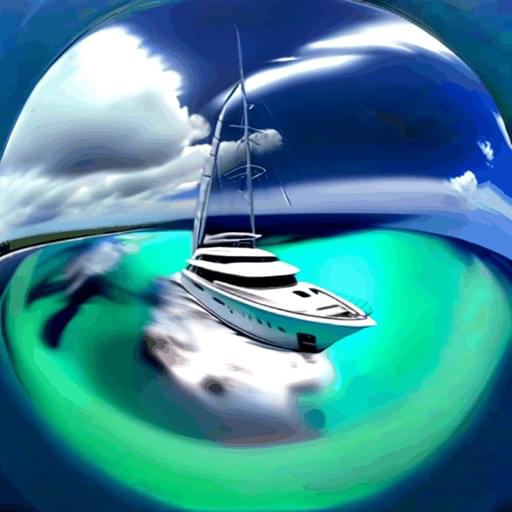} & \includegraphics[width=0.112\linewidth]{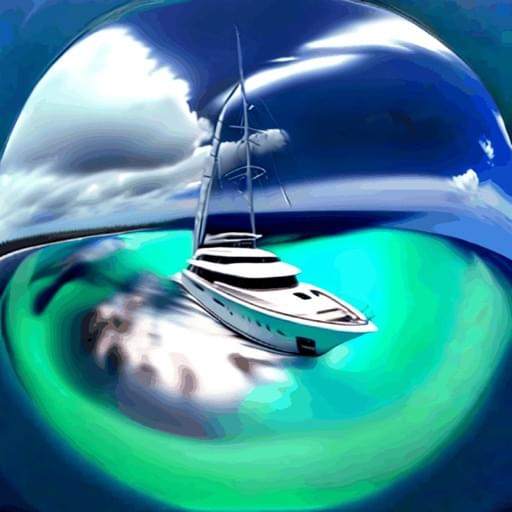} & \includegraphics[width=0.112\linewidth]{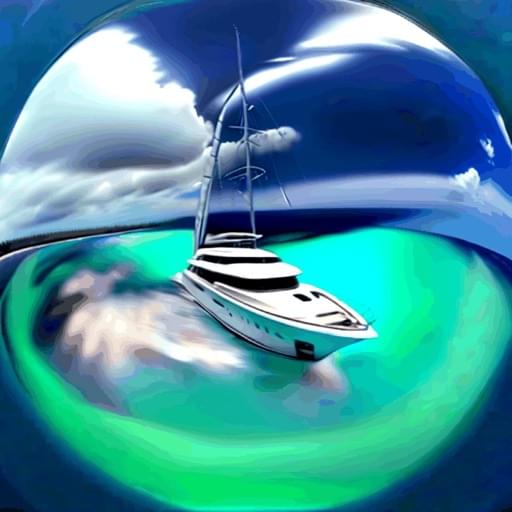} & \includegraphics[width=0.112\linewidth]{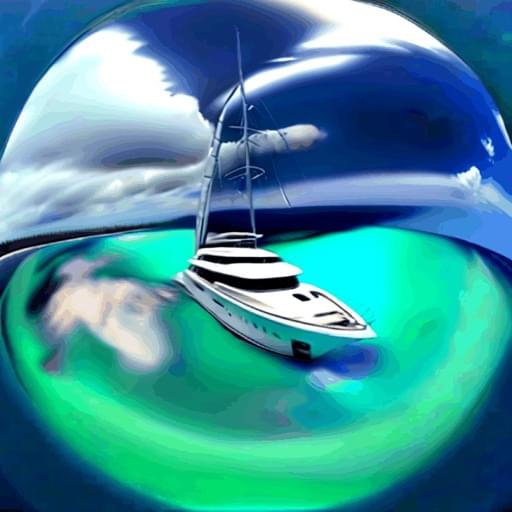} \\
\midrule
\rotatebox[origin=l]{90}{\makebox[.1\linewidth]{w/o constraint}} & \rotatebox[origin=l]{90}{\makebox[.115\linewidth]{\scriptsize 0--48 frames}} & \includegraphics[width=0.112\linewidth]{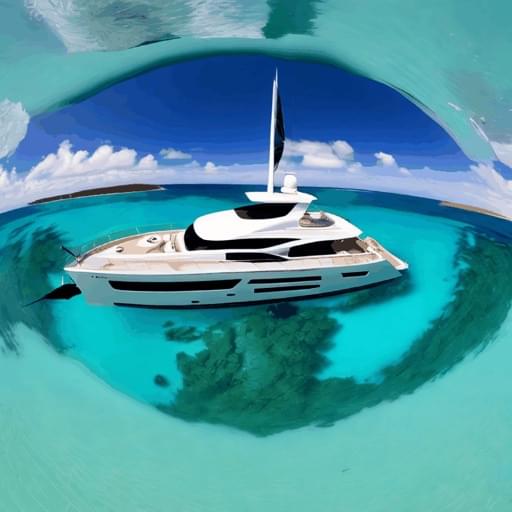} & \includegraphics[width=0.112\linewidth]{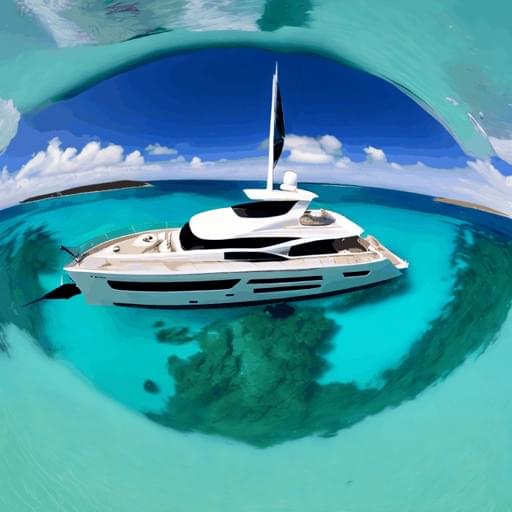} & \includegraphics[width=0.112\linewidth]{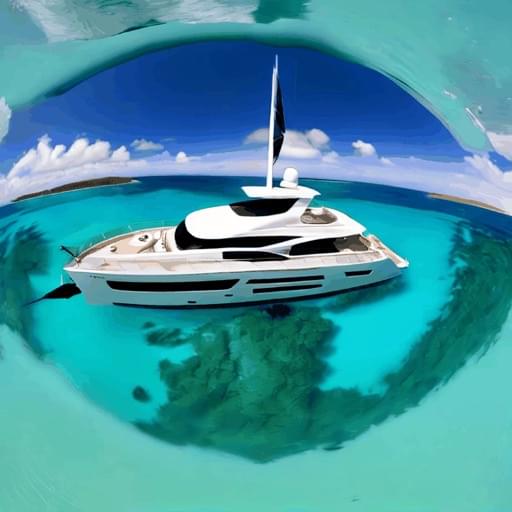} & \includegraphics[width=0.112\linewidth]{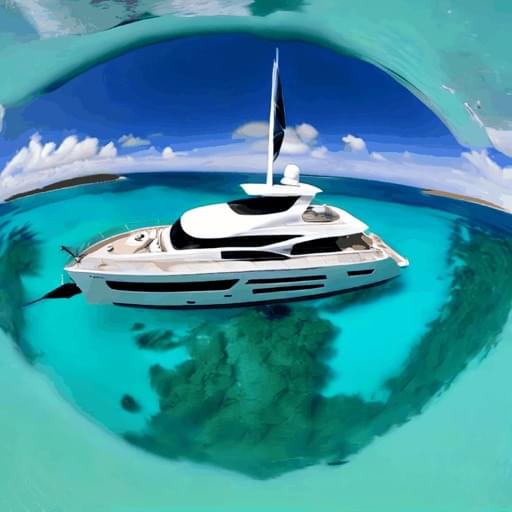} & \includegraphics[width=0.112\linewidth]{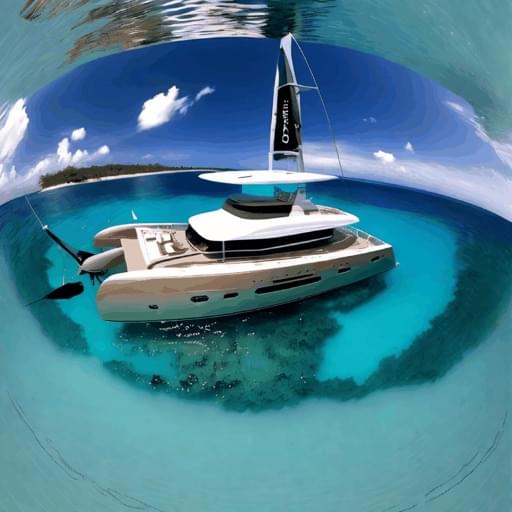} & \includegraphics[width=0.112\linewidth]{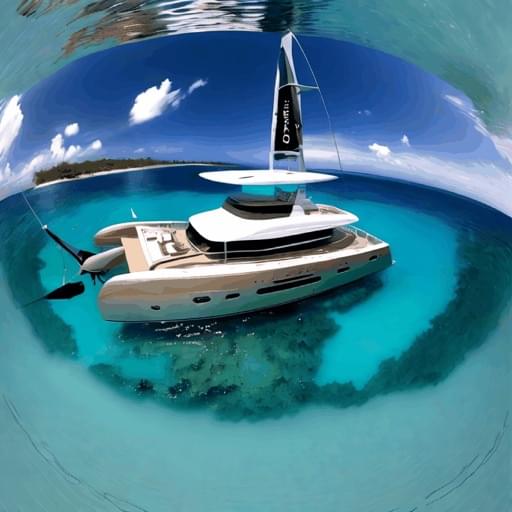} & \includegraphics[width=0.112\linewidth]{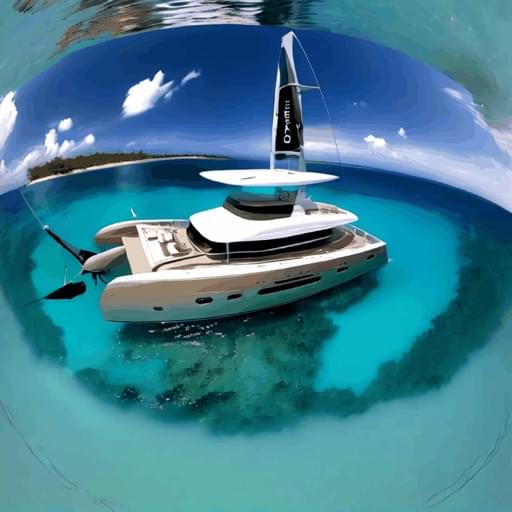} & \includegraphics[width=0.112\linewidth]{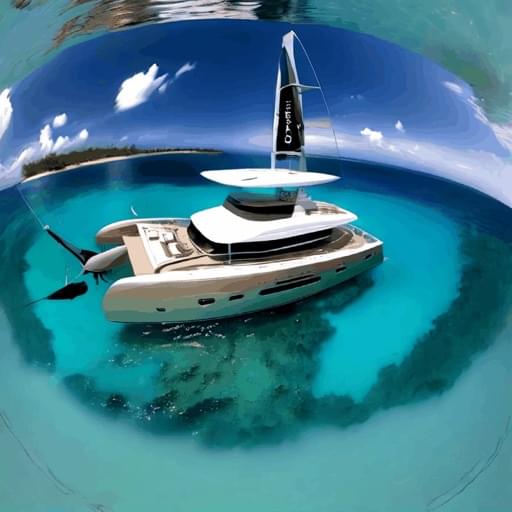} \\[-5pt]
\end{tabular}
\end{small}
\caption{Long video generation example with \textit{``a 360 shot of a sleek yacht sailing gracefully through the crystal-clear waters of the Caribbean''}. The top two rows use the noise constraint in~\cref{eq:inverse} to improve temporal consistency, while the bottom row does not.}
\label{fig:long}
\end{center}
\vskip -0.15in
\end{figure*}

The text-to-video and image-to-video generation results are visualized in~\cref{fig:t2v_vis}. For text-to-video generation, our method can produce visual quality not much far from the closed-source model Gen-2~\cite{runaway2023gen}, thanks to the unified pre-training framework with images. Meanwhile, \ours~is advantageous in reasoning abilities, such as inferring better motion (in the top-left example) and adding artistic touches based on the text prompt (as in both cases).
For image-to-video generation, our method is comparable to the state-of-the-art model SVD~\cite{blattmann2023stable} in generating both coherent and highly aesthetic video clips (the bottom-left example). In addition, the decomposed video representation enables the video decoder to produce more salient and vivid movements given a relatively difficult synthetic image prompt (the bottom-right example).

Furthermore, our autoregressive model can be naturally extended to long video generation, as shown in~\cref{fig:long}. Thanks to the proposed explicit noise constraint when decoding consecutive video clips, the temporal consistency between decoded clips is greatly improved. In contrast, decoding each video clip separately will result in the incoherence of fine-grained visual details among the video frames of different clips (See the bottom of ~\cref{fig:long}).

\subsection{Ablation Study}
This section investigates the impact of motion tokenization and different motion token lengths. Due to limited space, the ablation for proposed enhanced motion conditioning strategy is provided in~\cref{app:gen}.

\textbf{Effect of Motion Tokenization}.
We design two baselines to validate the effectiveness of motion tokenization in video pre-training. For video understanding, the w/o motion in~\cref{tab:ablation1} indicates the independent tokenization of 16 uniformly sampled video frames by the 2D visual encoder without any motion tokens. Most existing methods use a similar strategy to encode video content fed into the LLM. As observed in~\cref{tab:ablation1}, the question-answering accuracy decreases without explicitly modeling temporal information. For the video synthesis, the w/o motion baseline divides the text-to-video process into two separate stages: text-to-image and image-to-video. Specifically, given a textual prompt, we generate only a keyframe (without producing motion tokens) and then feed this keyframe into the image-to-video generation model svd-img2vid-xt~\cite{blattmann2023stable} to synthesize the final video. Since svd-img2vid-xt takes only an image as condition, the video generation process of this baseline lacks temporal motion as guidance. In comparison, our model can generate text-related motion tokens and thus generate more accurate video content following the prompt. 

\begin{table}[t]
\vskip -0.1in
\caption{Zero-shot text-to-long video generation performance. It is evaluated on 2048 long videos (64 frames) generated using the prompts from EvalCrafter~\cite{liu2023evalcrafter}.}
\label{tab:long_video_quant}
\begin{center}
\begin{small}
\setlength{\tabcolsep}{6.5pt}
\resizebox{1.0\linewidth}{!}{
\begin{tabular}{lccc}
\toprule
Method & FVD ($\downarrow$) & KVD ($\downarrow$) & CLIPSIM ($\uparrow$) \\
\midrule
Direct &  737.61 &  359.11  &  0.9104 \\
Sliding &  224.55 &  44.09  &  0.9438 \\
Gen-L-Video~(\citeauthor{wang2023gen}) &  177.63	& 21.06	& 0.9370 \\
FreeNoise~(\citeauthor{qiu2023freenoise}) &  \textbf{85.83}  &	6.07   & \textbf{0.9732} \\
\cmidrule{1-4} \ours &  113.37  & \textbf{4.94}  & 0.9621 \\
\bottomrule
\end{tabular}
}
\end{small}
\end{center}
\vskip -0.3in
\end{table}

\textbf{Effect of Token Length}.
We also explore the influence of different motion token lengths when encoding temporal motion information. The detailed results are reported in~\cref{tab:ablation2}. It can be observed that a very small number suffice to yield high understanding and generation performance. More token numbers may lead to representation redundancy and bring more duplicate motion token IDs when encoding videos without obvious motions, rendering the next-token prediction learning paradigm of LLM less effective. Using fewer motion tokens also allows for more video clips as input conditions under the same context length of LLM, which is useful for long video understanding.

\section{Conclusion}
This paper introduces \ours, a multimodal generative pre-training method that empowers LLMs with unified comprehension and generation of videos, images, and language. At the core of our method is a video decomposition scheme that allows for more effective modeling of temporal information while reusing visual knowledge from image-only multimodal LLMs. The decomposed keyframes and motion vectors can be efficiently tokenized to be adapted to LLMs for unified generative pre-training. Finally, the understanding and generative capabilities of \ours\\are verified by extensive quantitative and qualitative results.

\begin{table}[t]
\vskip -0.1in
\caption{Ablation of proposed motion tokenization strategy in zero-shot video understanding (left) and generation (right).}
\label{tab:ablation1}
\begin{center}
\begin{small}
\setlength{\tabcolsep}{6.5pt}
\begin{tabular}{lcccc}
\toprule
\multirow{2}{*}{Method} & MSVD & ActivityNet & \multicolumn{2}{c}{UCF-101} \\
\cmidrule(lr){2-2} \cmidrule(lr){3-3} \cmidrule(lr){4-5} & Accuracy & Accuracy & IS ($\uparrow$) & FVD ($\downarrow$) \\
\midrule
w/o motion & 67.3  & 47.4  & 29.56 & 442.80 \\
w/ motion & \textbf{73.2} & \textbf{50.1} & \textbf{44.26} &  \textbf{280.57}\\
\bottomrule
\end{tabular}
\end{small}
\end{center}
\vskip -0.2in
\end{table}


\section*{Impact Statement}

While this work advances the pre-training of large multimodal models in both performance and efficiency, its reasoning and generative capabilities should be treated carefully. Some well-known problems include hallucination in multimodal understanding and exploitation to create misinformation through personalized generation. The model could also produce harmful responses due to inherent data bias and lack of alignment procedures.

One positive aspect we'd like to highlight is that, unlike many generation methods compared, this model is fully trained with \textit{public} datasets. This allows the research community to correct for bias and harmful content in the training data. We hope that such a practice will help address important safety and alignment issues in multimodal models.

\begin{table}[t]
\vskip -0.1in
\caption{Ablation of the number of motion tokens (denoted by $N$) in zero-shot video understanding (left) and generation (right).}
\label{tab:ablation2}
\begin{center}
\begin{small}
\setlength{\tabcolsep}{5.7pt}
\begin{tabular}{lcccc}
\toprule
\multirow{2}{*}{Method} & MSVD & ActivityNet & \multicolumn{2}{c}{UCF-101} \\
\cmidrule(lr){2-2} \cmidrule(lr){3-3} \cmidrule(lr){4-5} & Accuracy & Accuracy & IS ($\uparrow$) & FVD ($\downarrow$) \\
\midrule
$N=256$  & 69.2  & 48.8 & 37.57 & 281.24  \\
$N=135$ & \textbf{73.2} & \textbf{50.1} &  \textbf{44.26} &  \textbf{280.57} \\
\bottomrule
\end{tabular}
\end{small}
\end{center}
\vskip -0.2in
\end{table}

\section*{Acknowledgements}

This research work is supported by National Key R\&D Program of China (2022ZD0160305), a research grant from China Tower Corporation Limited, and a grant from Beijing Aerospace Automatic Control Institute. We also sincerely thank for the very constructive comments from all reviewers.


\bibliography{bibs/video, bibs/image, bibs/LLM}
\bibliographystyle{icml2024}


\newpage
\appendix
\onecolumn
\section{Experimental Settings}

\subsection{Model Implementation Details}
\label{app:model}


\textbf{Video Tokenizer} 
We employ the off-the-shelf visual tokenizer from LaVIT~\cite{jin2024unified} to transform the video keyframe into 90 visual tokens on average, which follows most existing MLLMs to utilize the ViT-G/14 of EVA-CLIP~\citep{fang2023eva} as the visual encoder. The visual codebook size is set to 16384. Please refer to the original paper for more details. During training and inference, images and keyframes are resized to 224$\times$224 resolution as input. 

As for motion tokenization, we downsample the original videos at 6 fps and then take 24 consecutive frames as a video clip to compute the motion vector $M$. It is further divided by the width and height of the corresponding video to normalize the value within the range of $[-1, 1]$. Before feeding into the motion tokenizer, the motion vector $M$ is resized to a resolution of 20$\times$36, resulting in the final input tensor shape being $B \times 24 \times 20 \times 36 \times 2$. The encoder $f_{\mathcal{E}}$ and decoder $f_{\mathcal{D}}$ in our motion tokenizer both have $L=12$ transformer blocks with 512 hidden states and 8 attention heads. Each block consists of spatial, temporal attention, and feed-forward layers. Before the attention computation, the motion input is reshaped into $[(B T) \times (H W) \times D]$ and $[(B H W) \times T \times D]$ for the spatial and temporal layers, respectively. We insert the spatial or temporal downsampling layers after the $[3, 6, 9, 12]$ encoder blocks to reduce the dimension of motion embeddings, which will then be quantized into 135 ($3 \times 9 \times 5$) discrete motion tokens. The decoder $f_{\mathcal{D}}$ includes symmetrical upsampling layers to recover the original input motion vector during training. The size of learned motion codebook is set to 1024. To improve the training stability of the motion codebook, we leverage exponential moving average (EMA) updates with a weight of $0.995$. Before quantization, the motion embeddings are projected into a low-dimensional space (dim=32) to improve the codebook usage, following the experience of Yu et al. (\citeyear{yu2022vector}). 


\textbf{Video Detokenizer} During training of the video detokenizer, we randomly sampled 24 consecutive frames from videos downsampled at 6 fps. The motion conditioning encoder has the same transformer architecture (12 blocks) as $f_{\mathcal{E}}$, except that the downsample layers are removed to keep the same temporal dimension with the input video frames. This strategy reduces the compression of motion information during encoding and provides explicit guidance for each frame to be denoised in the 3D U-Net. The detailed architecture of the 3D U-Net employed follows the same implementations as Blattmann et al. (\citeyear{blattmann2023align,blattmann2023stable}). During the EDM-preconditioning optimization for the detokenizer, the distribution of $\log \sigma$ is set to $\mathcal{N}(1.0, 1.2^2)$ to encourage a higher noise level, which is found effective for the high-resolution generation~\cite{girdhar2023emu}. We train the motion conditioning encoder, the input encoding layer, and all the cross-attention layers in the 3D U-Net from scratch and initialize the other weights from the SVD img2vid-xt~\cite{blattmann2023stable}. To reduce the computational complexity, the detokenizer is first trained with a resolution 384 $\times$ 384 for 50k steps, and then further fine-tuned at two types of resolutions: 768 $\times$ 768 or 1024 $\times$ 576 for another 10k steps.

\textbf{Language Model} We utilize Llama 2 7B~\citep{touvron2023llama2} as the default large language model for the generative pre-training. The weight of the language model is initialized from LaVIT~\cite{jin2024unified} to preserve the learned visual prior knowledge to support the comprehension and generation for the image domain. During pre-training, we mix the image-text, video-text pairs, and textual data in one batch to form the final multimodal input sequence.

\subsection{Pre-training Data}
The training dataset used by \ours~only consists of publicly available image and video datasets. In the following, we present a detailed elaboration of the dataset usage at each training stage.

\textbf{Stage 1:} The video tokenizer and detokenizer are trained on the WebVid-10M~\cite{bain2021frozen}, which is an open-source video-text dataset containing 10 million video-text pairs scraped from the stock footage sites. Since both our tokenizer and detokenizer do not rely on textual data, we only employ pure video data at this stage. Due to the common watermarks in WebVid-10M, during the training of the video detokenizer, we incorporate a subset of InterVid-14M-aesthetics~\cite{wang2024internvid} to remove watermarks in the generated videos. It has also been shown useful in PixelDance~\cite{zeng2023make}. Specifically, we first select a subset of 4s--10s video clips with the highest aesthetic scores and then follow SVD~\cite{blattmann2023stable} in applying CRAFT~\cite{baek2019character} to filter out those videos with unwanted written text. The result contains about 300k publicly available video clips. \textbf{Noting that the 300k video subset is only used during the training of the video detokenizer to improve the aesthetics of the generated videos, the results reported in all the experiments are tested on the checkpoint that uses only the WebVid-10M dataset.}

\textbf{Stage 2:} The language model is pre-trained on a mixture of video, image and text data, including WebVid-10M~\cite{bain2021frozen}; 93M samples from Conceptual Caption~\cite{sharma2018conceptual, changpinyo2021conceptual}, SBU~\cite{ordonez2011im2text}, and BLIP-Capfilt~\cite{li2022blip}. Moreover, we also employ the English text corpus from RedPajama~\citep{together2023redpajama}, which is open-source data like the original one to train LLaMA from scratch. The purpose of including the English text corpus during pre-training is to preserve the already learned language understanding ability of LLM (e.g., the performance on linguistic benchmarks like MMLU~\cite{hendrycks2021measuring}) while acquiring good multimodal capabilities. 

\textbf{Stage 3:} For a fair comparison, we employ the same instruction tuning dataset as the existing works~\cite{lin2023video, li2023llama} during this stage. It includes a 665k image-text instruction dataset from LLaVA v1.5~\cite{liu2023improved} and a 100k video-text instruction dataset from Video-ChatGPT~\cite{maaz2023video}. All the understanding results are tested by the model trained after Stage 3.

\subsection{Training Settings}
\label{sec:supp_detail}

The detailed training hyper-parameter settings for the video tokenizer, detokenizer, and language model in \ours~are reported in Table~\ref{tab:hyper}. We adopt the same instruction tuning setting as LLaVA v1.5~\cite{liu2023improved}.

\begin{table*}[h]
    \centering
    \resizebox{0.80\linewidth}{!}{
        \begin{tabular}{lccc}
        \toprule
        Configuration & Language Model & Tokenizer & Detokenizer \\ 
        \midrule
        LLM init  & LaVIT-7B & - & -\\
        Optimizer  & AdamW & AdamW & AdamW \\
        Optimizer Hyperparameters & $\beta_1=0.9$, $\beta_2=0.95$, $\epsilon=1e^{-6}$ & \multicolumn{2}{c}{$\beta_1=0.9$, $\beta_2=0.99$, $\epsilon=1e^{-6}$} \\
        Global batch size & 2048 & 512 & 128 \\
        Peak learning rate of LLM & 2e-5 & - & - \\
        Peak learning rate of other Part & 5e-5 & 1e-4 & 5e-5\\
        Learning rate schedule &  Cosine & Cosine & Cosine \\
        Training Steps & 30K & 100K & 60K \\
        Warm-up steps & 2k & 5K  & 3K \\
        Weight decay & 0.1 & 0.001 & 0.001 \\
        Gradient clipping & 1.0 & 1.0 & 1.0\\
        Input sequence to LLM & 2048 & - & -\\
        Numerical precision & bfloat16 & bfloat16 & bfloat16 \\
        GPU Usage & 128 NVIDIA A100 & 64 NVIDIA A100 & 64 NVIDIA A100\\
        Framework & Megatron & DeepSpeed & DeepSpeed \\
        Training Time & 60h & 10h & 48h \\
        \bottomrule
        \end{tabular}
    }
    \caption{The detailed training hyperparameters of \ours}
    \label{tab:hyper}
    \vspace{-0.1in}
\end{table*}

\subsection{Evaluation}
\label{app:eval}

\textbf{Image Understanding} is evaluated using eight popular image question answering and multimodal benchmarks: VQA v2~\cite{goyal2017making}, GQA~\cite{hudson2019gqa}, VizWiz~\cite{gurari2018vizwiz}, ScienceQA-IMG~\cite{lu2022learn}, MME~\cite{fu2023mme}, MMBench~\cite{liu2023mmbench}, SEED~\cite{li2023seed}, MM-Vet~\cite{yu2023mm}. For question-answering datasets, we use the same prompts as in LLaVA-1.5~\cite{liu2023improved}, and adopt the widely used VQA accuracy as the evaluation metric.

\textbf{Video Question Answering}. Three common datasets are considered: MSVD-QA~\cite{chen2011collecting}, MSRVTT-QA~\cite{xu2016msr} and ActivityNet-QA~\cite{yu2019activitynet}. To assess model accuracy, a GPT-3.5 assistant~\cite{maaz2023video} is employed, which also produces outputs a relative score ranging from 0 to 5.

\textbf{Text-to-Image Generation}. We adopt the validation set of MS-COCO~\cite{lin2014microsoft} and randomly select 30K samples. The quality of the generated images is evaluated by Fr{\'e}chet Inception distance (FID)~\cite{heusel2017gans}, which computes its Fr{\'e}chet distance to the ground truth in the feature space of a pre-trained Inception V3 model.

\textbf{Text-to-Video Generation} is measured on MSR-VTT~\cite{xu2016msr} and UCF-101~\cite{soomro2012dataset}. For MSR-VTT, we use all 2990 videos and sample one caption for each video, resulting in 2990 video-text pairs; for UCF-101, we sample 20 videos per class and follow PYoCo~\cite{ge2023preserve} to curate prompts for each class, producing 2020 video-text pairs. Their evaluation metrics are detailed below.
\begin{itemize}
    \item CLIP similarity (CLIPSIM)~\cite{wu2021godiva} measures the semantic similarity between video-text pairs. We follow Phenaki~\cite{villegas2023phenaki} and VideoPoet~\cite{kondratyuk2023videopoet} in using a ViT-B/16~\cite{radford2021learning} to compute CLIP scores between 224$\times$224 sized video frames and their corresponding captions. The final score is averaged over all generated video frames.
    \item Fr{\'e}chet video distance (FVD)~\cite{unterthiner2018towards} evaluates the Fr{\'e}chet distance between generated and real videos in the feature space of an I3D action classification model~\cite{carreira2017quo} pre-trained on Kinetics-400~\cite{kay2017kinetics}.
    \item Fr{\'e}chet Inception distance (FID)~\cite{heusel2017gans} measures the Fr{\'e}chet distance between generated and real video frames. Following PYoCo~\cite{ge2023preserve}, we use a ViT-B/32 model~\cite{kynkaanniemi2023role} to extract the frame features. The final result is averaged over all video frames.
    \item Inception score (IS)~\cite{saito2020train} evaluates the distribution of our generated video frames. We employ a C3D model~\cite{tran2015learning} fine-tuned on UCF-101 to calculate a video version of the inception score. The model takes the central 16 frames of each video as the input.
\end{itemize}
Note that there are slight variations in the evaluation protocols of different papers. We have sought to keep our protocol the same as or similar to most of the top-ranked methods.

\section{Additional Results}

\subsection{Multimodal Generation}
\label{app:gen}

\begin{figure}[t]
\begin{center}
\begin{small}
\begin{tabular}{lc@{}c@{}cc@{}c@{}c}
& \multicolumn{3}{c}{\it Close up headshot, futuristic young woman, wild hair sly} & \multicolumn{3}{c}{\multirow{2}{*}{\it A steaming cup of coffee with mountains in the}} \\  
& \multicolumn{3}{c}{\it smile in front of gigantic UFO, dslr, sharp focus, dynamic} & \multicolumn{3}{c}{\multirow{2}{*}{\it background. Resting during road trip.}} \\
& \multicolumn{3}{c}{\it composition.} \\
\rotatebox[origin=l]{90}{\makebox[.145\linewidth]{SDXL}} & \includegraphics[width=0.145\linewidth]{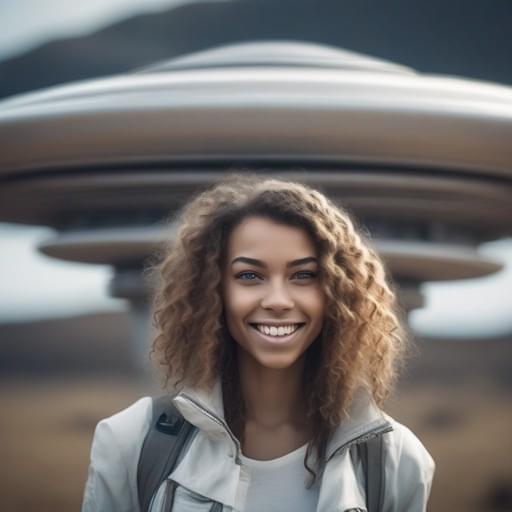} & \includegraphics[width=0.145\linewidth]{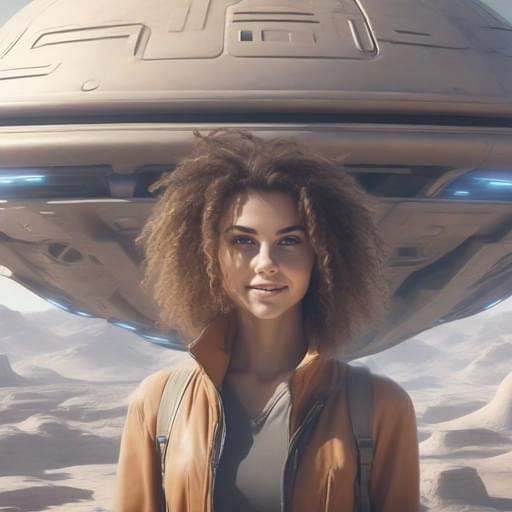} & \includegraphics[width=0.145\linewidth]{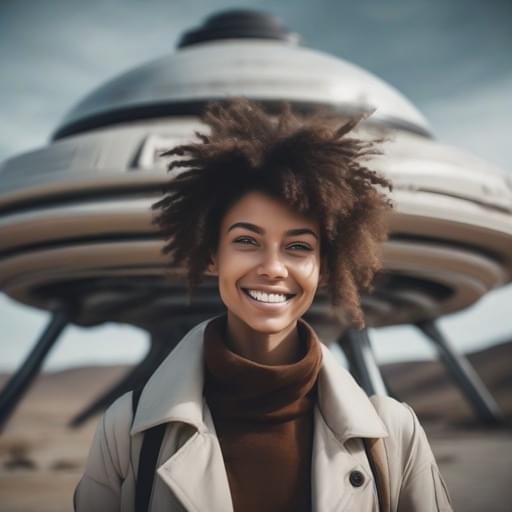} & \includegraphics[width=0.145\linewidth]{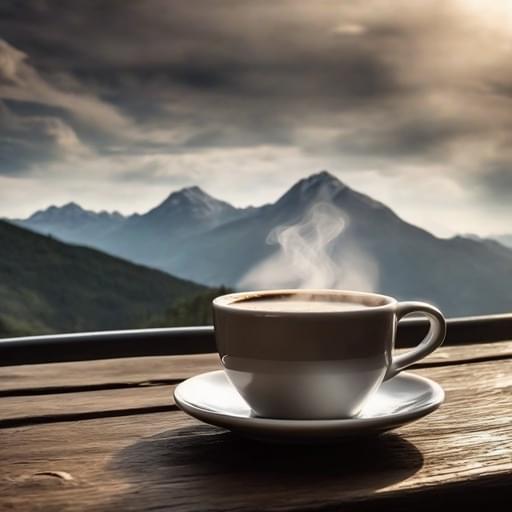} & \includegraphics[width=0.145\linewidth]{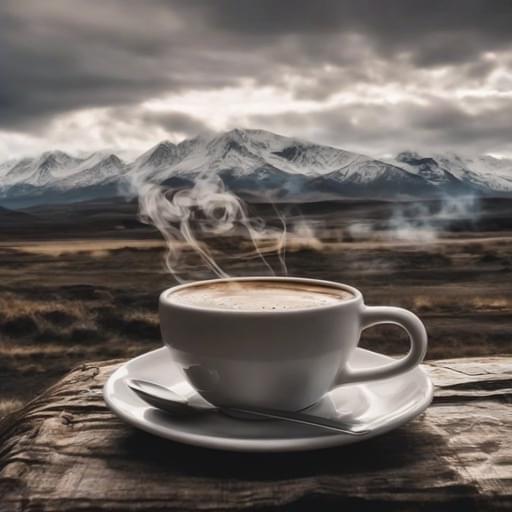} & \includegraphics[width=0.145\linewidth]{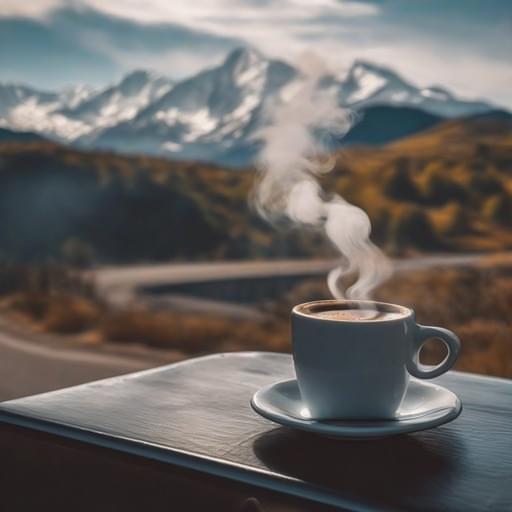} \\
\rotatebox[origin=l]{90}{\makebox[.135\linewidth]{\ours}} & \includegraphics[width=0.145\linewidth]{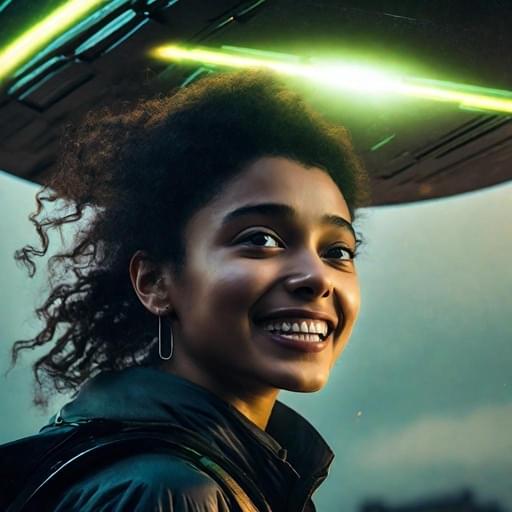} & \includegraphics[width=0.145\linewidth]{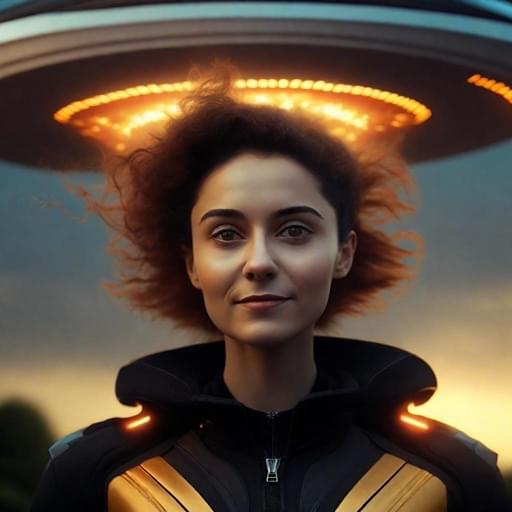} & \includegraphics[width=0.145\linewidth]{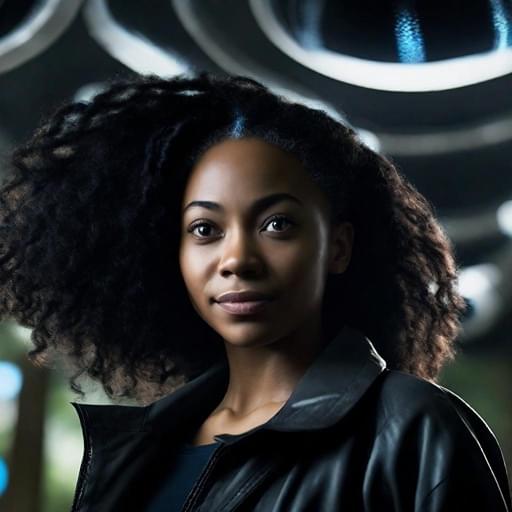} & \includegraphics[width=0.145\linewidth]{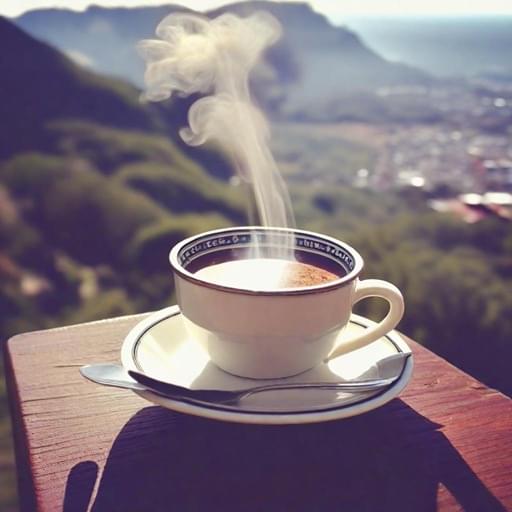} & \includegraphics[width=0.145\linewidth]{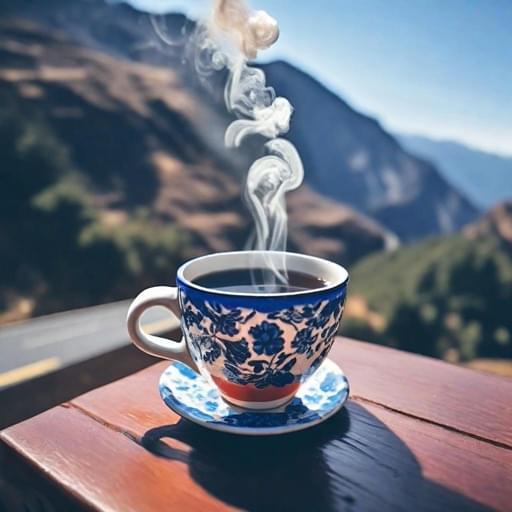} & \includegraphics[width=0.145\linewidth]{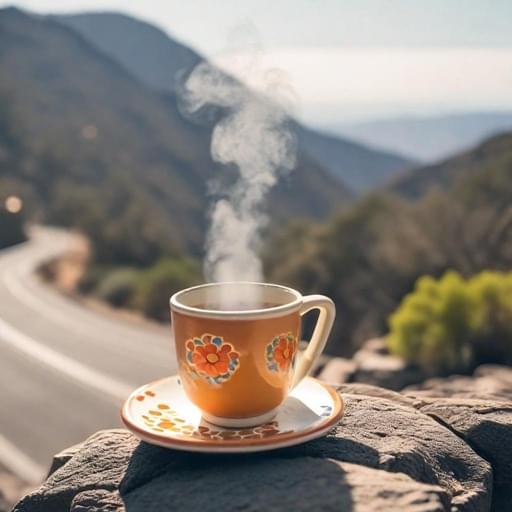} \\
\midrule
& \multicolumn{3}{c}{\it A squirrel is inside a giant bright shiny crystal ball on the} & \multicolumn{3}{c}{\multirow{2}{*}{\it An origami fox walking through the forest.}} \\  
& \multicolumn{3}{c}{\it surface of blue ocean. There are few clouds in the sky.} \\
\rotatebox[origin=l]{90}{\makebox[.145\linewidth]{SDXL}} & \includegraphics[width=0.145\linewidth]{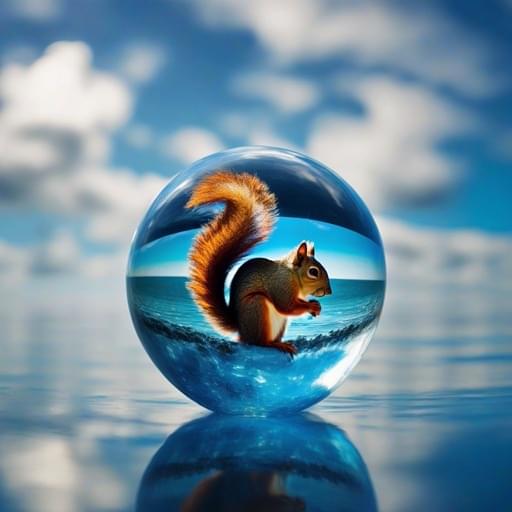} & \includegraphics[width=0.145\linewidth]{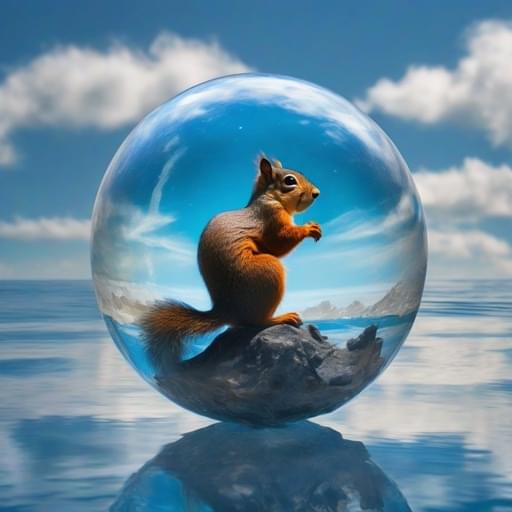} & \includegraphics[width=0.145\linewidth]{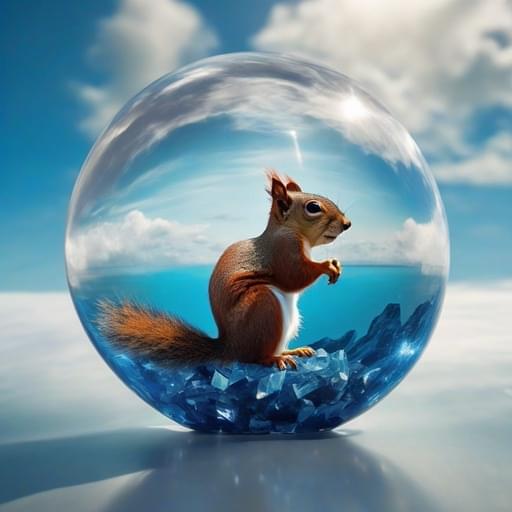} & \includegraphics[width=0.145\linewidth]{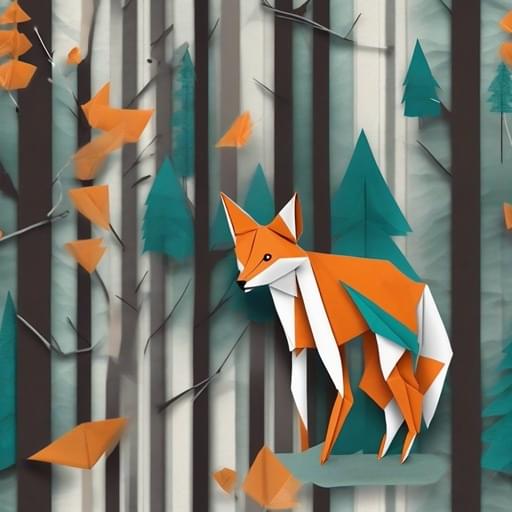} & \includegraphics[width=0.145\linewidth]{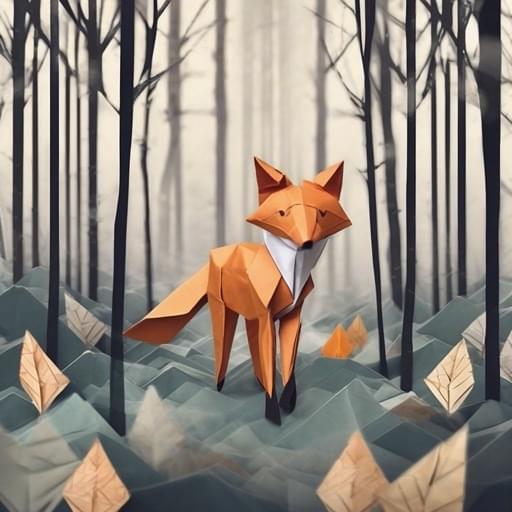} & \includegraphics[width=0.145\linewidth]{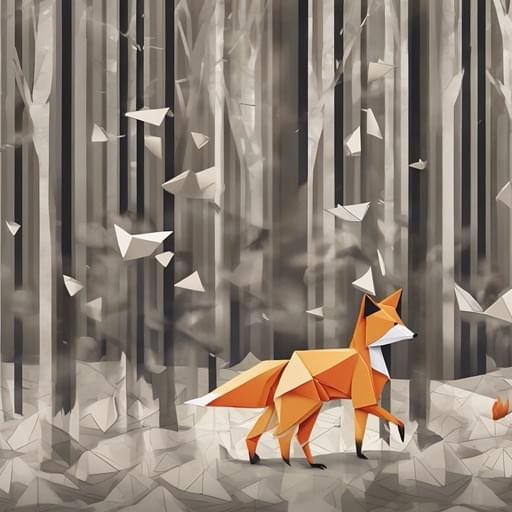} \\
\rotatebox[origin=l]{90}{\makebox[.135\linewidth]{\ours}} & \includegraphics[width=0.145\linewidth]{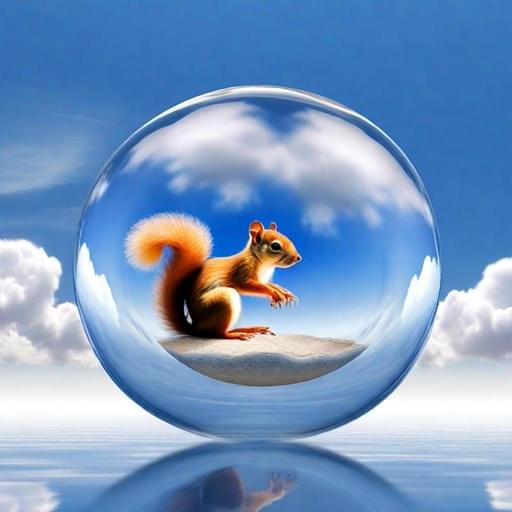} & \includegraphics[width=0.145\linewidth]{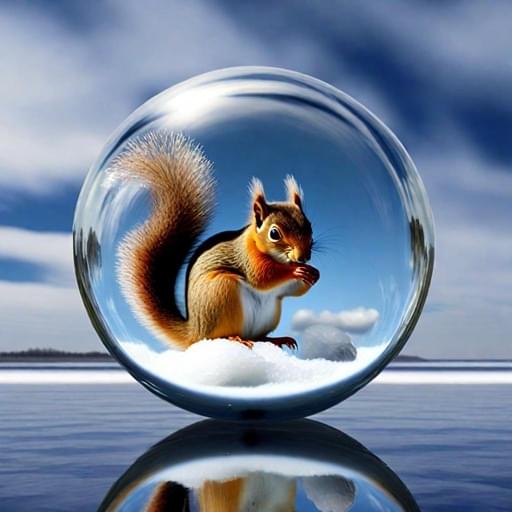} & \includegraphics[width=0.145\linewidth]{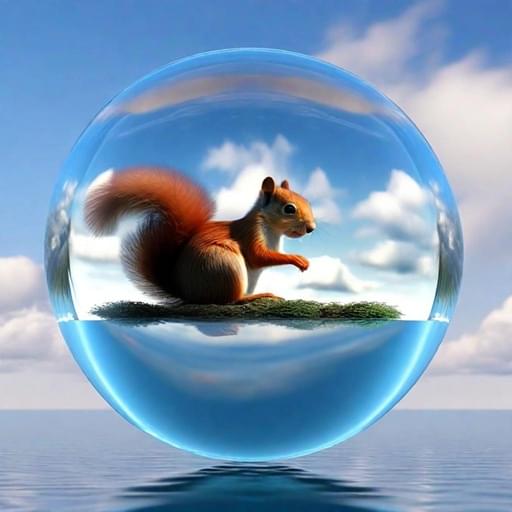} & \includegraphics[width=0.145\linewidth]{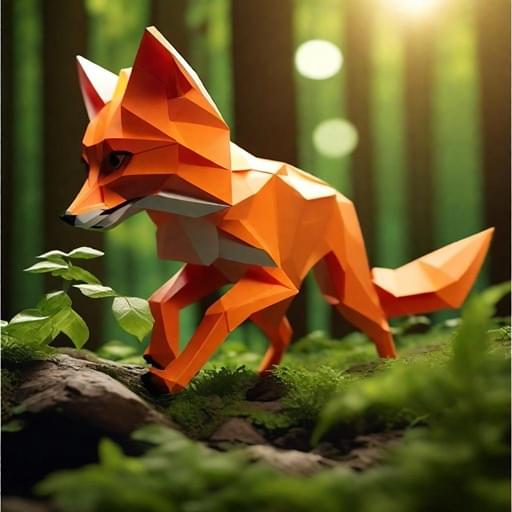} & \includegraphics[width=0.145\linewidth]{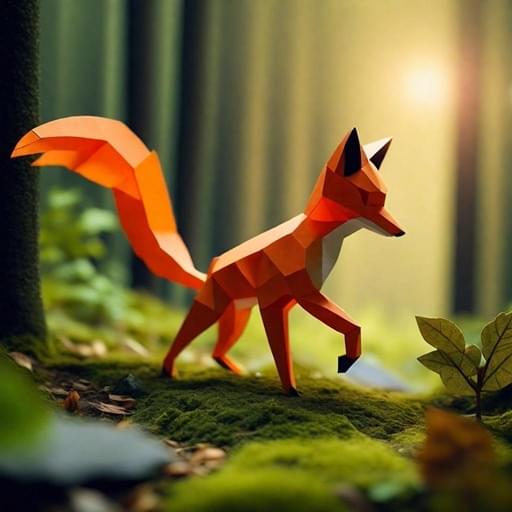} & \includegraphics[width=0.145\linewidth]{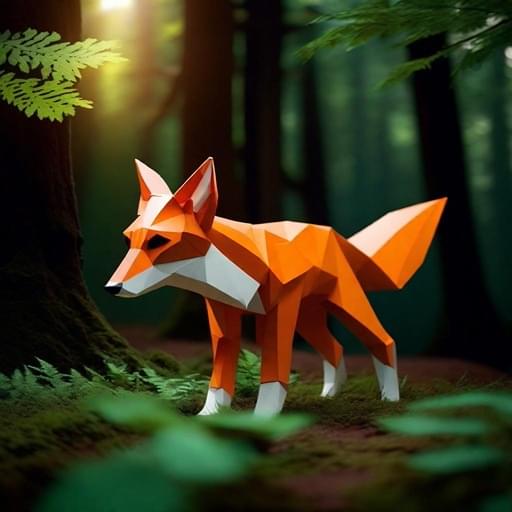} \\
\midrule
& \multicolumn{3}{c}{\it A watercolor painting of two apples on a wooden table,} & \multicolumn{3}{c}{\multirow{2}{*}{\it A cat is sitting on a basket under a bench.}} \\  
& \multicolumn{3}{c}{\it  neither is red and both are green.} \\
\rotatebox[origin=l]{90}{\makebox[.145\linewidth]{SDXL}} & \includegraphics[width=0.145\linewidth]{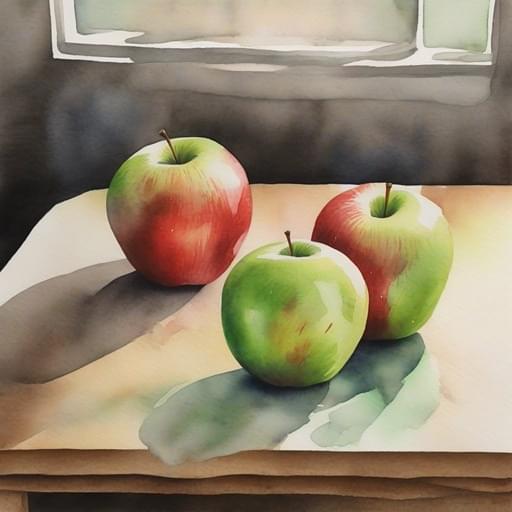} & \includegraphics[width=0.145\linewidth]{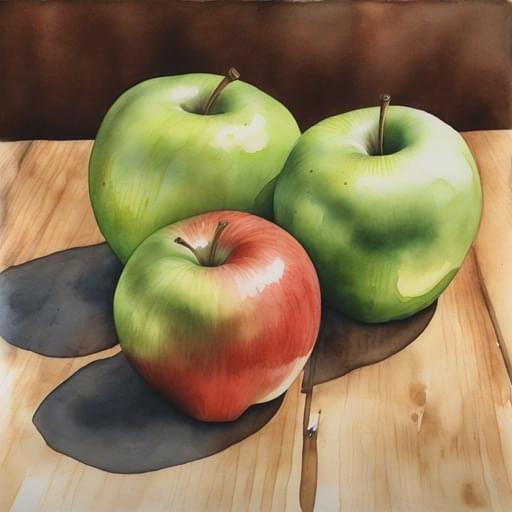} & \includegraphics[width=0.145\linewidth]{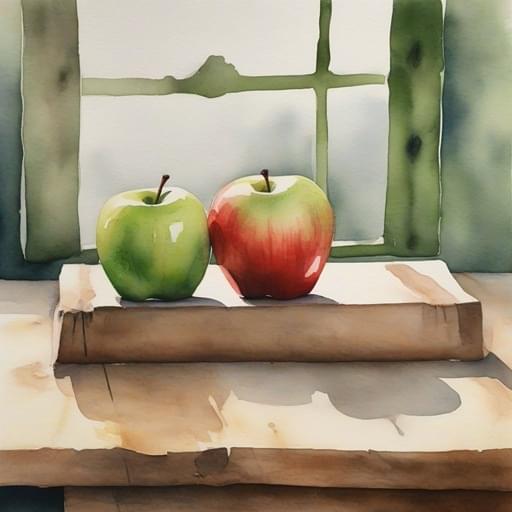} & \includegraphics[width=0.145\linewidth]{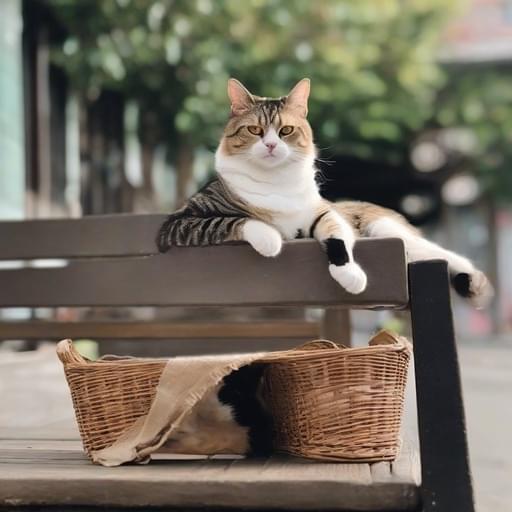} & \includegraphics[width=0.145\linewidth]{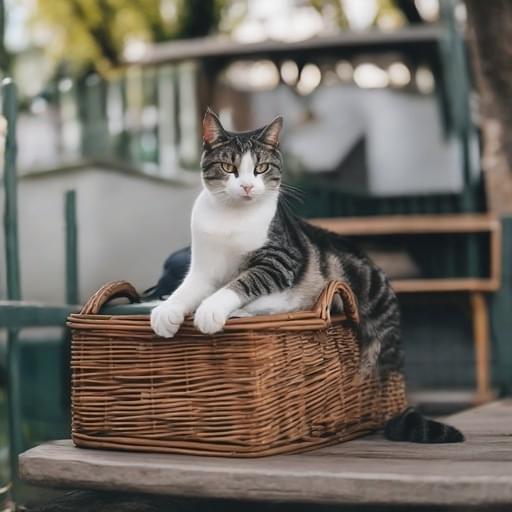} & \includegraphics[width=0.145\linewidth]{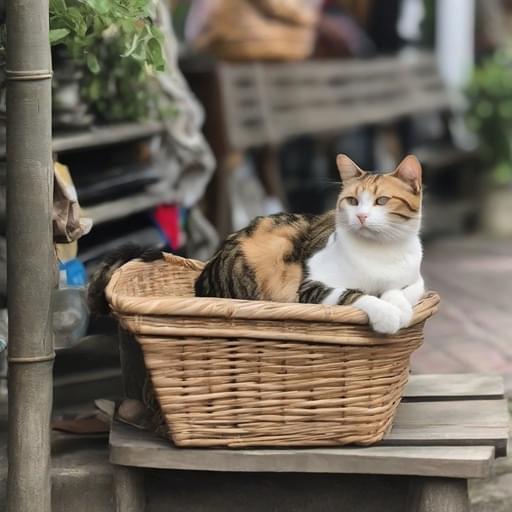} \\
\rotatebox[origin=l]{90}{\makebox[.135\linewidth]{\ours}} & \includegraphics[width=0.145\linewidth]{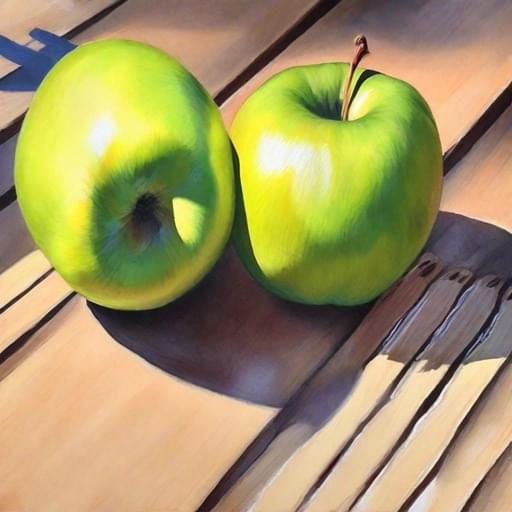} & \includegraphics[width=0.145\linewidth]{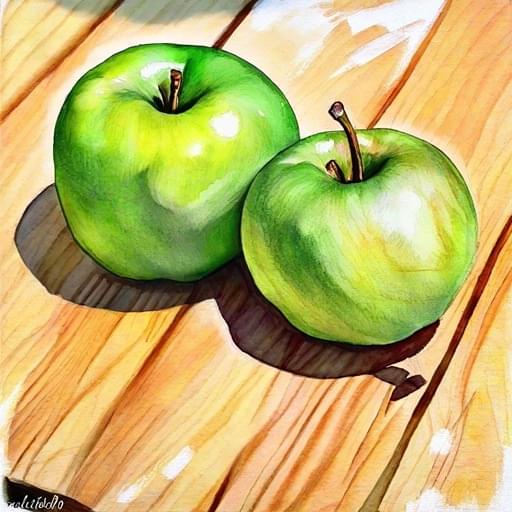} & \includegraphics[width=0.145\linewidth]{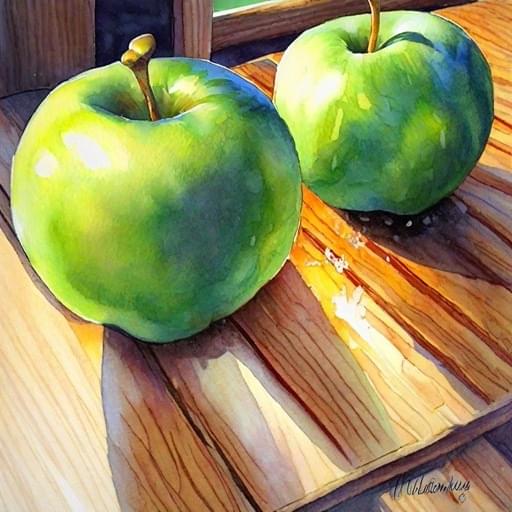} & \includegraphics[width=0.145\linewidth]{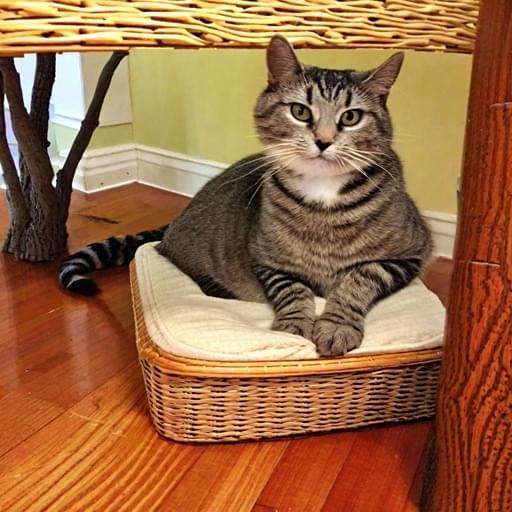} & \includegraphics[width=0.145\linewidth]{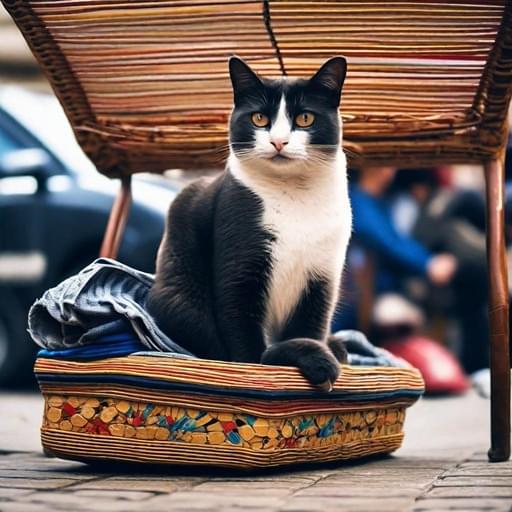} & \includegraphics[width=0.145\linewidth]{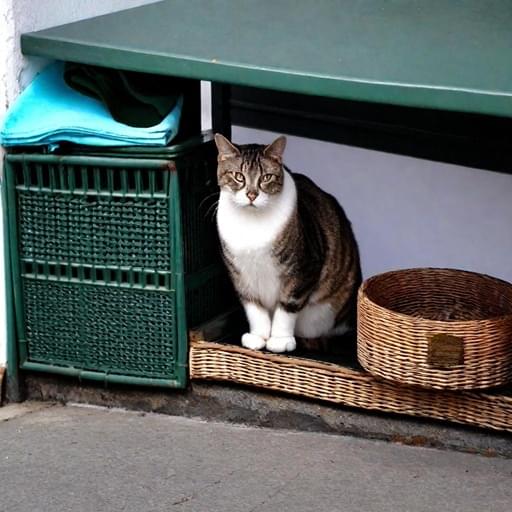} \\
\end{tabular}
\end{small}
\caption{Text-to-image generation comparison with SDXL~\cite{podell2024sdxl}. Prompts are from SDXL, CM3Leon~\cite{aghajanyan2022cm3}, Imagen~\cite{saharia2022photorealistic}, VideoPoet~\cite{kondratyuk2023videopoet}, LMD~\cite{lian2023llm}, and LayoutGPT~\cite{feng2024layoutgpt}. Our model provides comparable visual quality while showing better logical and spatial reasoning abilities (see the last two cases).}
\label{fig:t2i_vis2}
\end{center}
\vskip -0.1in
\end{figure}

This section provides additional qualitative results and an ablation study to demonstrate the effectiveness of our design for multimodal generation, complementing the existing comparisons in the main paper.

\textbf{Text-to-Image Generation}. \Cref{fig:t2i_vis2} illustrates the comparison of text-to-image generation between \ours~and SDXL~\cite{podell2024sdxl}. Overall, our method achieves competitive visual quality while having better language understanding and reasoning capabilities. For example, in the top-left case of a young woman in front of a UFO, our method produces highly aesthetic headshots of the woman, while capturing the detail of ``sharp focus'' in the text prompt. And in the bottom-left example of apple painting, our model successfully infers from the prompt ``neither is red and both are green'' to draw two green apples, thanks to the better logical reasoning ability of the LLM-based generation approach we adopted.


\begin{figure}[t]
\begin{center}
\begin{small}
\begin{tabular}{lc@{}c@{}c@{}c@{}c@{}c}
& \multicolumn{6}{c}{\it First-person view running through the woods and approaching a large beautiful cabin, highly detailed.} \\  
\rotatebox[origin=l]{90}{\makebox[.15\linewidth]{Gen-2}} & \includegraphics[width=0.15\linewidth]{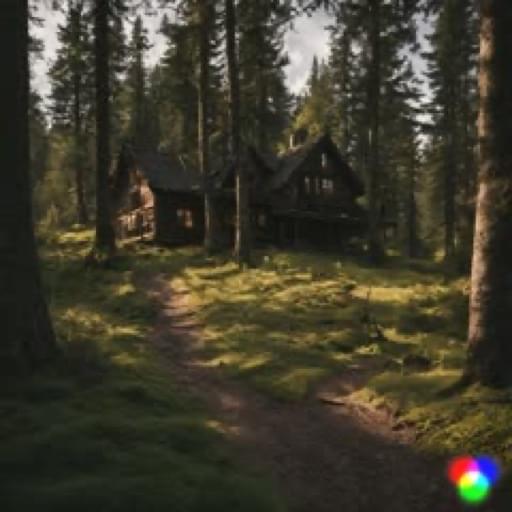} & \includegraphics[width=0.15\linewidth]{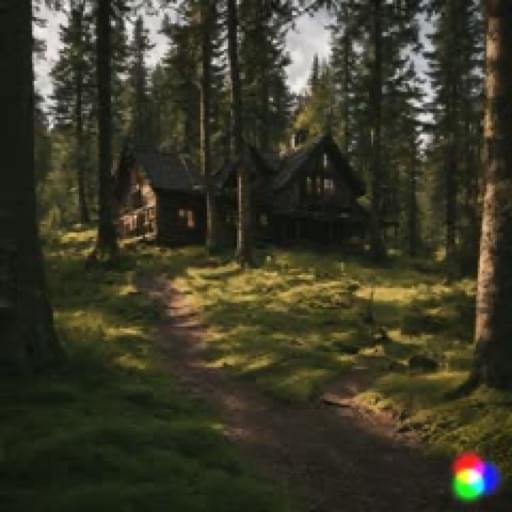} & \includegraphics[width=0.15\linewidth]{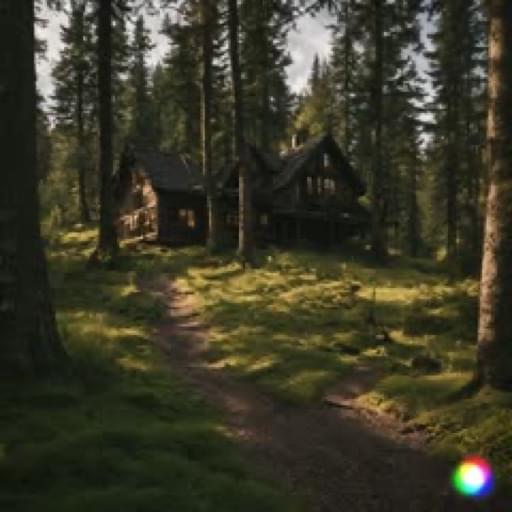} & \includegraphics[width=0.15\linewidth]{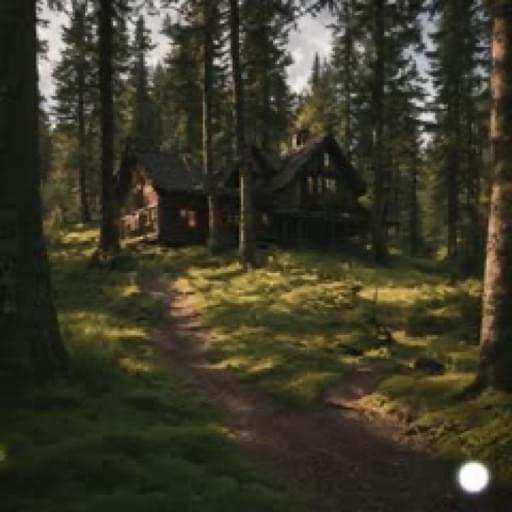} & \includegraphics[width=0.15\linewidth]{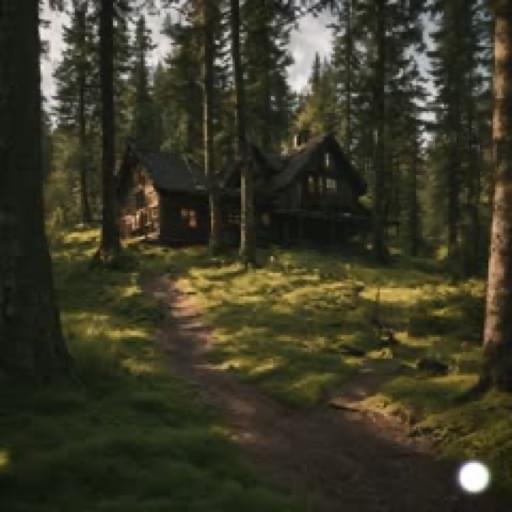} & \includegraphics[width=0.15\linewidth]{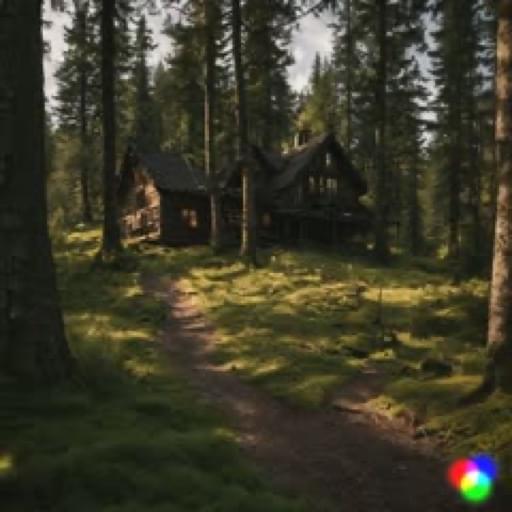} \\
\rotatebox[origin=l]{90}{\makebox[.14\linewidth]{\ours}} & \includegraphics[width=0.15\linewidth]{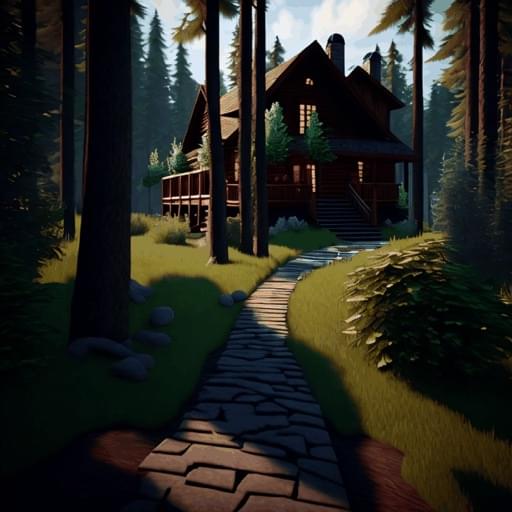} & \includegraphics[width=0.15\linewidth]{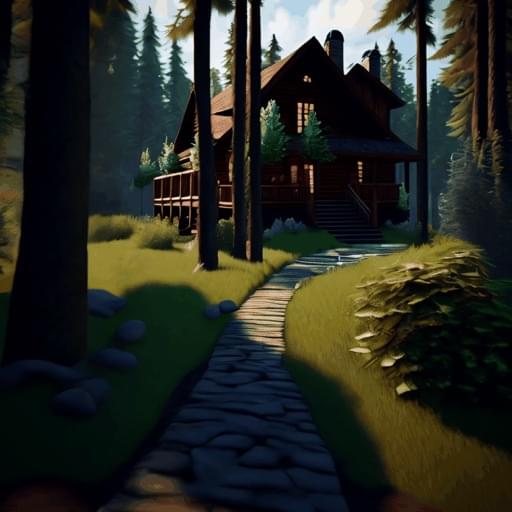} & \includegraphics[width=0.15\linewidth]{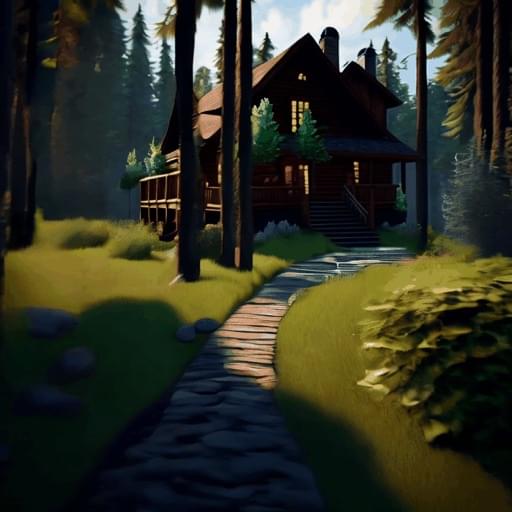} & \includegraphics[width=0.15\linewidth]{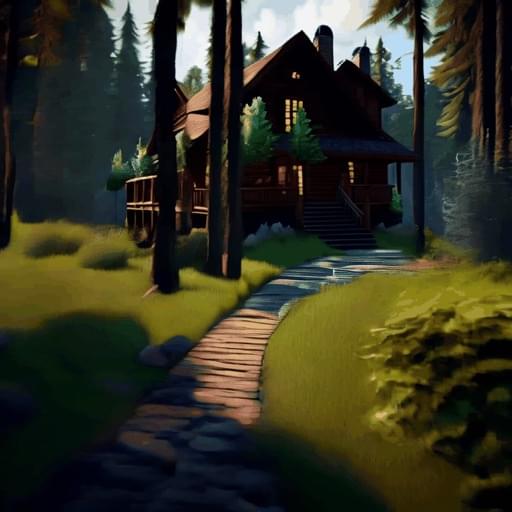} & \includegraphics[width=0.15\linewidth]{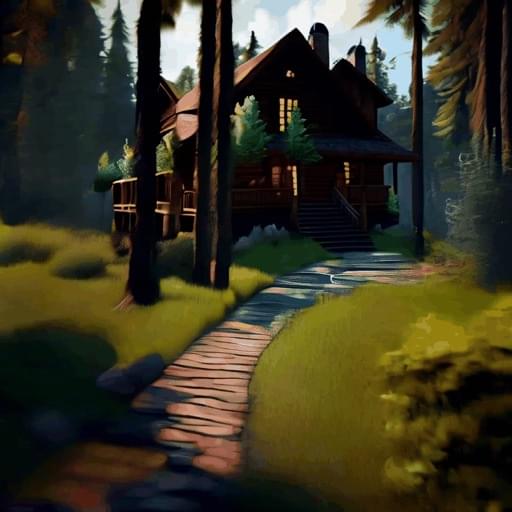} & \includegraphics[width=0.15\linewidth]{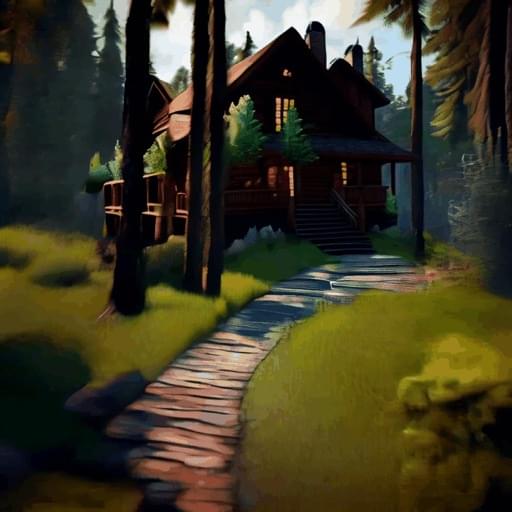} \\
\midrule
& \multicolumn{6}{c}{\it Flying through an intense battle between pirate ships in a stormy ocean..} \\  
\rotatebox[origin=l]{90}{\makebox[.15\linewidth]{Gen-2}} & \includegraphics[width=0.15\linewidth]{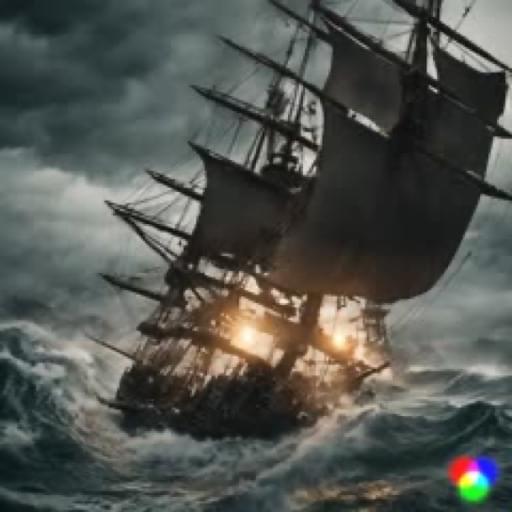} & \includegraphics[width=0.15\linewidth]{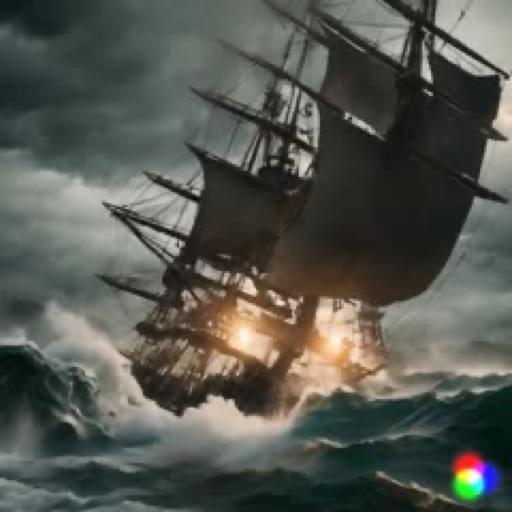} & \includegraphics[width=0.15\linewidth]{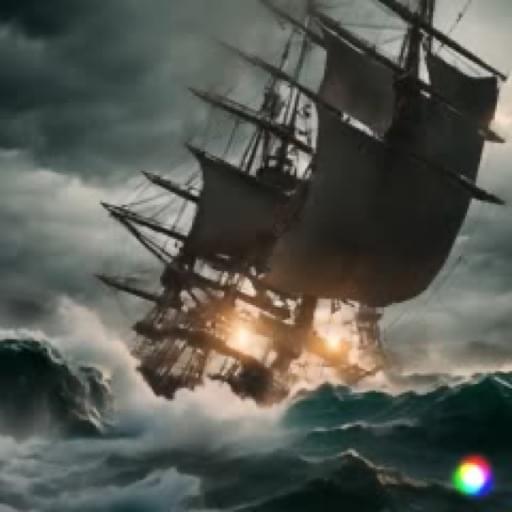} & \includegraphics[width=0.15\linewidth]{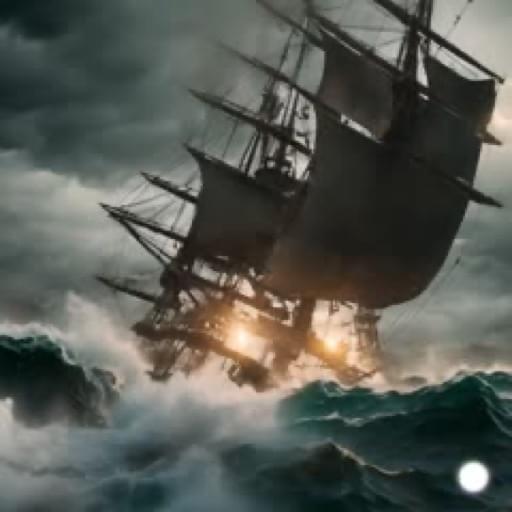} & \includegraphics[width=0.15\linewidth]{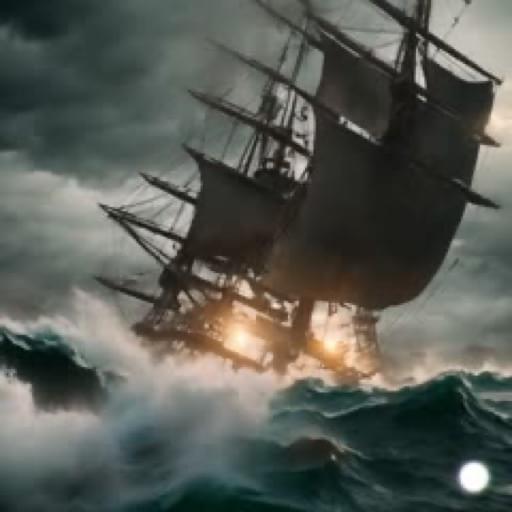} & \includegraphics[width=0.15\linewidth]{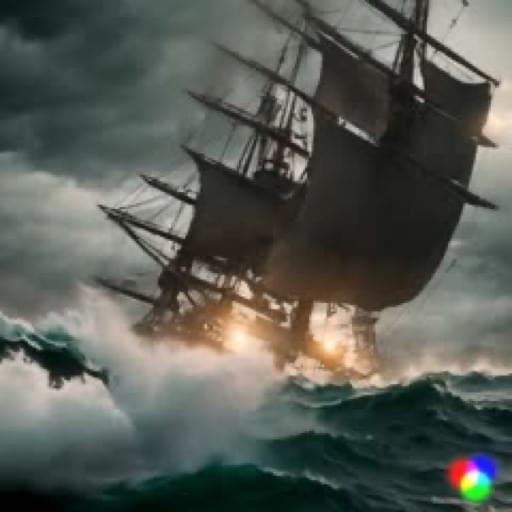} \\
\rotatebox[origin=l]{90}{\makebox[.14\linewidth]{\ours}} & \includegraphics[width=0.15\linewidth]{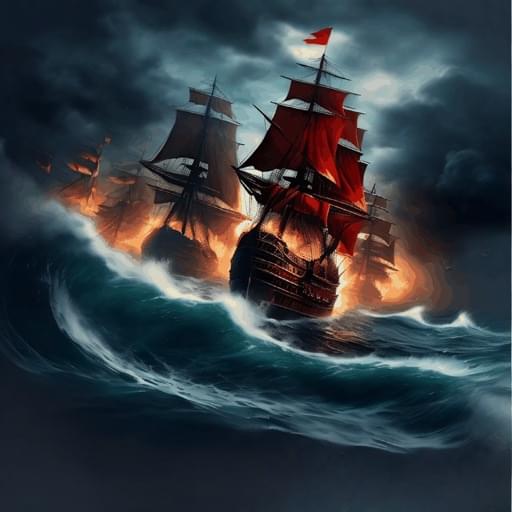} & \includegraphics[width=0.15\linewidth]{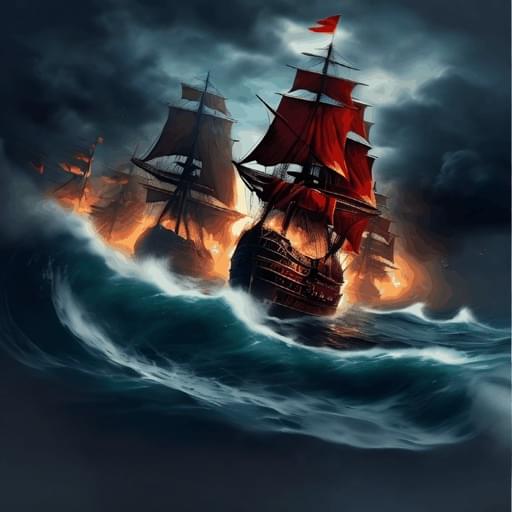} & \includegraphics[width=0.15\linewidth]{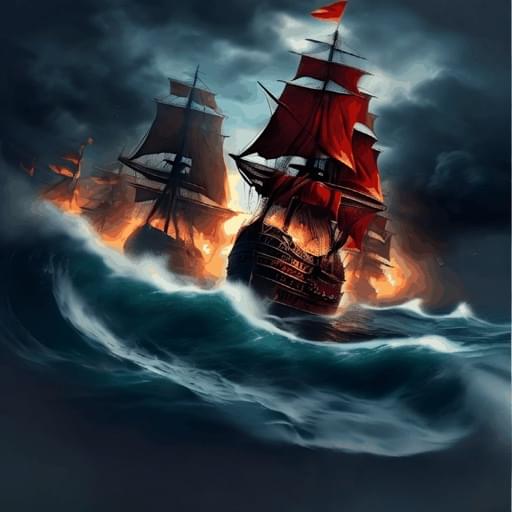} & \includegraphics[width=0.15\linewidth]{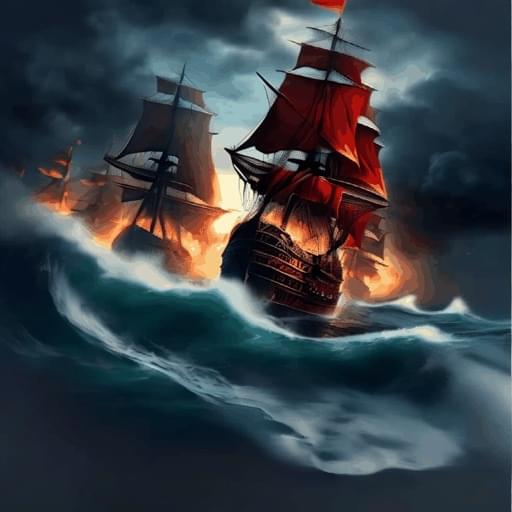} & \includegraphics[width=0.15\linewidth]{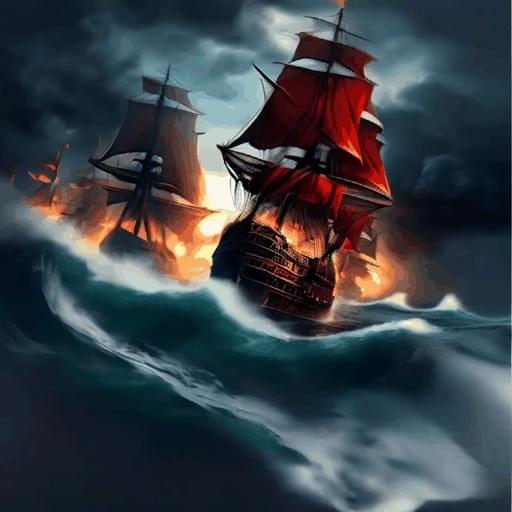} & \includegraphics[width=0.15\linewidth]{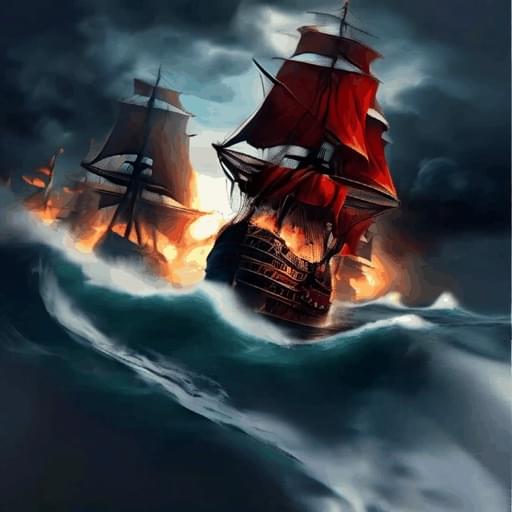} \\
\midrule
& \multicolumn{6}{c}{\it POV footage of approaching a small cottage covered in moss and many flowers, tilt shift, arc shot.} \\  
\rotatebox[origin=l]{90}{\makebox[.08\linewidth]{Gen-2}} & \includegraphics[width=0.15\linewidth]{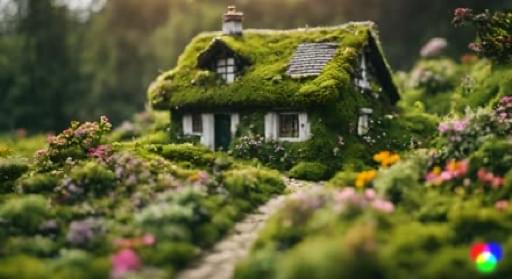} & \includegraphics[width=0.15\linewidth]{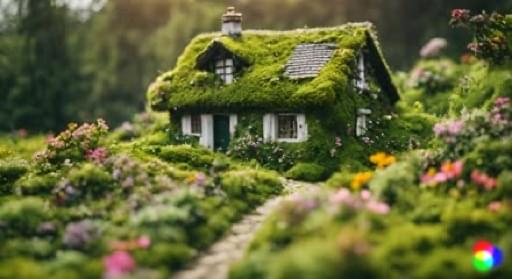} & \includegraphics[width=0.15\linewidth]{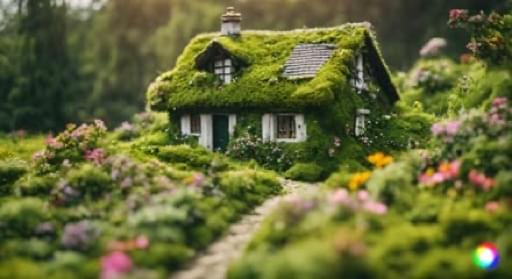} & \includegraphics[width=0.15\linewidth]{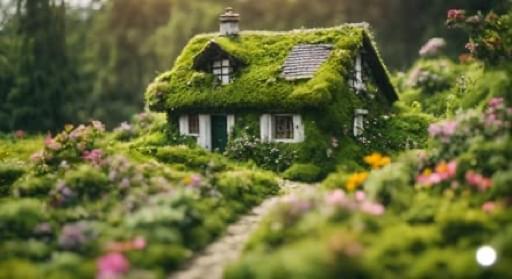} & \includegraphics[width=0.15\linewidth]{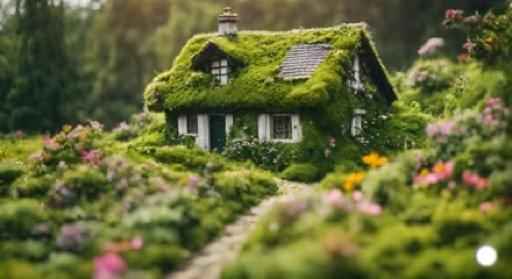} & \includegraphics[width=0.15\linewidth]{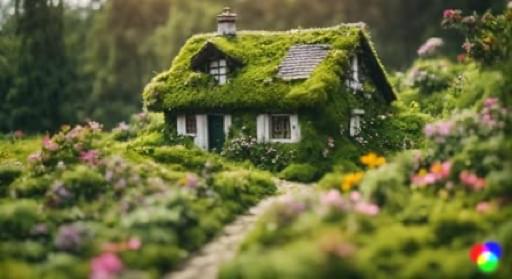} \\
\rotatebox[origin=l]{90}{\makebox[.075\linewidth]{\ours}} & \includegraphics[width=0.15\linewidth]{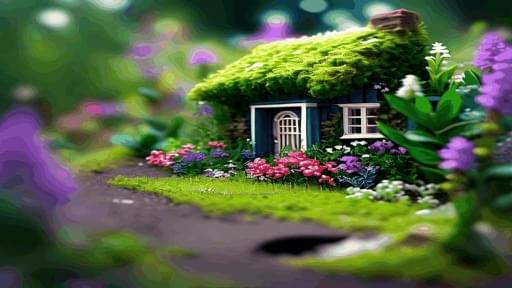} & \includegraphics[width=0.15\linewidth]{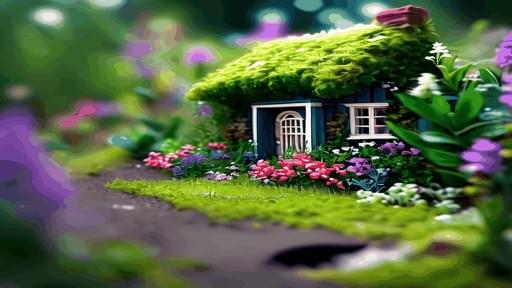} & \includegraphics[width=0.15\linewidth]{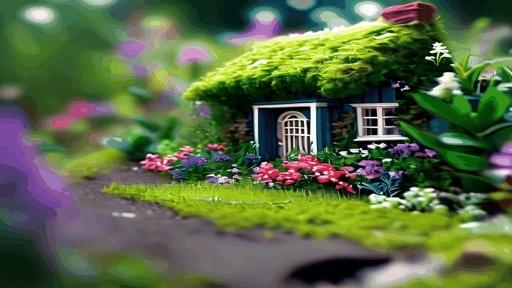} & \includegraphics[width=0.15\linewidth]{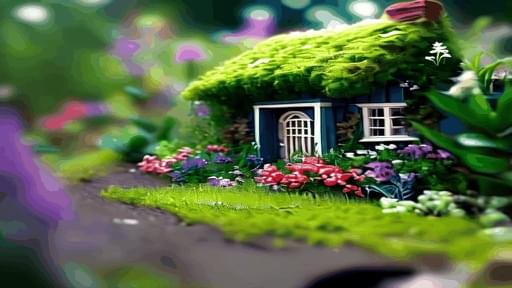} & \includegraphics[width=0.15\linewidth]{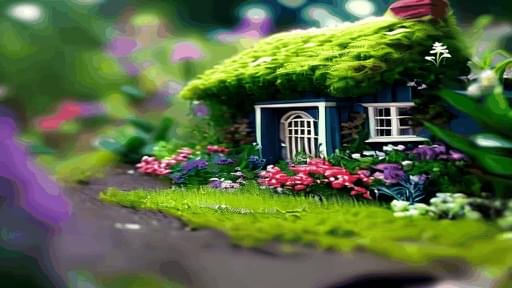} & \includegraphics[width=0.15\linewidth]{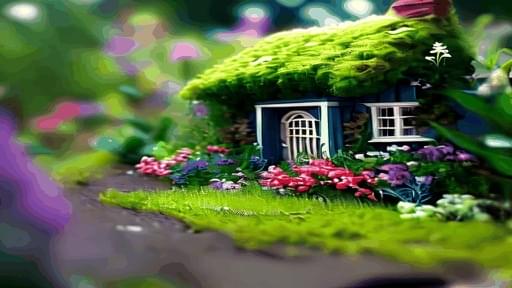} \\
\midrule
& \multicolumn{6}{c}{\it FPV drone footage of an ancient city in autumn.} \\  
\rotatebox[origin=l]{90}{\makebox[.08\linewidth]{Gen-2}} & \includegraphics[width=0.15\linewidth]{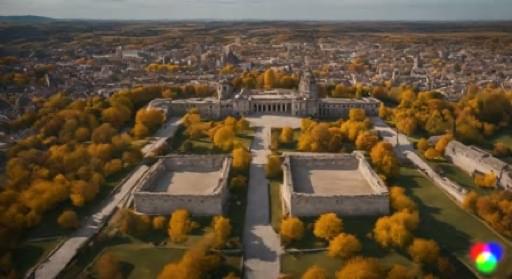} & \includegraphics[width=0.15\linewidth]{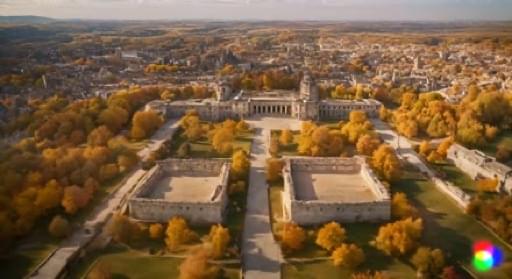} & \includegraphics[width=0.15\linewidth]{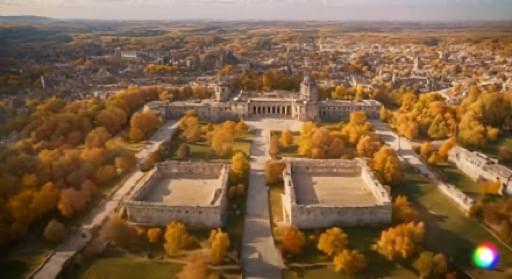} & \includegraphics[width=0.15\linewidth]{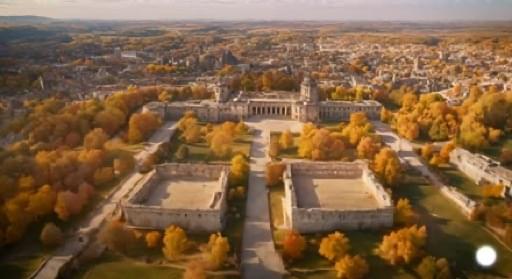} & \includegraphics[width=0.15\linewidth]{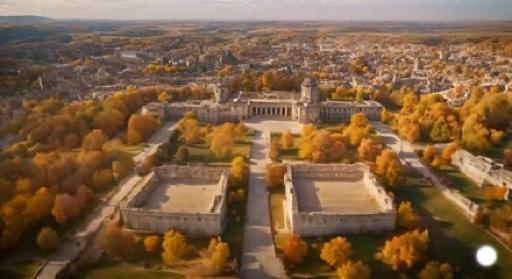} & \includegraphics[width=0.15\linewidth]{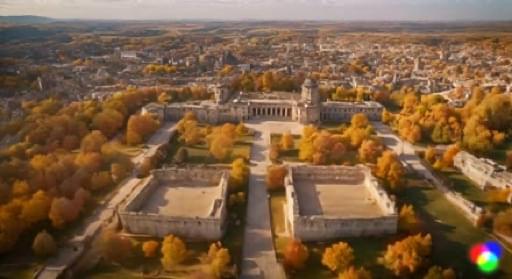} \\
\rotatebox[origin=l]{90}{\makebox[.075\linewidth]{\ours}} & \includegraphics[width=0.15\linewidth]{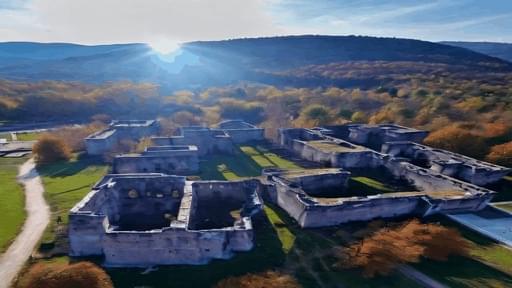} & \includegraphics[width=0.15\linewidth]{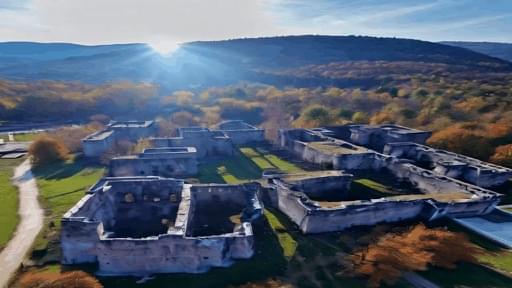} & \includegraphics[width=0.15\linewidth]{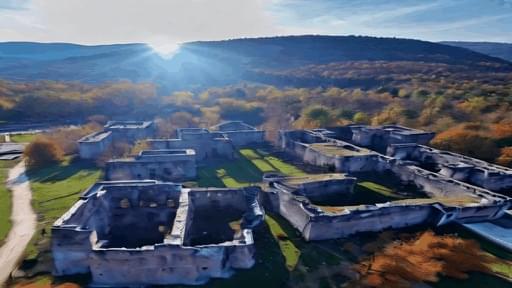} & \includegraphics[width=0.15\linewidth]{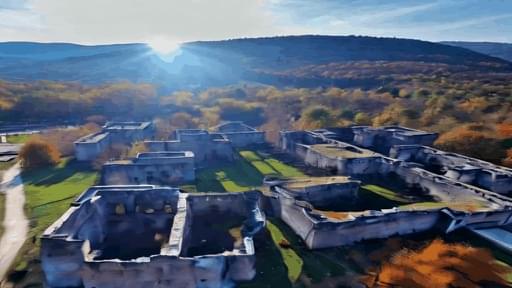} & \includegraphics[width=0.15\linewidth]{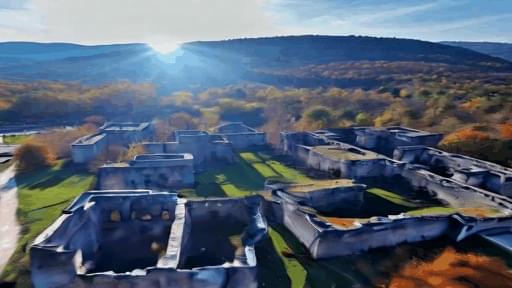} & \includegraphics[width=0.15\linewidth]{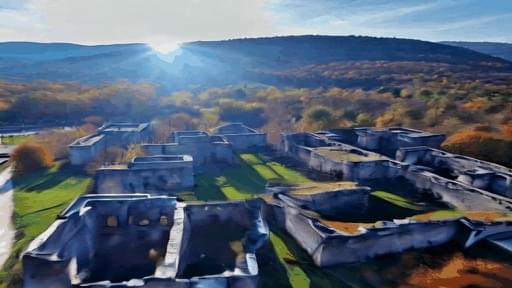} \\
\end{tabular}
\end{small}
\caption{Text-to-video generation comparison with Gen-2~\cite{runaway2023gen} using default parameters. Prompts are from VideoPoet~\cite{kondratyuk2023videopoet} and PixelDance~\cite{zeng2023make}. Our model provides a similarly high visual quality (in the bottom two cases) while following the text prompt better (including ``running'' in the first example and ``pirate ships'' in the second examples).}
\label{fig:t2v_vis2}
\end{center}
\vskip -0.1in
\end{figure}

\begin{figure}[t]
\begin{center}
\begin{small}
\begin{tabular}{lc@{}c@{}c@{}c@{}c@{}c}
\rotatebox[origin=l]{90}{\makebox[.14\linewidth]{SVD}} & \includegraphics[width=0.15\linewidth]{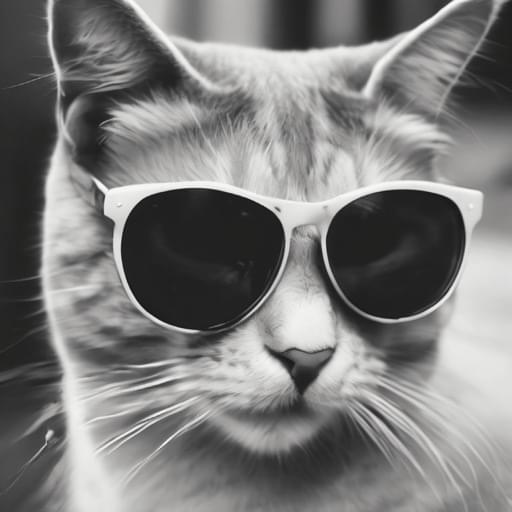} & \includegraphics[width=0.15\linewidth]{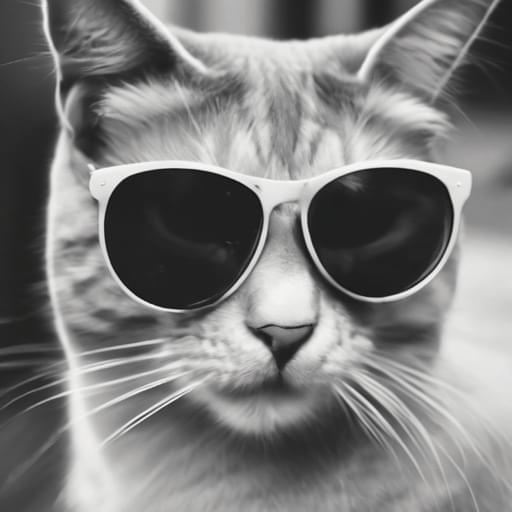} & \includegraphics[width=0.15\linewidth]{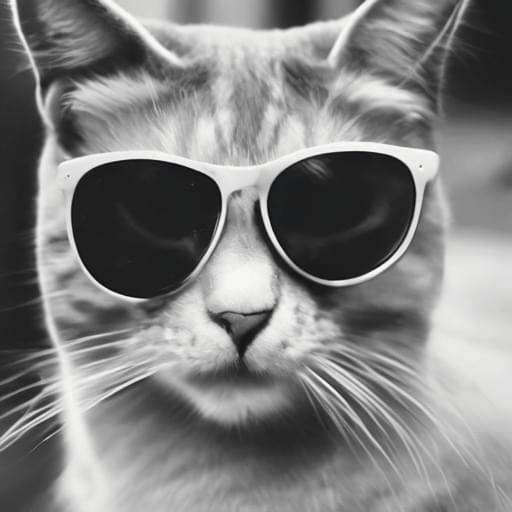} & \includegraphics[width=0.15\linewidth]{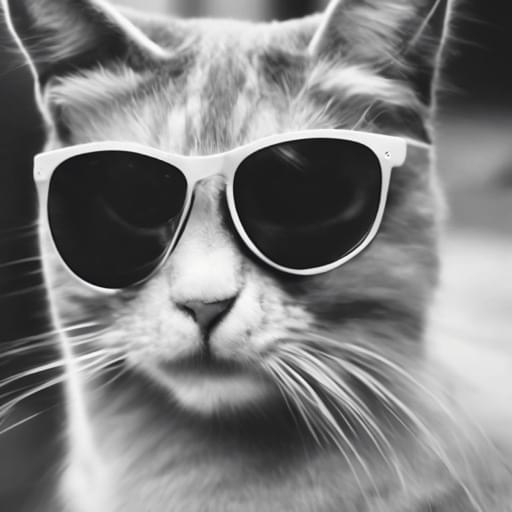} & \includegraphics[width=0.15\linewidth]{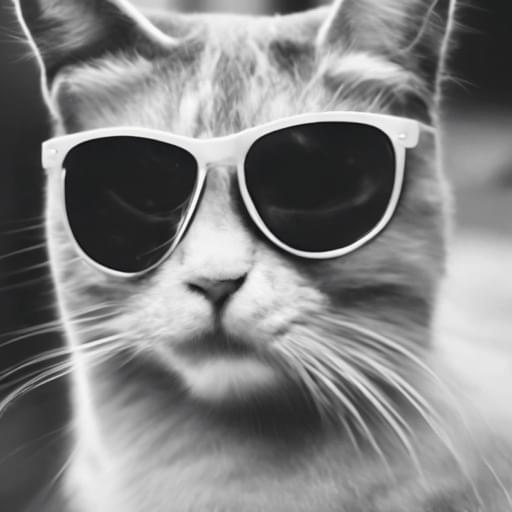} & \includegraphics[width=0.15\linewidth]{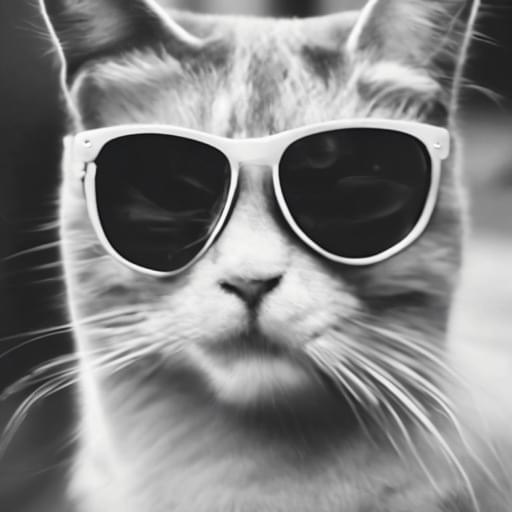} \\
\rotatebox[origin=l]{90}{\makebox[.14\linewidth]{\ours}} & \includegraphics[width=0.15\linewidth]{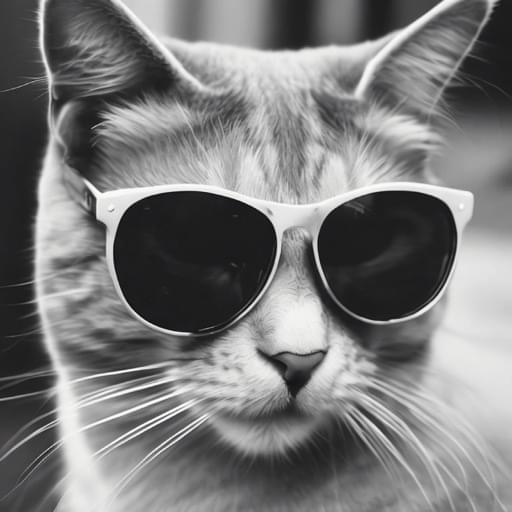} & \includegraphics[width=0.15\linewidth]{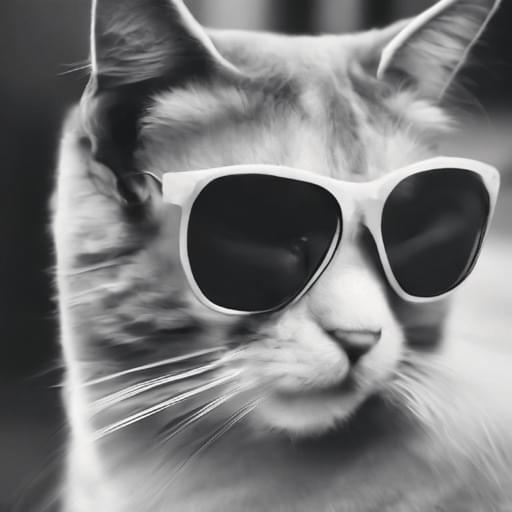} & \includegraphics[width=0.15\linewidth]{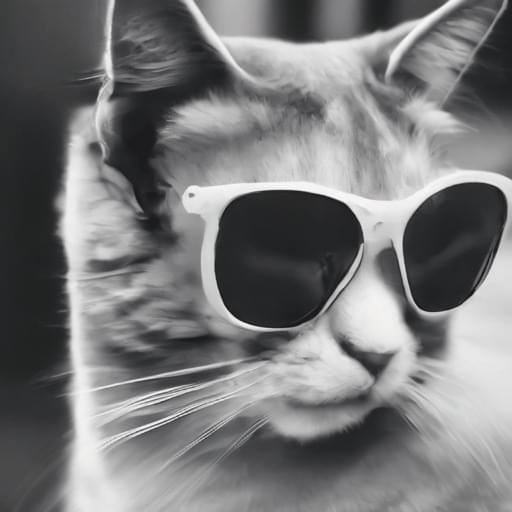} & \includegraphics[width=0.15\linewidth]{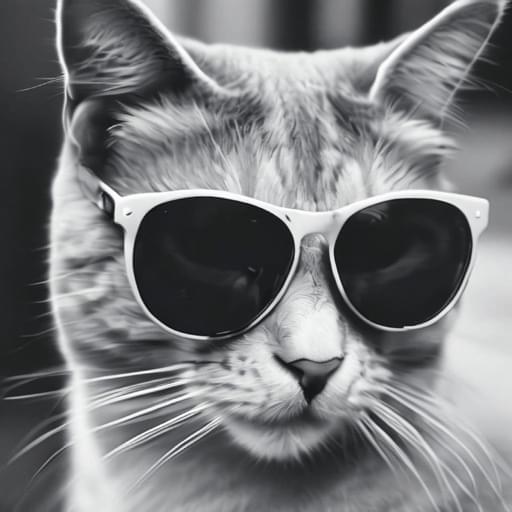} & \includegraphics[width=0.15\linewidth]{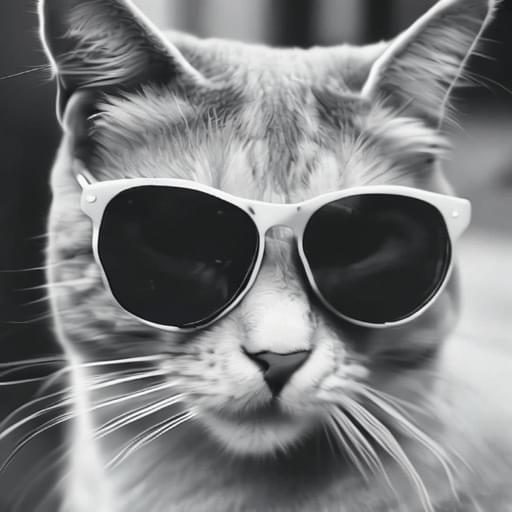} & \includegraphics[width=0.15\linewidth]{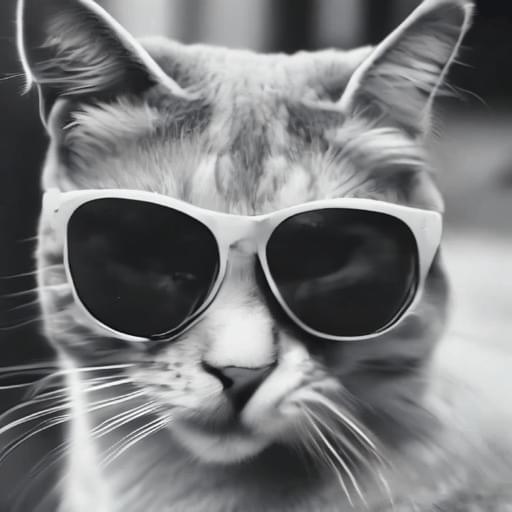} \\
\midrule
\rotatebox[origin=l]{90}{\makebox[.14\linewidth]{SVD}} & \includegraphics[width=0.15\linewidth]{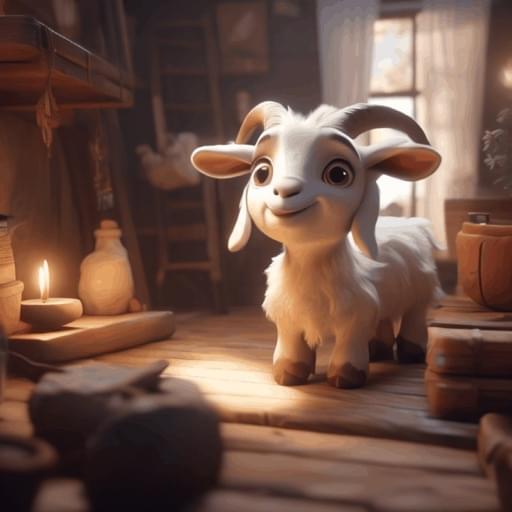} & \includegraphics[width=0.15\linewidth]{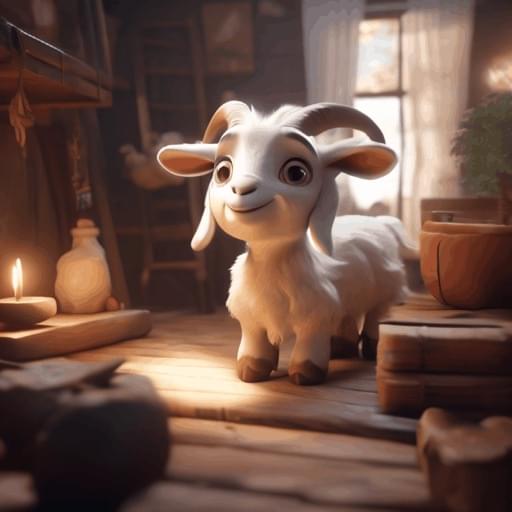} & \includegraphics[width=0.15\linewidth]{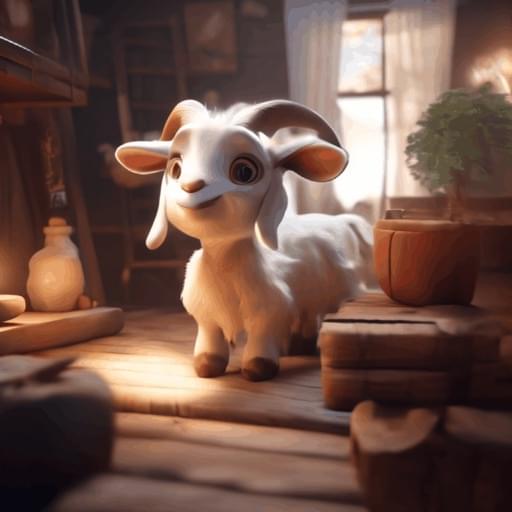} & \includegraphics[width=0.15\linewidth]{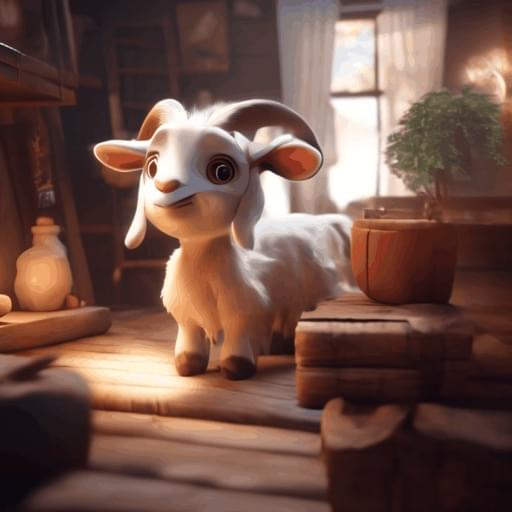} & \includegraphics[width=0.15\linewidth]{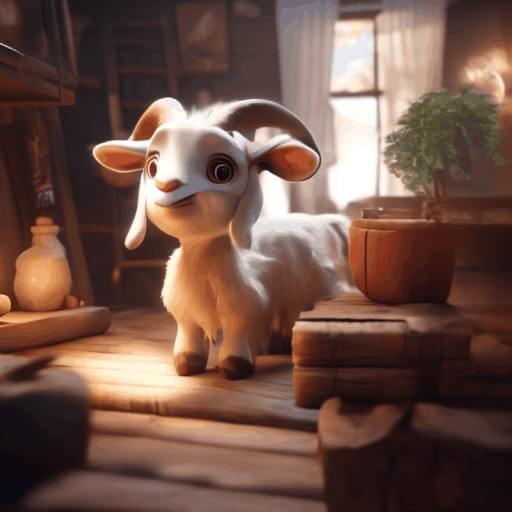} & \includegraphics[width=0.15\linewidth]{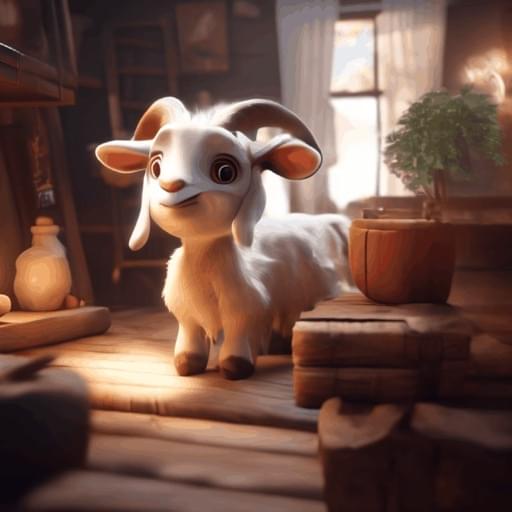} \\
\rotatebox[origin=l]{90}{\makebox[.14\linewidth]{\ours}} & \includegraphics[width=0.15\linewidth]{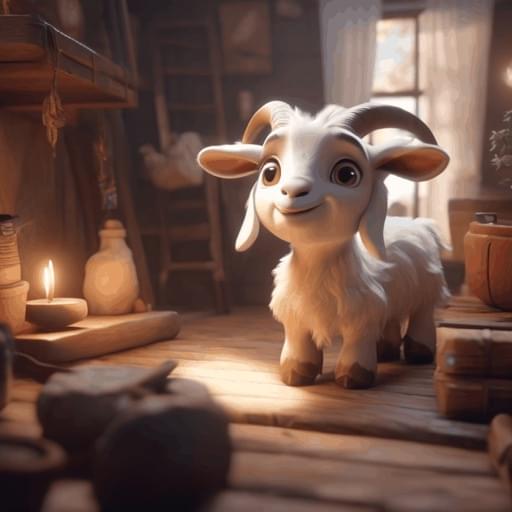} & \includegraphics[width=0.15\linewidth]{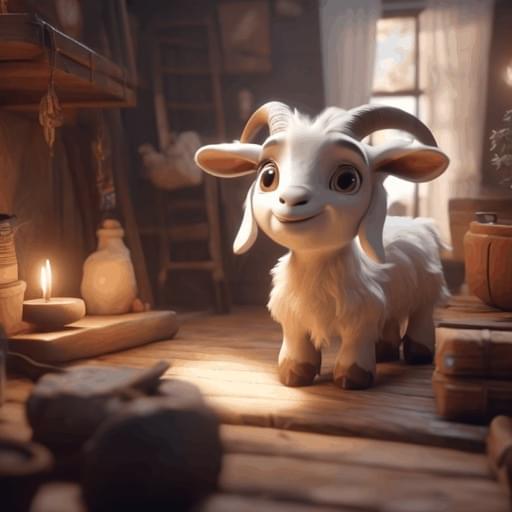} & \includegraphics[width=0.15\linewidth]{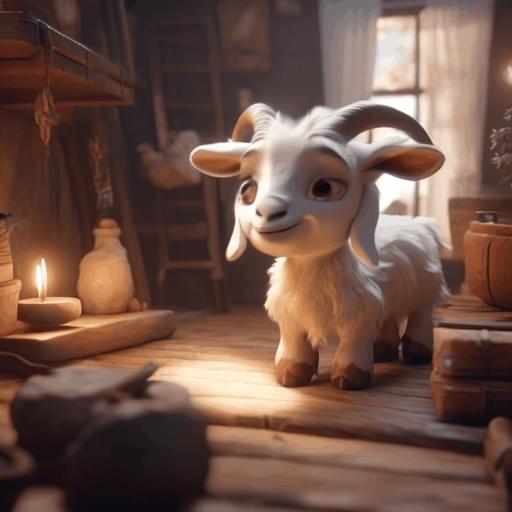} & \includegraphics[width=0.15\linewidth]{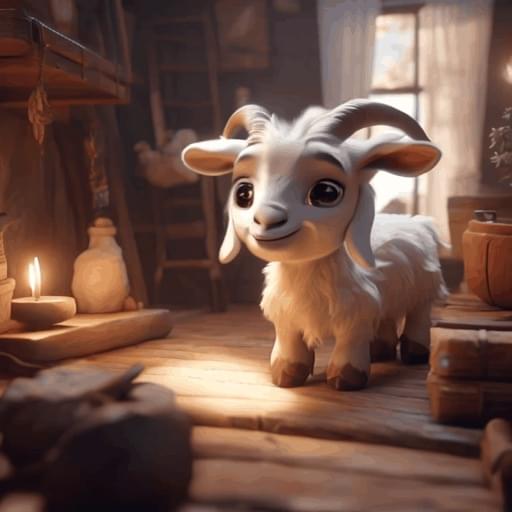} & \includegraphics[width=0.15\linewidth]{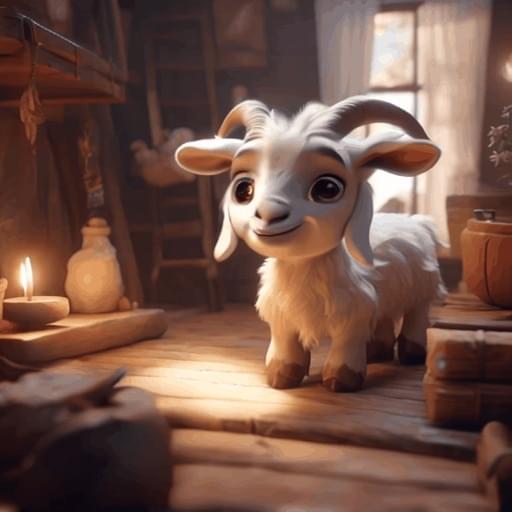} & \includegraphics[width=0.15\linewidth]{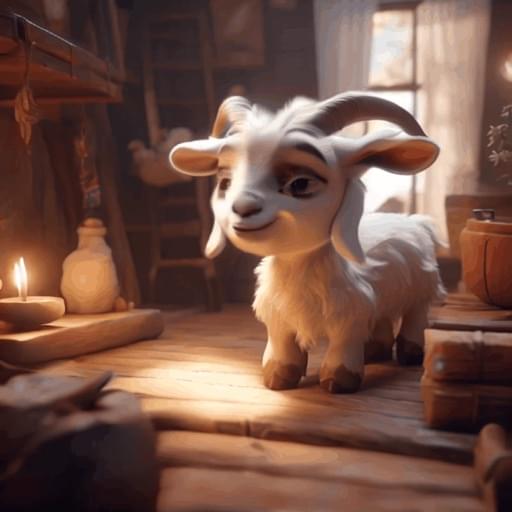} \\
\midrule
\rotatebox[origin=l]{90}{\makebox[.14\linewidth]{SVD}} & \includegraphics[width=0.15\linewidth]{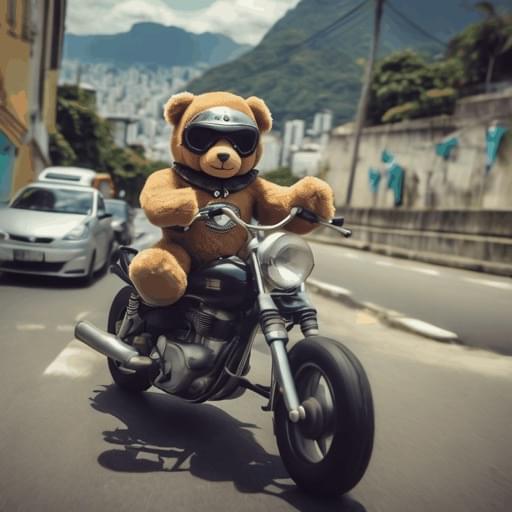} & \includegraphics[width=0.15\linewidth]{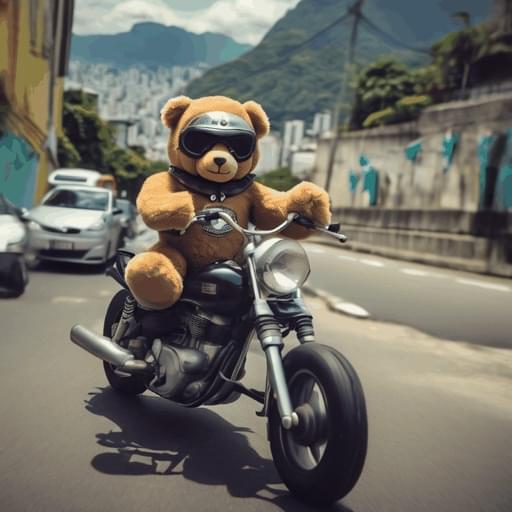} & \includegraphics[width=0.15\linewidth]{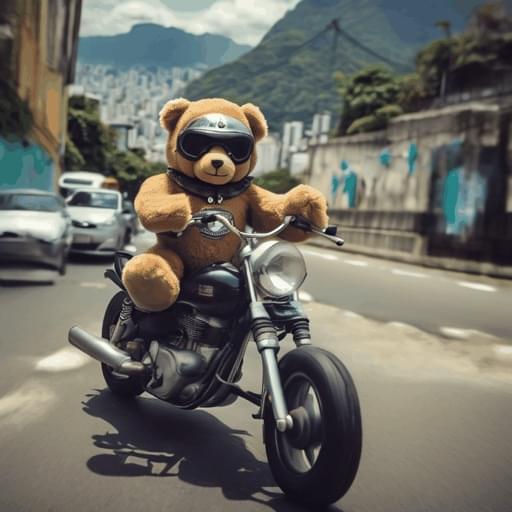} & \includegraphics[width=0.15\linewidth]{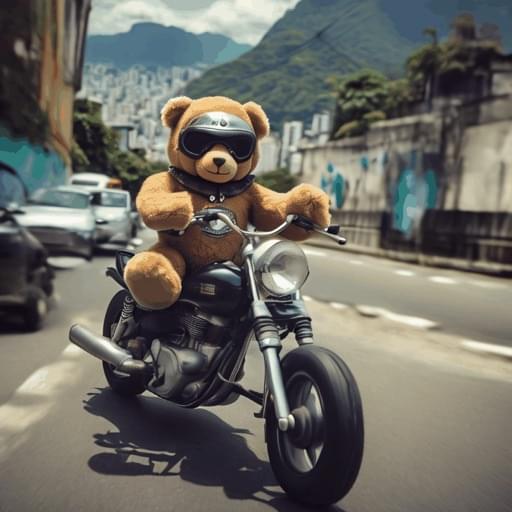} & \includegraphics[width=0.15\linewidth]{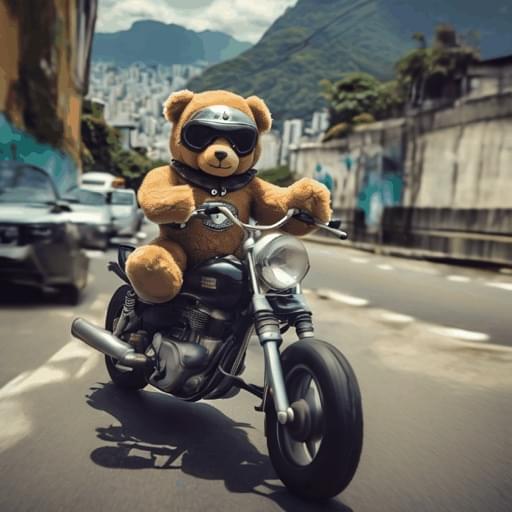} & \includegraphics[width=0.15\linewidth]{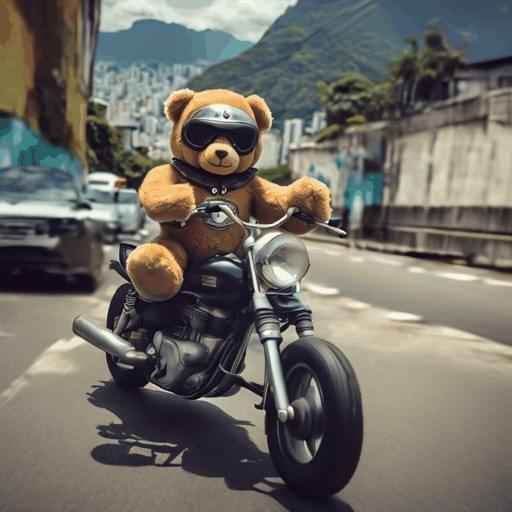} \\
\rotatebox[origin=l]{90}{\makebox[.14\linewidth]{\ours}} & \includegraphics[width=0.15\linewidth]{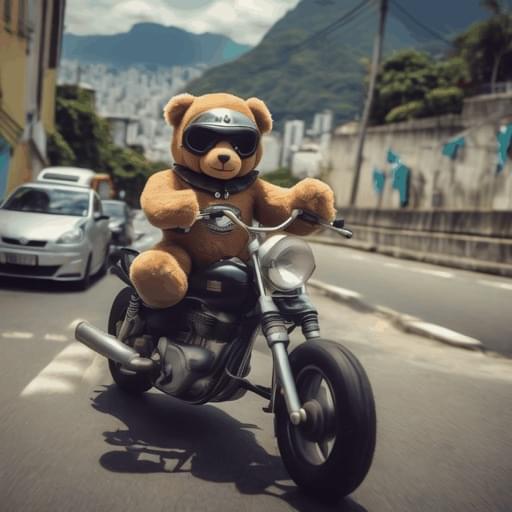} & \includegraphics[width=0.15\linewidth]{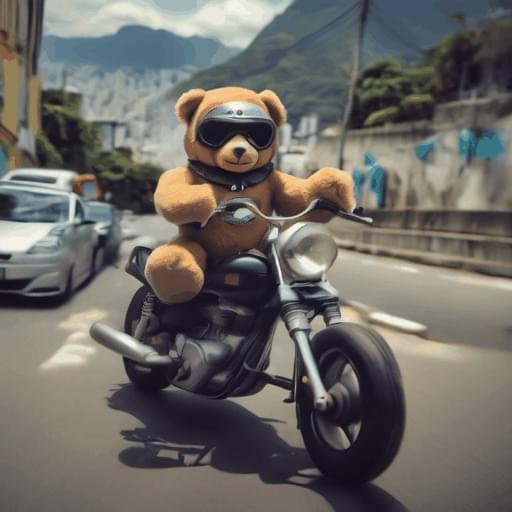} & \includegraphics[width=0.15\linewidth]{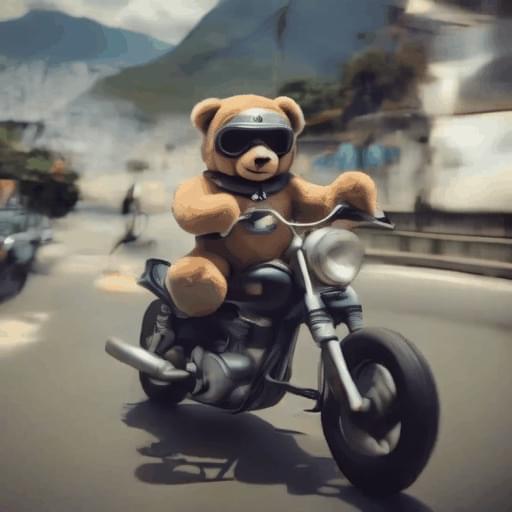} & \includegraphics[width=0.15\linewidth]{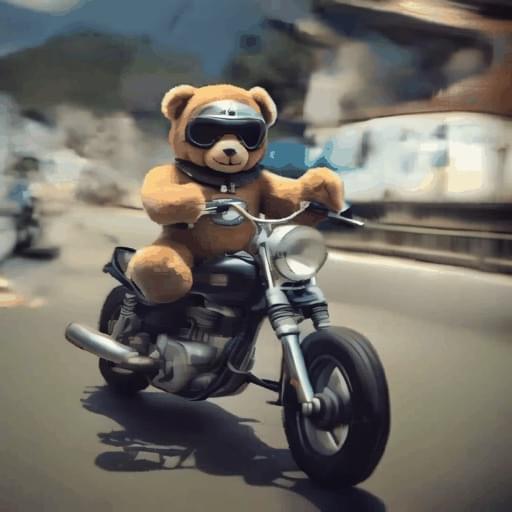} & \includegraphics[width=0.15\linewidth]{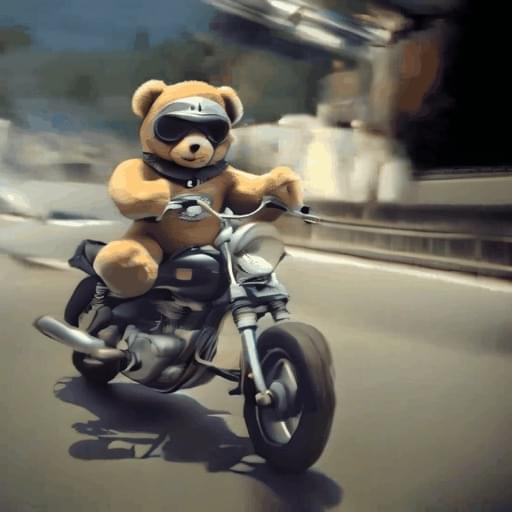} & \includegraphics[width=0.15\linewidth]{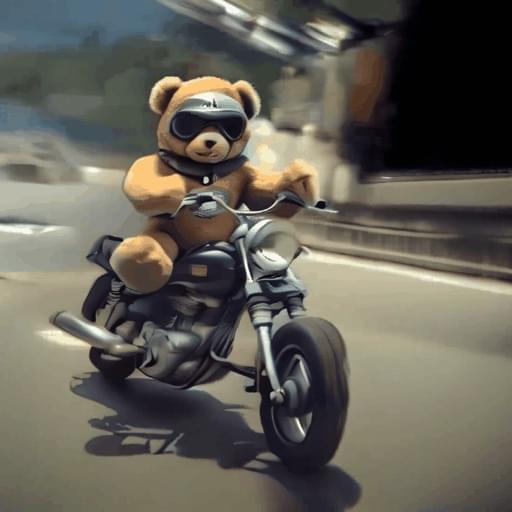} \\
\end{tabular}
\end{small}
\caption{Image-to-video generation comparison with SVD~\cite{blattmann2023stable} using the stable-video-diffusion-img2vid-xt version. The generation is conditioned on the leftmost frame. Our model can produce more sophisticated animal motions (see the top two cases) while not violating the physical rules (e.g., in the second last row, the motorcycle is not moving in the direction of its tire).}
\label{fig:i2v_vis}
\end{center}
\vskip -0.1in
\end{figure}

\begin{figure}[t]
\begin{center}
\begin{small}
\begin{tabular}{lc@{}c@{}c@{}c@{}c@{}c}
& \multicolumn{6}{c}{\it A dog in the sun.} \\  
\rotatebox[origin=l]{90}{\makebox[.15\linewidth]{0--24 frames}} & \includegraphics[width=0.15\linewidth]{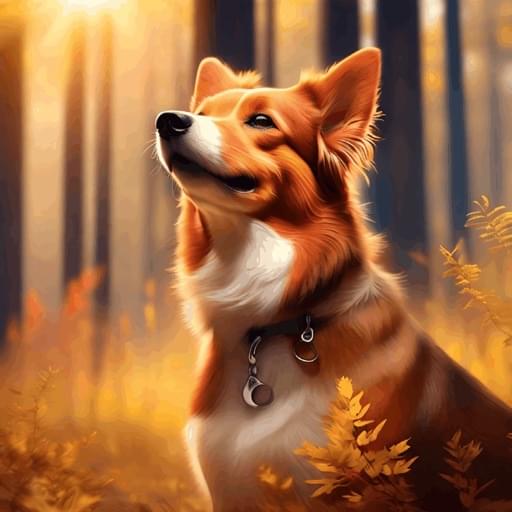} & \includegraphics[width=0.15\linewidth]{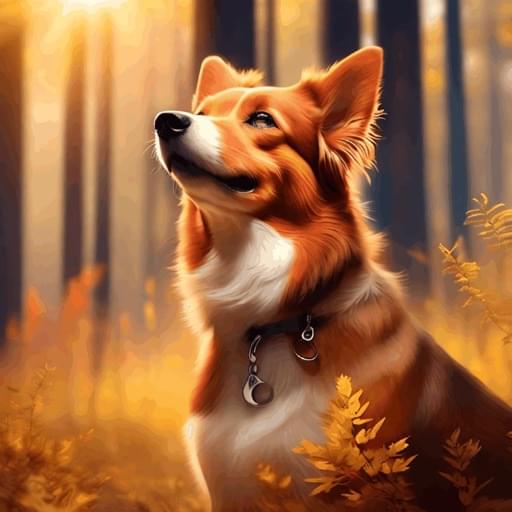} & \includegraphics[width=0.15\linewidth]{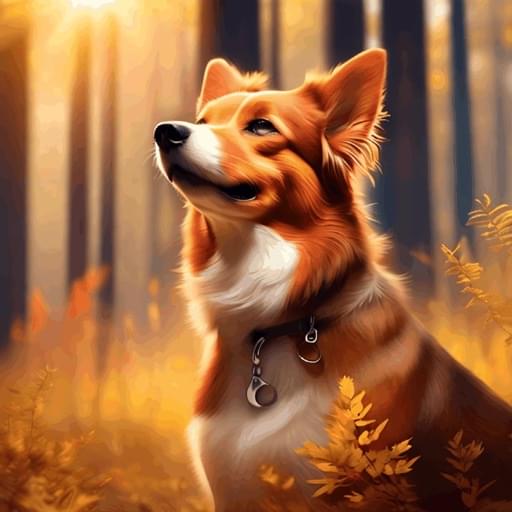} & \includegraphics[width=0.15\linewidth]{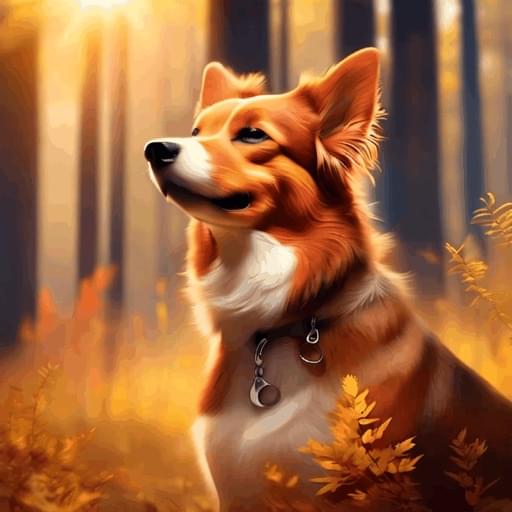} & \includegraphics[width=0.15\linewidth]{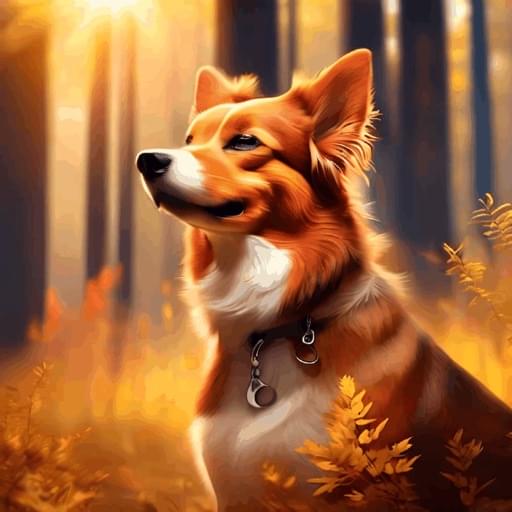} & \includegraphics[width=0.15\linewidth]{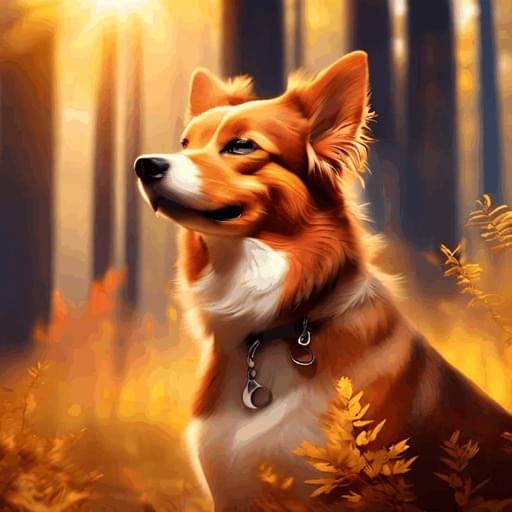} \\
\rotatebox[origin=l]{90}{\makebox[.14\linewidth]{24--48 frames}} & \includegraphics[width=0.15\linewidth]{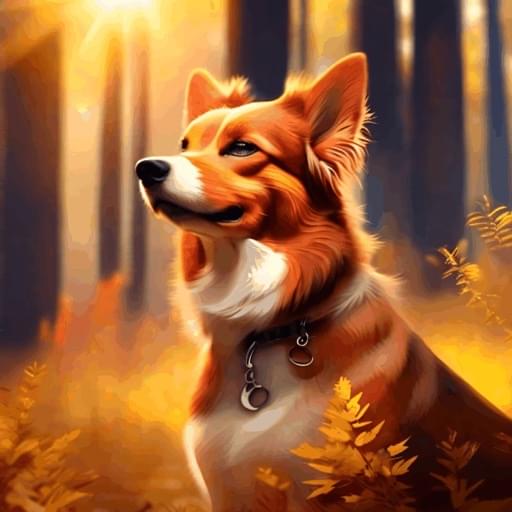} & \includegraphics[width=0.15\linewidth]{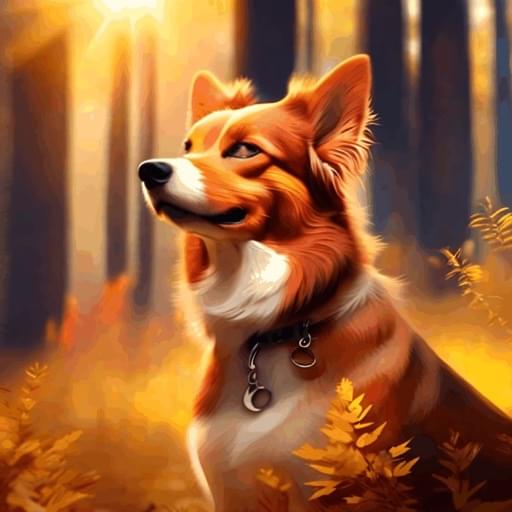} & \includegraphics[width=0.15\linewidth]{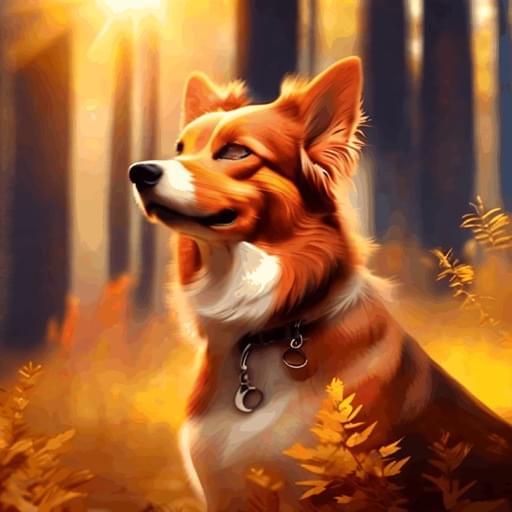} & \includegraphics[width=0.15\linewidth]{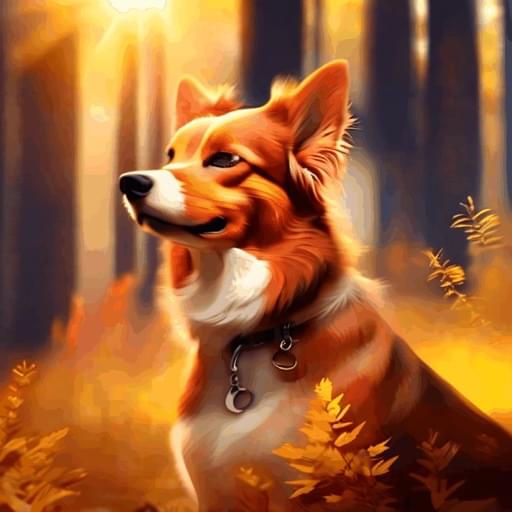} & \includegraphics[width=0.15\linewidth]{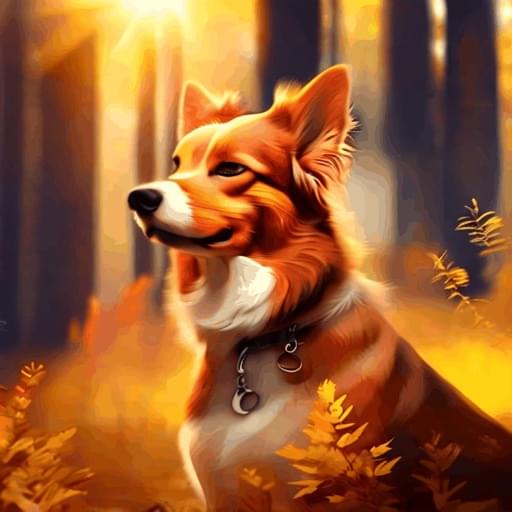} & \includegraphics[width=0.15\linewidth]{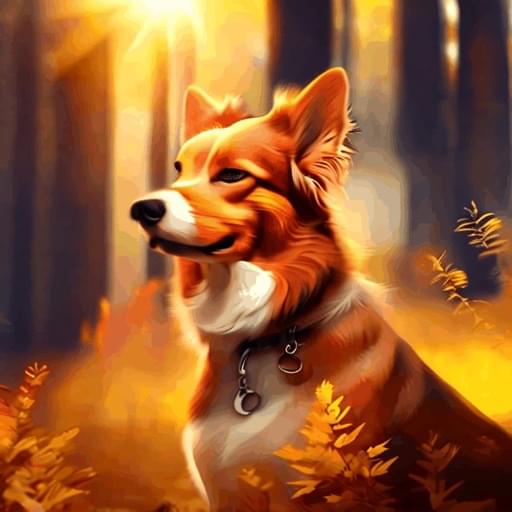} \\
\midrule
& \multicolumn{6}{c}{\it A jeep car is moving on the beach.} \\  
\rotatebox[origin=l]{90}{\makebox[.15\linewidth]{0--24 frames}} & \includegraphics[width=0.15\linewidth]{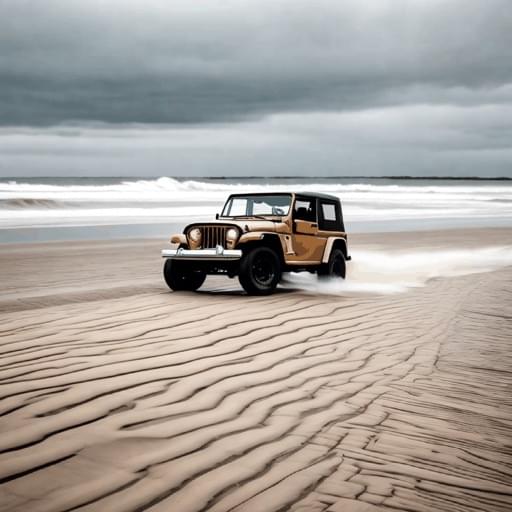} & \includegraphics[width=0.15\linewidth]{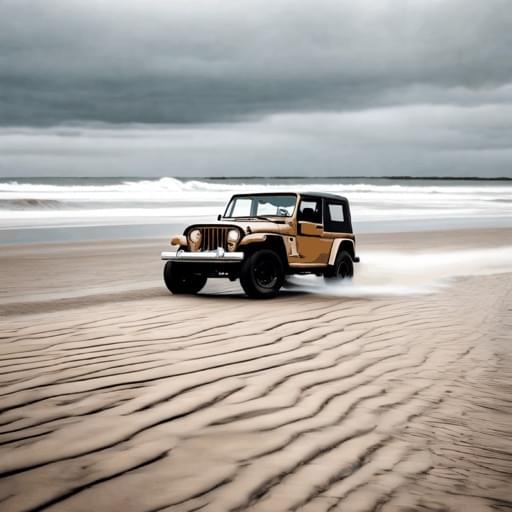} & \includegraphics[width=0.15\linewidth]{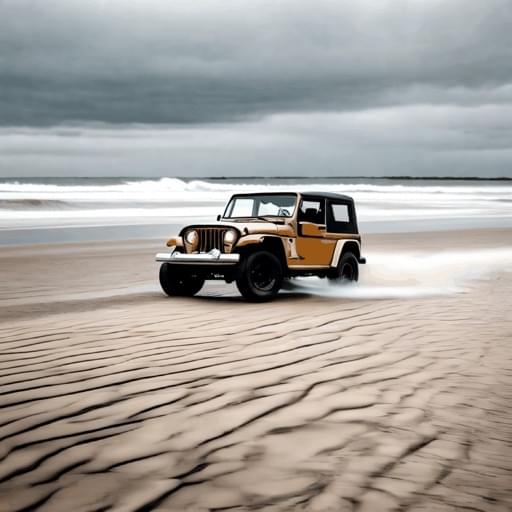} & \includegraphics[width=0.15\linewidth]{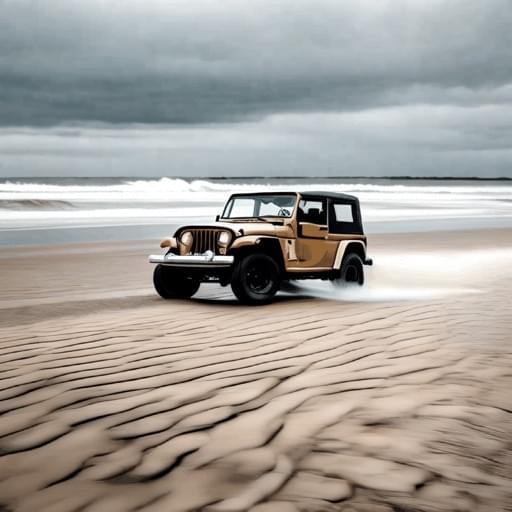} & \includegraphics[width=0.15\linewidth]{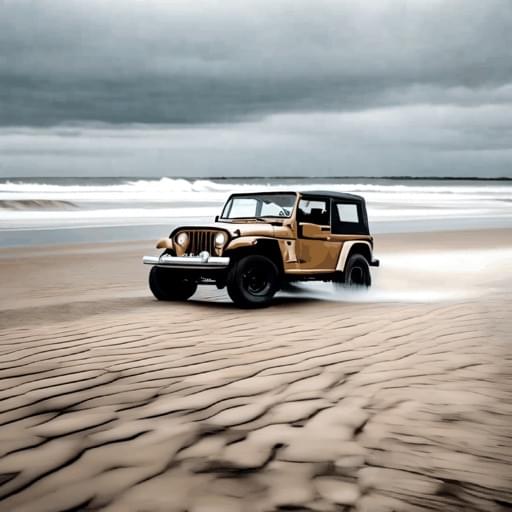} & \includegraphics[width=0.15\linewidth]{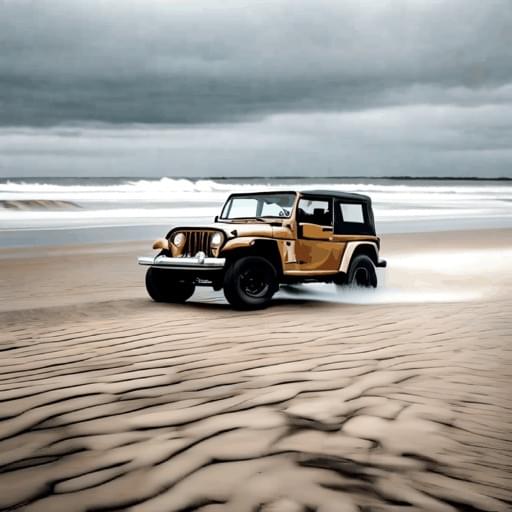} \\
\rotatebox[origin=l]{90}{\makebox[.14\linewidth]{24--48 frames}} & \includegraphics[width=0.15\linewidth]{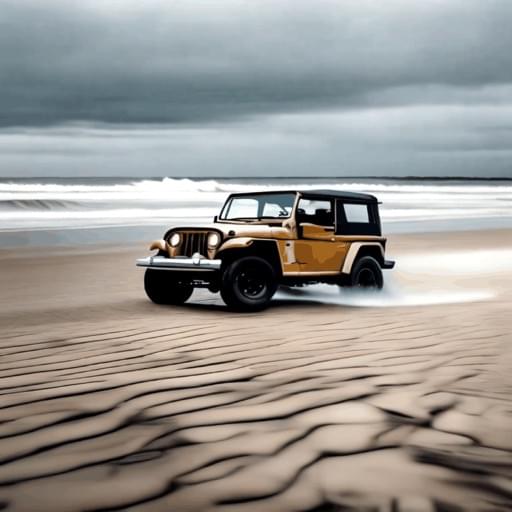} & \includegraphics[width=0.15\linewidth]{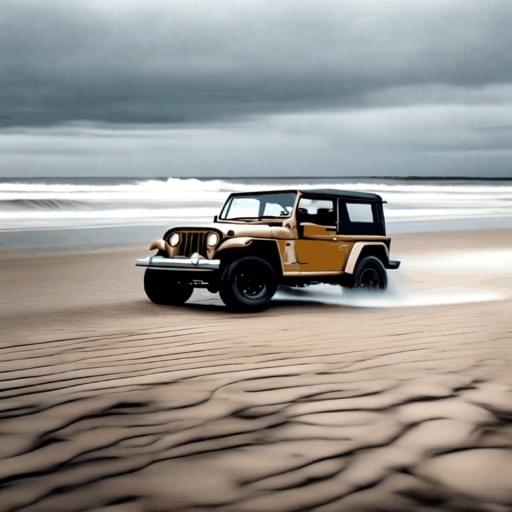} & \includegraphics[width=0.15\linewidth]{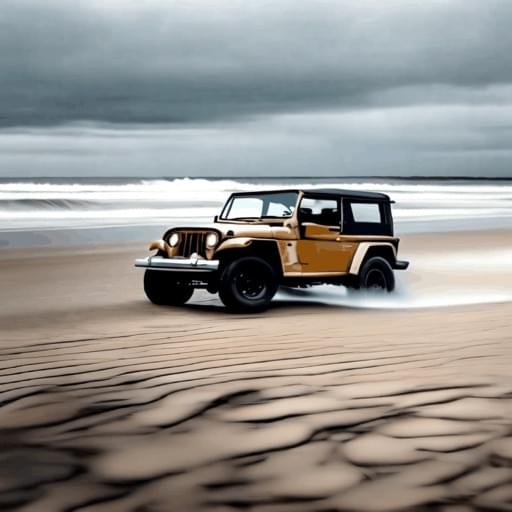} & \includegraphics[width=0.15\linewidth]{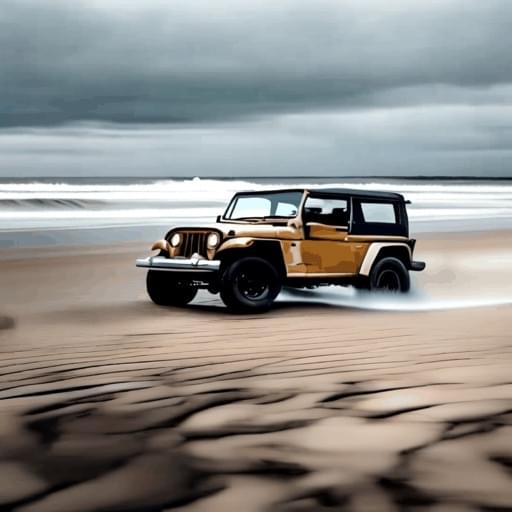} & \includegraphics[width=0.15\linewidth]{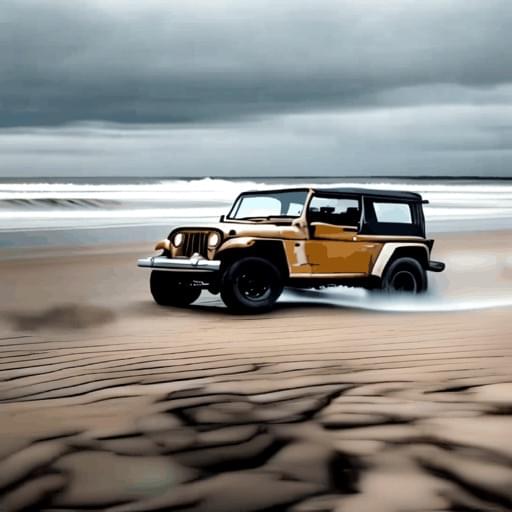} & \includegraphics[width=0.15\linewidth]{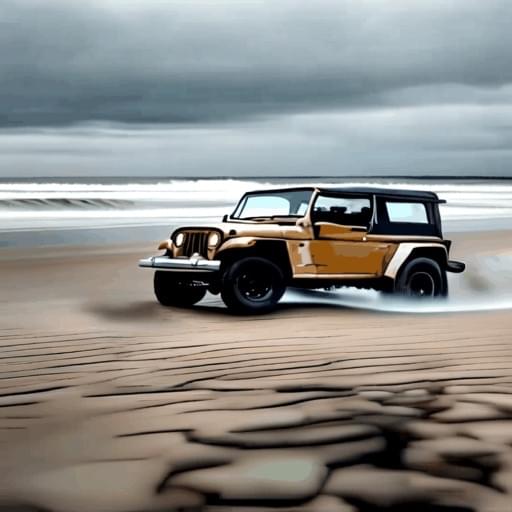} \\
\midrule
& \multicolumn{6}{c}{\it First-person view running through the woods and approaching a large beautiful cabin, highly detailed.} \\  
\rotatebox[origin=l]{90}{\makebox[.15\linewidth]{0--24 frames}} & \includegraphics[width=0.15\linewidth]{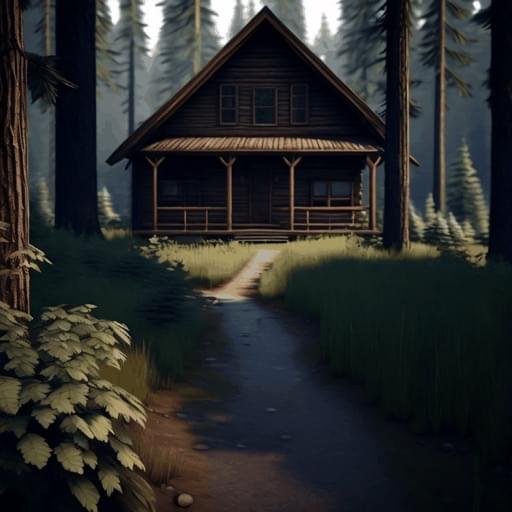} & \includegraphics[width=0.15\linewidth]{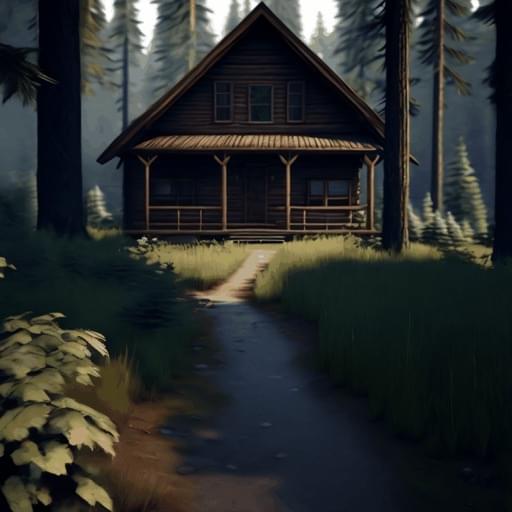} & \includegraphics[width=0.15\linewidth]{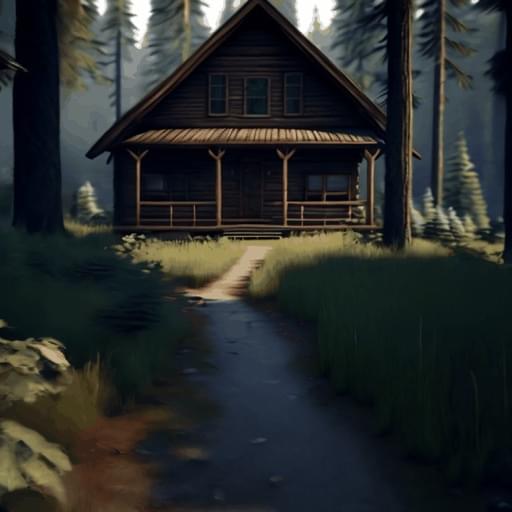} & \includegraphics[width=0.15\linewidth]{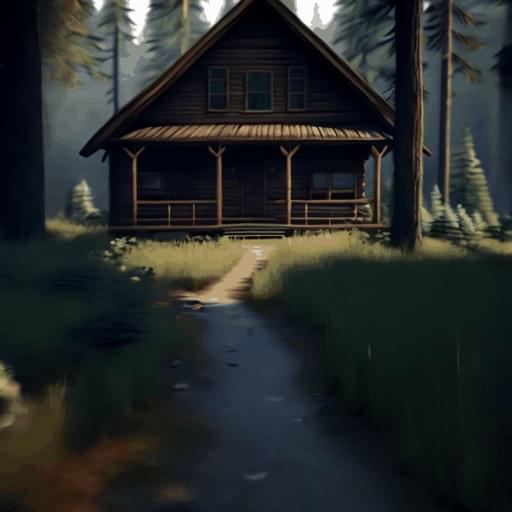} & \includegraphics[width=0.15\linewidth]{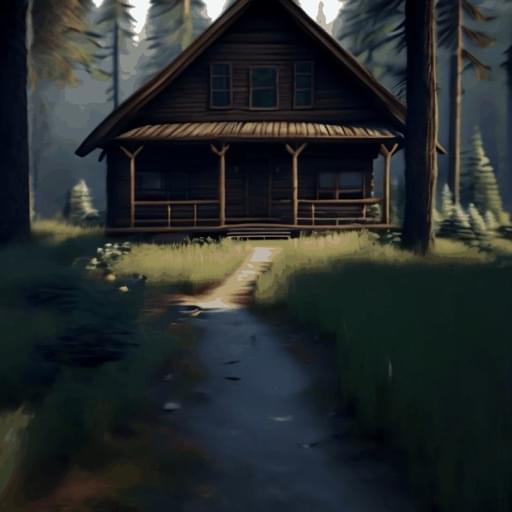} & \includegraphics[width=0.15\linewidth]{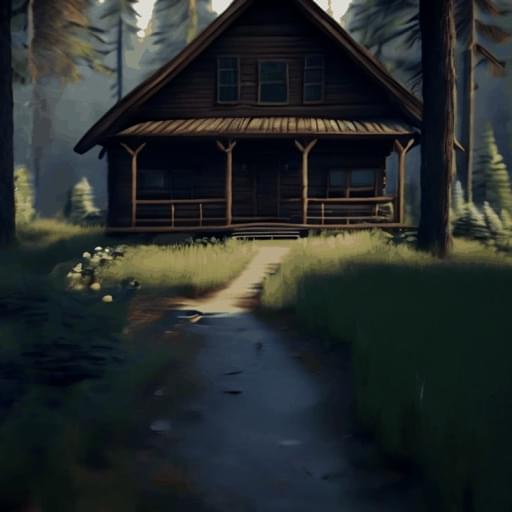} \\
\rotatebox[origin=l]{90}{\makebox[.14\linewidth]{24--48 frames}} & \includegraphics[width=0.15\linewidth]{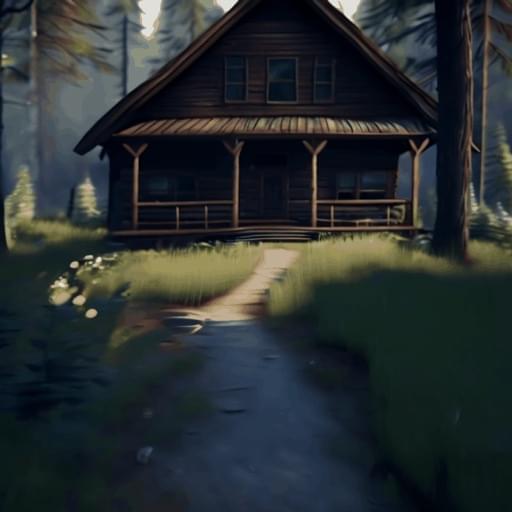} & \includegraphics[width=0.15\linewidth]{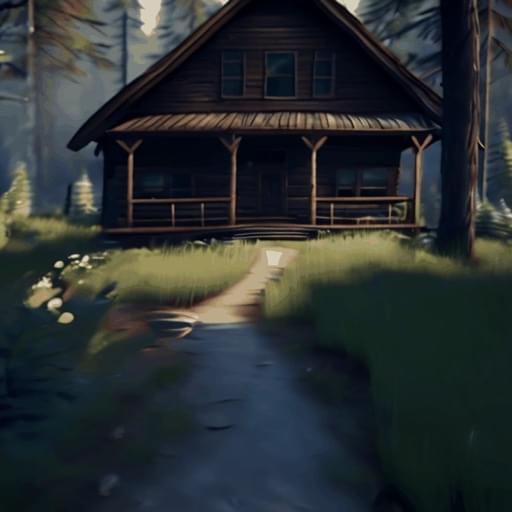} & \includegraphics[width=0.15\linewidth]{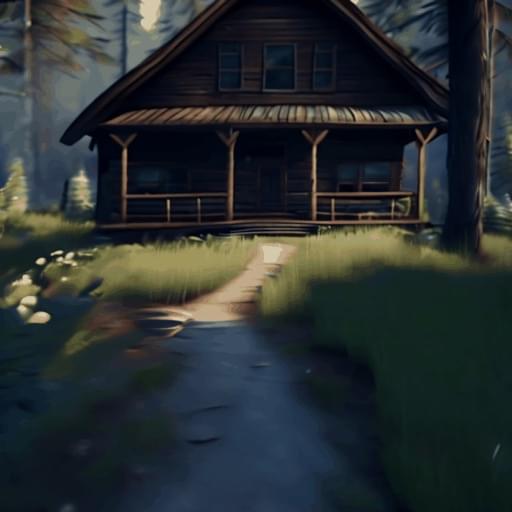} & \includegraphics[width=0.15\linewidth]{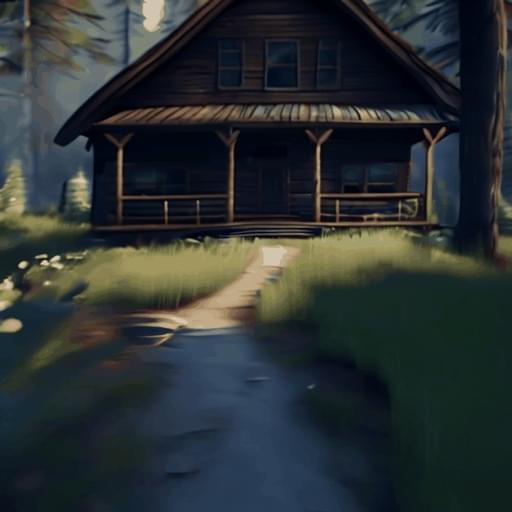} & \includegraphics[width=0.15\linewidth]{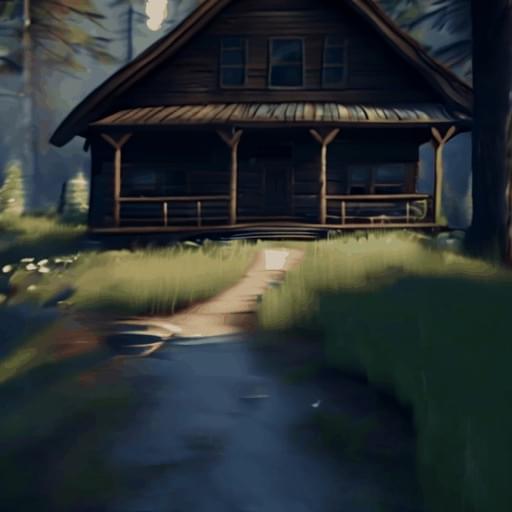} & \includegraphics[width=0.15\linewidth]{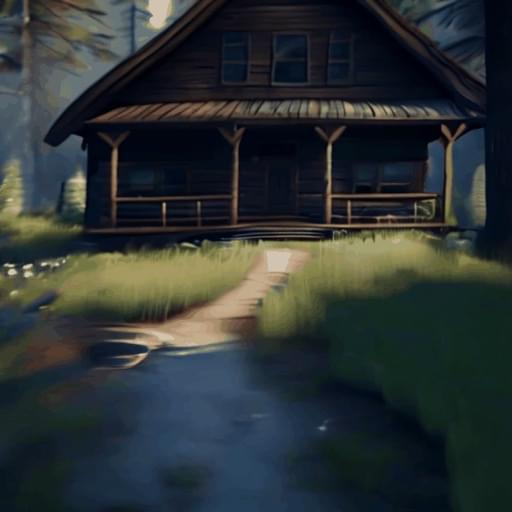} \\
\end{tabular}
\end{small}
\caption{Long video generation examples. Prompts are Gen-L-Video~\cite{wang2023gen} and VideoPoet~\cite{kondratyuk2023videopoet}. Our generated videos are temporally coherent even across different decoded clips, thanks to our proposed explicit noise constraint.}
\label{fig:long2}
\end{center}
\vskip -0.1in
\end{figure}

\textbf{Text-to-Video Generation}. \Cref{fig:t2v_vis2} compares \ours~to a closed-source model Gen-2~\cite{runaway2023gen}. As can be seen, our model produces high-quality videos that are generally comparable to Gen-2, which is especially evident in the last two examples where it successfully captures details such as "moss and many flowers" and "autumn" in the text prompt and yields very similar results to Gen-2. Moreover, the first two comparisons demonstrate a favorable prompt following ability of our model. In the first case with the keyword ``running'', our model produces significant camera motion toward the cabin, while the movement in Gen-2 is relatively nuanced. In the second case, our model correctly displays multiple ``pirate ships'' as the prompt specified, with artistic details such as all the ships being on fire, according to the implication of ``intense battle''. These results support the benefits of unified video-language pre-training in prompt following capabilities.

\textbf{Image-to-video generation}. \Cref{fig:i2v_vis} presents a comparison of \ours~with the open-source model SVD~\cite{blattmann2023stable}, both conditioned on synthetic image prompts. Moving to some unseen test cases, our method produces video clips featuring both natural and refined motions, thanks to the decomposed video representation that can better transfer motion-related knowledge to new visual inputs. For example, in the middle case, our generated goat smoothly lowers its head and blinks as if it were a human to think, while the goat in the video produced by SVD hardly moved. In the bottom case, where the image prompt shows a teddy is riding a motorcycle, our generated full video looks very natural and similar to a human riding a motorcycle, while SVD constantly produces a scenario where the motorcycle is moving a different direction from where its tire is pointing (which is physically wrong). Overall, our model demonstrates superior image-to-video generation performance with the inclusion of decoupled visual-motion tokenization and LLM pre-training.

\textbf{Long Video Generation} is showcased in~\cref{fig:long2}. By explicitly constraining the noise when decoding successive video clips, our model can provide a high temporal consistency during long video generation. For example, in the first two cases, the dog and the jeep car maintain the same identity across different clips with highly coherent visual details. In the last example which features large camera movement, the moving trajectory remains consistent as it approaches the cabin according to the text prompt. These examples all illustrate our reasonably good quality of long video generation. Note that all the generated videos are provided at \web.

\begin{figure}[t]
\begin{center}
\begin{small}
\begin{tabular}{lc@{}c@{}c@{}c@{}c@{}c}
\rotatebox[origin=l]{90}{\makebox[.08\linewidth]{Original}} & 
\includegraphics[width=0.15\linewidth]{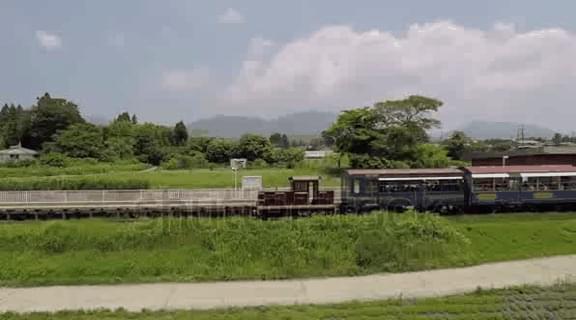} & \includegraphics[width=0.15\linewidth]{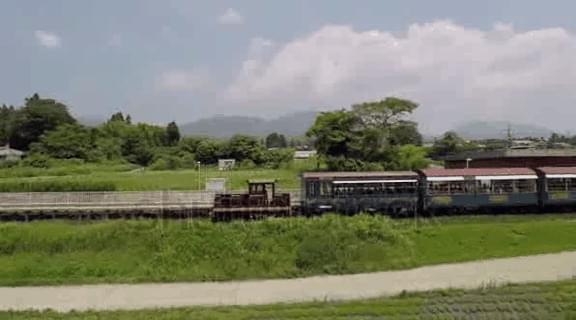} & \includegraphics[width=0.15\linewidth]{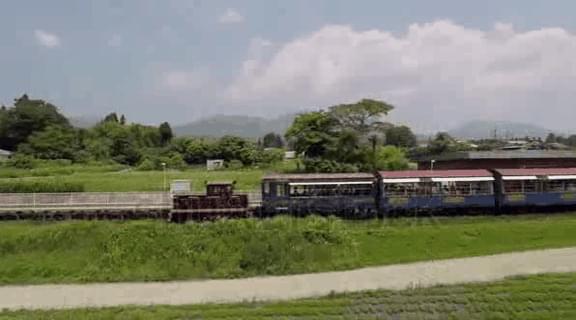} & \includegraphics[width=0.15\linewidth]{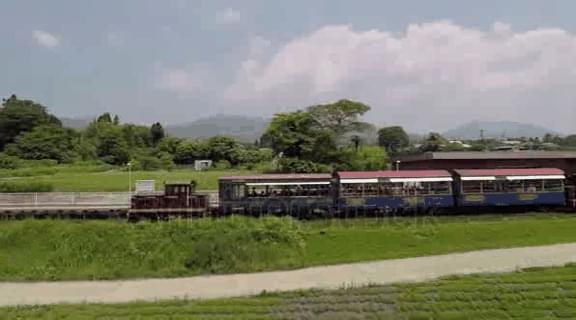} & \includegraphics[width=0.15\linewidth]{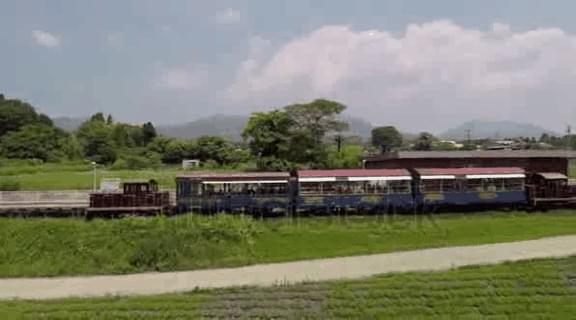} & \includegraphics[width=0.15\linewidth]{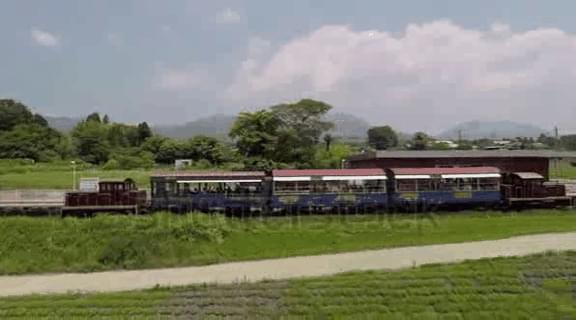} \\
\rotatebox[origin=l]{90}{\makebox[.08\linewidth]{w/ EMC}} & 
\includegraphics[width=0.15\linewidth]{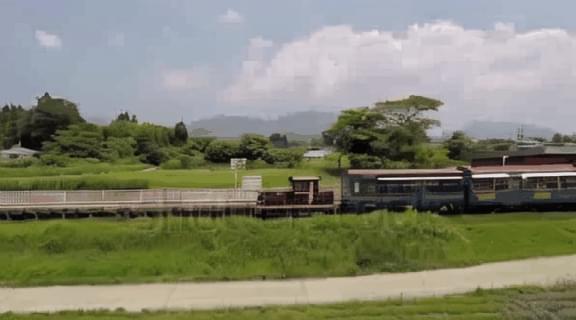} & \includegraphics[width=0.15\linewidth]{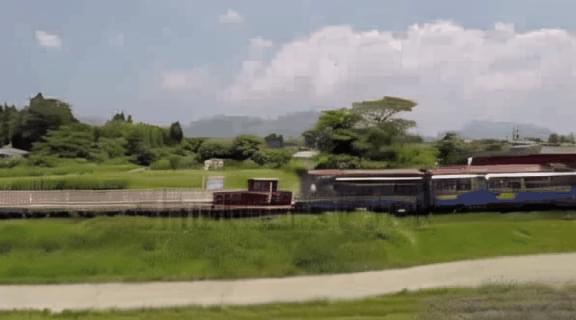} & \includegraphics[width=0.15\linewidth]{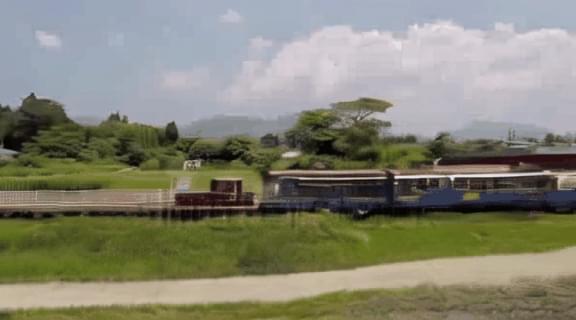} & \includegraphics[width=0.15\linewidth]{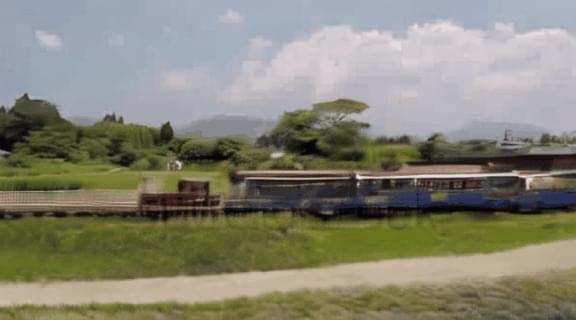} & \includegraphics[width=0.15\linewidth]{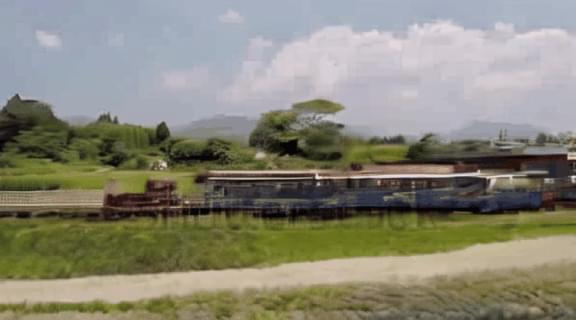} & \includegraphics[width=0.15\linewidth]{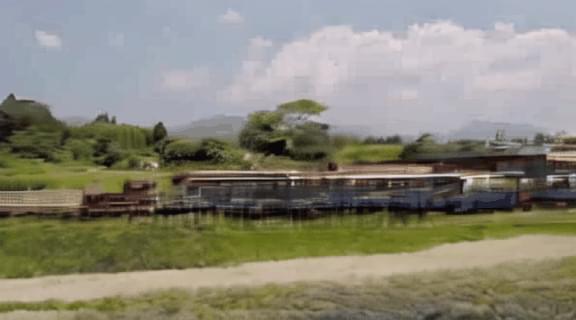} \\
\rotatebox[origin=l]{90}{\makebox[.08\linewidth]{w/o EMC}} & 
\includegraphics[width=0.15\linewidth]{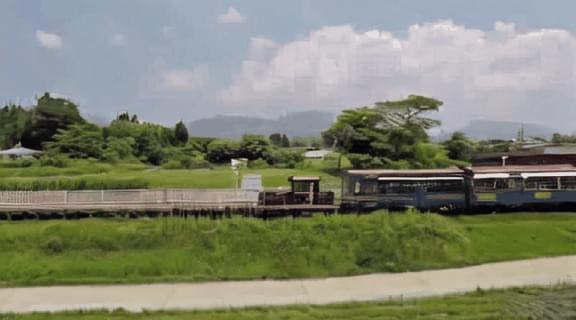} & \includegraphics[width=0.15\linewidth]{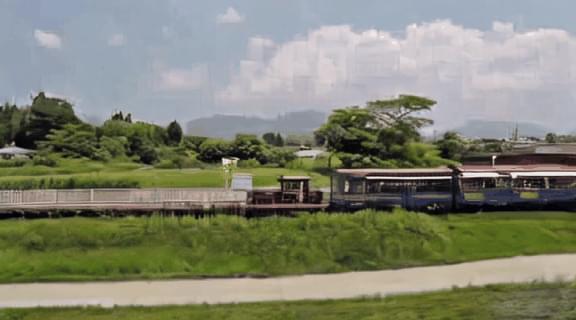} & \includegraphics[width=0.15\linewidth]{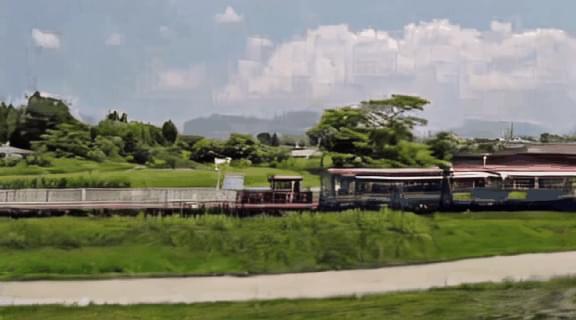} & \includegraphics[width=0.15\linewidth]{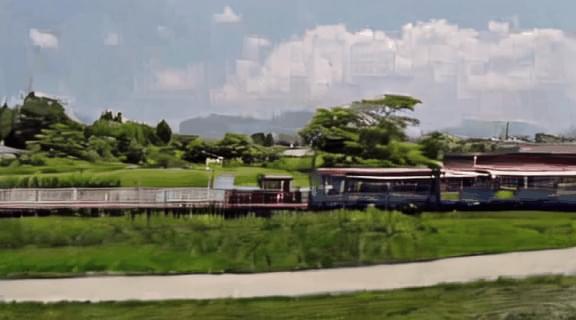} & \includegraphics[width=0.15\linewidth]{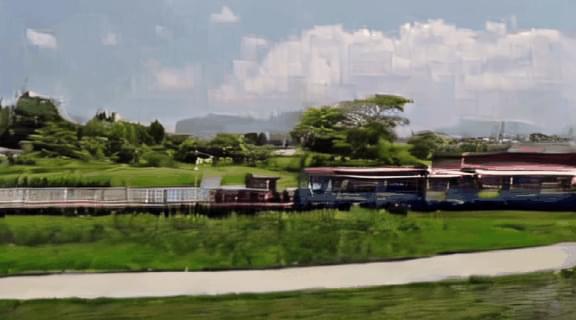} & \includegraphics[width=0.15\linewidth]{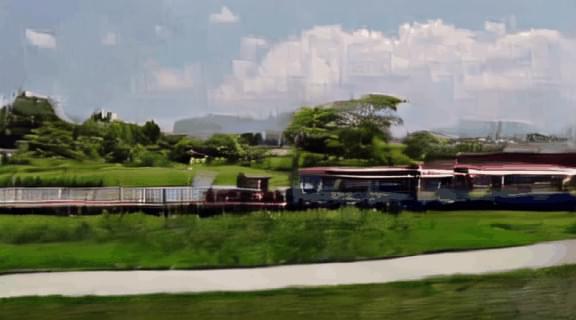} \\
\midrule
\rotatebox[origin=l]{90}{\makebox[.08\linewidth]{Original}} & 
\includegraphics[width=0.15\linewidth]{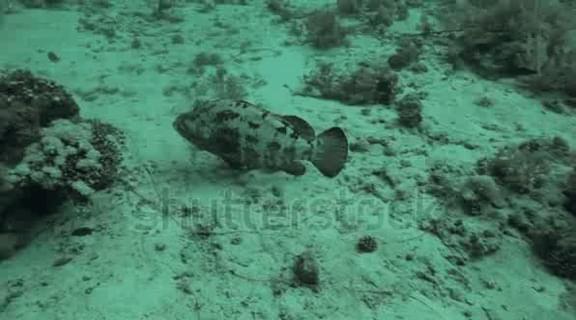} & \includegraphics[width=0.15\linewidth]{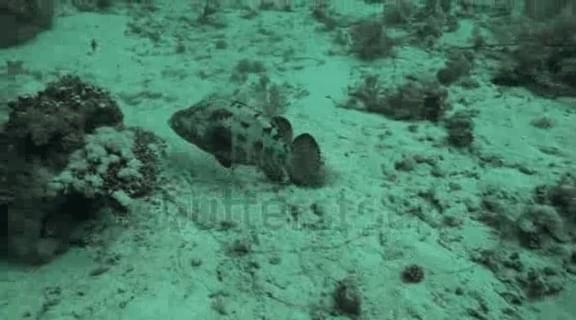} & \includegraphics[width=0.15\linewidth]{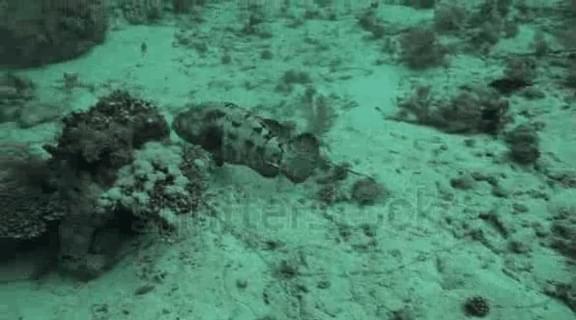} & \includegraphics[width=0.15\linewidth]{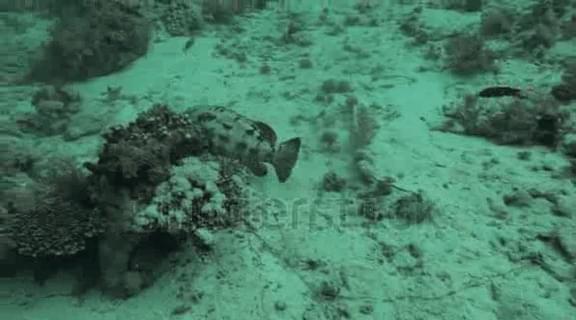} & \includegraphics[width=0.15\linewidth]{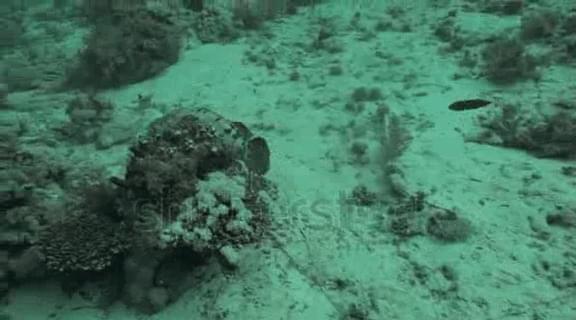} & \includegraphics[width=0.15\linewidth]{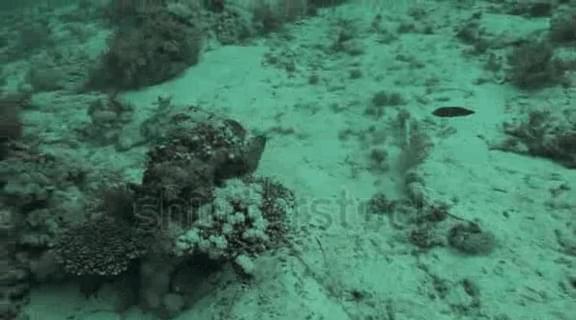} \\
\rotatebox[origin=l]{90}{\makebox[.08\linewidth]{w/ EMC}} & 
\includegraphics[width=0.15\linewidth]{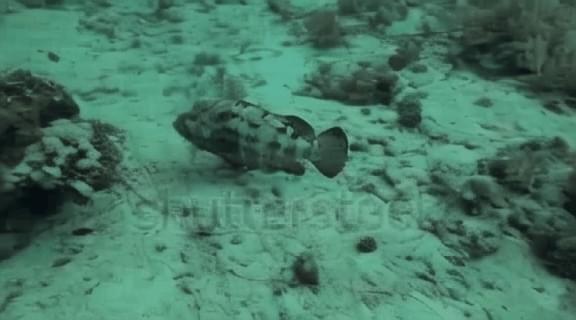} & \includegraphics[width=0.15\linewidth]{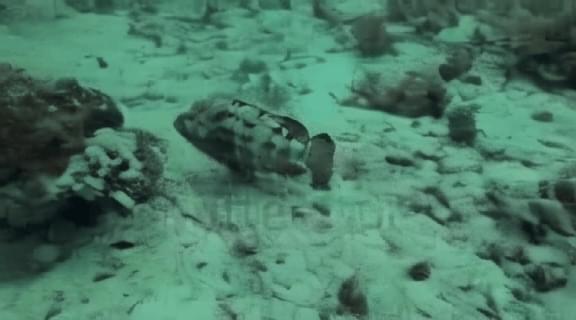} & \includegraphics[width=0.15\linewidth]{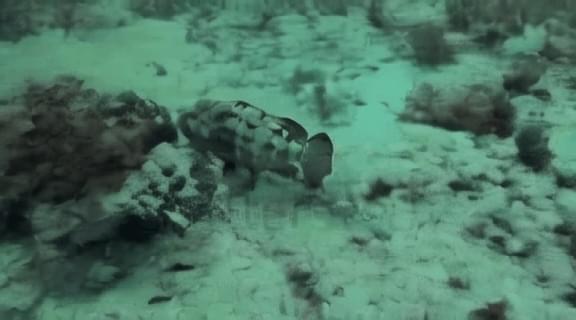} & \includegraphics[width=0.15\linewidth]{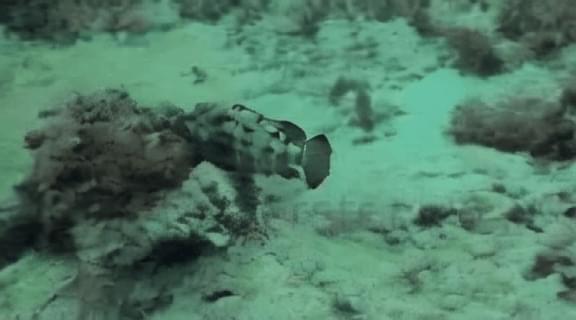} & \includegraphics[width=0.15\linewidth]{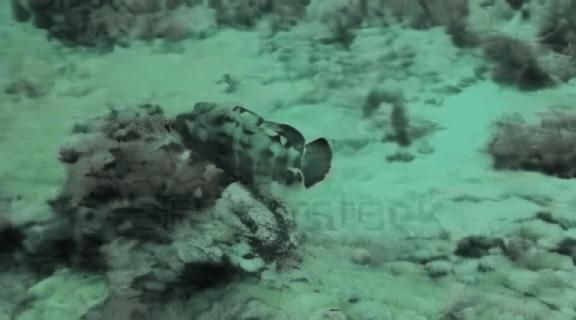} & \includegraphics[width=0.15\linewidth]{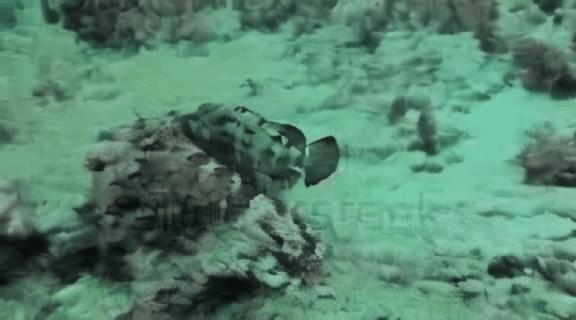} \\
\rotatebox[origin=l]{90}{\makebox[.08\linewidth]{w/o EMC}} & 
\includegraphics[width=0.15\linewidth]{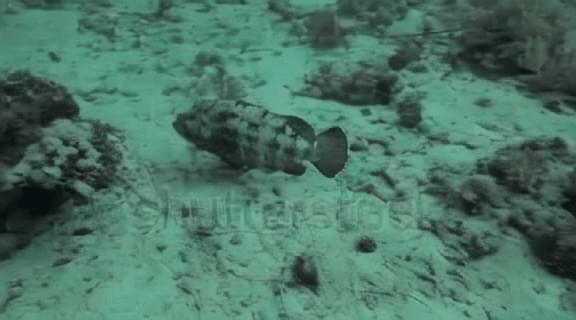} & \includegraphics[width=0.15\linewidth]{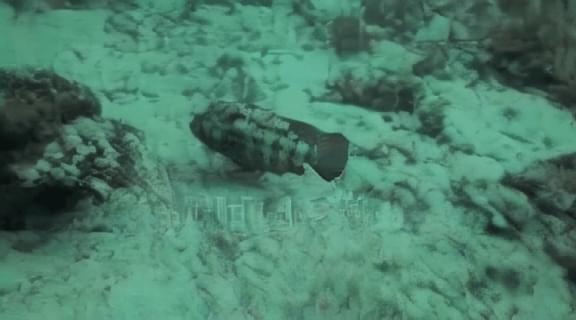} & \includegraphics[width=0.15\linewidth]{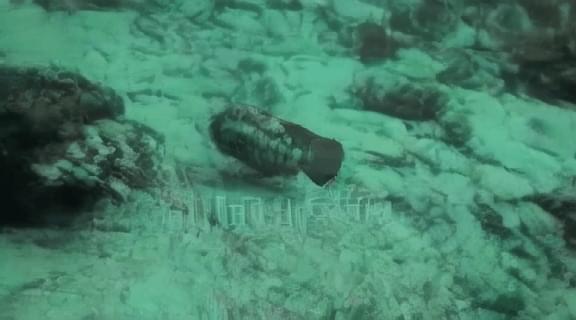} & \includegraphics[width=0.15\linewidth]{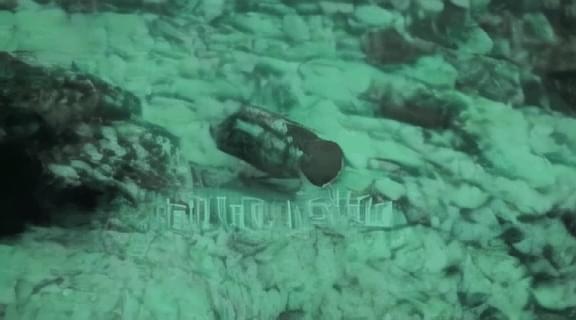} & \includegraphics[width=0.15\linewidth]{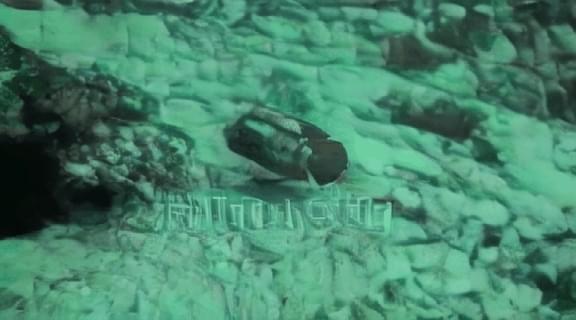} & \includegraphics[width=0.15\linewidth]{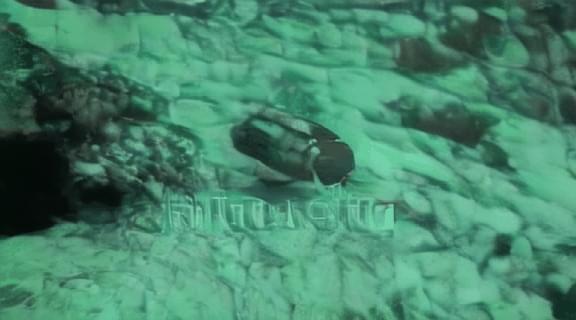} \\
\end{tabular}
\end{small}
\caption{Ablation study of enhanced motion conditioning (EMC) for video reconstruction. The last row (i.e., w/o EMC) indicates only motion vectors as the input condition for training 3D U-Net. As illustrated, incorporating EMC can improve the fidelity of the reconstructed videos. The video samples are taken from WebVid-10M~\cite{bain2021frozen}.}
\label{fig:recon}
\end{center}
\vskip -0.1in
\end{figure}

\begin{table*}[h]
\vskip -0.1in
\caption{The impact of incorporating motion tokens on image comprehension. }
\label{tab:supp_abla1}
\begin{center}
\begin{small}
\setlength{\tabcolsep}{6.5pt}
\resizebox{0.7\linewidth}{!}{
\begin{tabular}{lcccccccc}
\toprule
Method & VQAv2 & GQA & VizWiz & SQA & MME & MMB  & SEED & MM-Vet \\
\midrule
w/o motion & 80.0 & 63.7 & 54.4 & \textbf{71.5} & 1533.2 & \textbf{67.5} & \textbf{64.7} & \textbf{34.5}  \\
w motion & \textbf{80.3} & \textbf{64.4} & \textbf{56.0} & 70.0 & \textbf{1551.8} & 67.3  & 64.0 & 33.2 \\
\bottomrule
\end{tabular}
}
\end{small}
\end{center}
\vskip -0.1in
\end{table*}

\begin{table*}[h]
\vskip -0.1in
\caption{The impact of svd-img2vid-xt weight initialization on text-to-video generation.}
\label{tab:supp_abla2}
\begin{center}
\begin{small}
\resizebox{0.6\linewidth}{!}{
\begin{tabular}{lccccc}
\toprule
\multirow{2}{*}{Method} & \multicolumn{3}{c}{MSR-VTT} & \multicolumn{2}{c}{UCF-101} \\
\cmidrule(lr){2-4} \cmidrule(l){5-6} & CLIPSIM ($\uparrow$) & FVD ($\downarrow$) & FID ($\downarrow$) & IS ($\uparrow$) & FVD ($\downarrow$) \\
\midrule
w/ svd-img2vid-xt  & 0.3010 & \textbf{169.51} & 11.80 & 37.96 & \textbf{274.96} \\
w/o svd-img2vid-xt & \textbf{0.3012} & 188.36 & \textbf{11.27} & \textbf{44.26} & 280.57 \\
\bottomrule
\end{tabular}
}
\end{small}
\end{center}
\vskip -0.2in
\end{table*}

\textbf{The Effect of Enhanced Motion Conditioning}. 
To rigorously reconstruct original video content, we employ the enhanced conditioning: motion input condition and motion feature condition for training the 3D video U-Net $g_V$. We illustrate the effect of proposed enhanced motion conditioning (EMC) strategy on video decoding in~\cref{fig:recon}. The variant ``w/o EMC'' only leverages motion vectors as the input condition. Compared with using EMC, it is incapable of recovering the motion of original input videos. For example, the ``train'' and ``fish'' barely moved in the shown video samples, which demonstrates the effectiveness of our proposed conditioning strategy.


\subsection{Multimodal Understanding}

\begin{table}[t]
\vskip -0.1in
\caption{Image question answering comparison with LLaVA~\cite{liu2023visual}, GPT-4~\cite{openai2023gpt}, BLIP-2~\cite{li2023blip2}, and OpenFlamingo~\cite{awadalla2023openflamingo}. Our model answers correctly like the other multimodal LLMs, but without missing details such as the vehicle type (it is an SUV instead of a van or minivan). The table format follows GPT-4.}
\label{tab:iqa_ironing}
\vskip 0.1in
\begin{center}
\begin{small}
\begin{tabular}{l p{11.5cm}}
\toprule
\multicolumn{2}{l}{\bf Visual input example, Extreme Ironing:} \\
\midrule
&  \includegraphics[height=4.5cm]{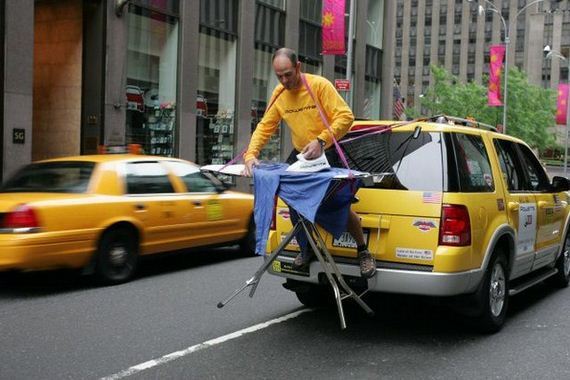} \\
& {\tiny Source:~\url{https://www.barnorama.com/wp-content/uploads/2016/12/03-Confusing-Pictures.jpg}} \\
User & What is unusual about this image? \\
\ours & The unusual aspect of this image is that a man is ironing clothes on an ironing board placed on the back of a yellow {\color{blue}SUV} while it is driving down the street.  This is not a typical scenario, as ironing clothes is usually done indoors, in a home or a laundry room, and not while the vehicle is in motion.  The man's actions pose a risk to himself and others on the road, as ironing on a moving vehicle can lead to accidents or distractions for the driver. \\
& [{\it Start a new conversation, and clear the history}] \\
User & What’s happening in the scene? \\
\ours & The image depicts a man ironing clothes on an ironing board placed on the back of a yellow {\color{blue}SUV}. The man is standing on the back of the vehicle, attentively ironing his clothes. The ironing board is positioned in the middle of the SUV, with the man standing on it to reach the clothes. In the background, there are two cars visible, one on the left side and another on the right side of the scene. The man appears to be focused on his task, despite the unusual location for ironing clothes. \\
\cmidrule{1-2}
User & What is unusual about this image? \\
LLaVA~\cite{liu2023visual} & The unusual aspect of this image is a man ironing clothes on the back of a {\color{red}minivan or van}. This is not a typical place to perform this activity, as one would usually iron clothes in a more stationary and safe location, such as a home, using a regular ironing board. The scene depicted in the image is peculiar as it involves a makeshift ironing setup on a vehicle, which can be both unsafe and unconventional. Additionally, it is not clear how the man is able to maintain balance and stability while ironing clothes in such an unstable environment. \\
& [{\it Start a new conversation, and clear the history}] \\
User & What’s happening in the scene? \\
LLaVA~\cite{liu2023visual} & The image depicts a man standing on top of a yellow SUV in a busy city street. He is holding a portable ladder, seemingly ironing clothes while standing on the vehicle. Around the scene, there are other cars, a traffic light, and a fire hydrant. Additionally, there are two other people in the vicinity, one close to the man on the car and another further away. The man on the ladder appears to be attracting attention due to his unconventional choice of ironing his clothes on top of a moving car. The city street around him is bustling with activity, adding to the unique nature of the scene. \\
\cmidrule{1-2}
User & What is unusual about this image? \\
GPT-4~\cite{openai2023gpt} &  The unusual thing about this image is that a man is ironing clothes on an ironing
board attached to the roof of a moving taxi. \\
\cmidrule{1-2}
User & Can you explain this meme in detail? \\
BLIP-2~\cite{li2023blip2} & a man is sitting on the back of a yellow cab \\
\cmidrule{1-2}
User & Can you explain this meme in detail? \\
OpenFlamingo~\cite{awadalla2023openflamingo} & The man is drying his clothes on the hood of his car. \\
\bottomrule
\end{tabular}
\end{small}
\end{center}
\vskip -0.1in
\end{table}

\begin{table}[t]
\vskip -0.1in
\caption{Video question answering comparison with Video-LLaVA~\cite{lin2023video} and Video-ChatGPT~\cite{maaz2023video}. Video examples are from Video-ChatGPT. In the first example, our model directly points to a salient explanation for why the video is funny. In~the second example, our model responses accurately without hallucinating about the video (the woman only swung the ax once).}
\label{tab:vqa_vis1}
\vskip 0.1in
\begin{center}
\begin{small}
\begin{tabular}{l p{11.5cm}}
\toprule
& \includegraphics[width=1.86cm]{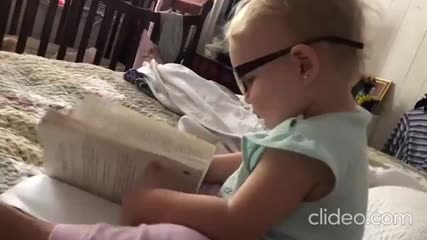} \includegraphics[width=1.86cm]{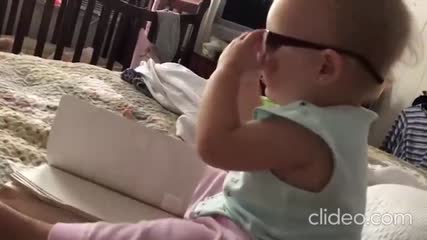} \includegraphics[width=1.86cm]{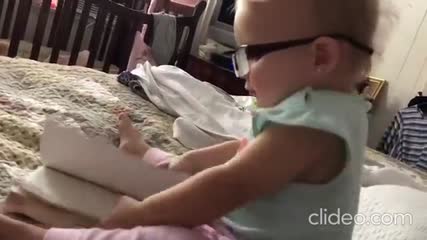} \includegraphics[width=1.86cm]{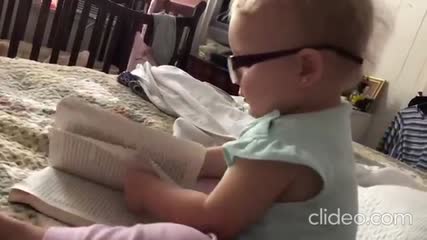} \includegraphics[width=1.86cm]{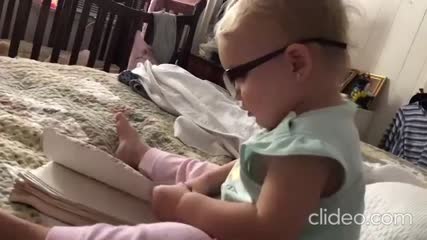} \includegraphics[width=1.86cm]{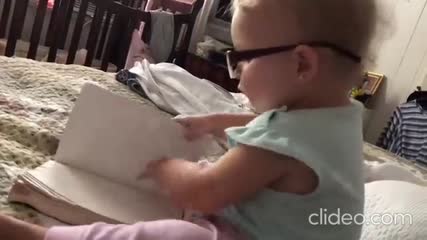} \\
User & Why is this video funny? \\
\ours & The video is funny because the baby is wearing glasses and trying to read a book, {\color{blue}{but she is not able to do so}}. The baby's attempts to read the book are adorable and amusing to watch. \\
\cmidrule{1-2}
User & Why is this video funny? \\
Video-LLaVA~\cite{lin2023video} & The video is funny because the baby is trying to read a book while wearing glasses, which is a humorous sight. The baby's attempts to read the book are also amusing, as they seem to be struggling with the task. Additionally, the baby's facial expressions and movements add to the overall humor of the video. \\
\cmidrule{1-2}
User & Why is this video funny? \\
Video-ChatGPT~\cite{maaz2023video} & The video is funny because it shows a baby girl who is learning to read. She is sitting on a bed and looking at a book, and she is trying to read it. The baby girl is wearing glasses, and she is holding the book with both hands. The baby girl is trying to read the book, and her actions are adorable and endearing. The video captures the innocence and curiosity of a young child as she learns to read, which is a joyful and amusing sight for the viewer. \\
\midrule
& \includegraphics[width=1.86cm]{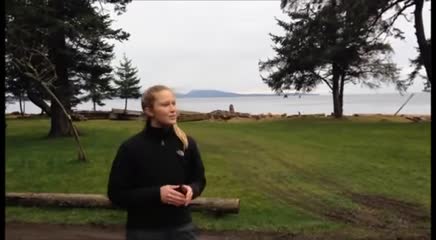} \includegraphics[width=1.86cm]{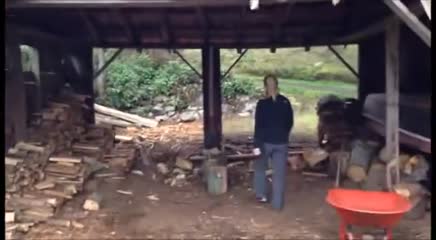} \includegraphics[width=1.86cm]{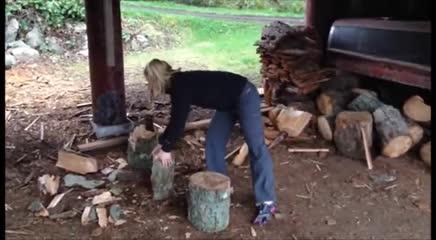} \includegraphics[width=1.86cm]{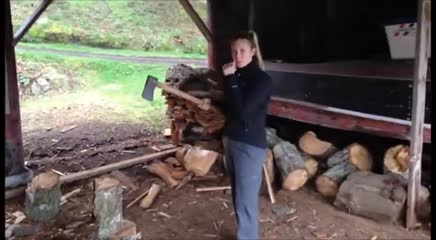} \includegraphics[width=1.86cm]{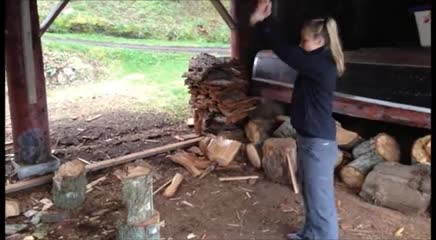} \includegraphics[width=1.86cm]{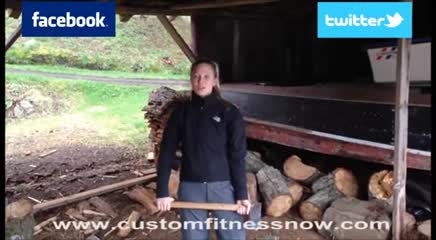} \\
User & What does the woman use to split the logs and how does she do it? \\
\ours &  The woman uses an ax to split the logs. She picks up the ax and swings it down with force, splitting the logs in half.  \\
\cmidrule{1-2}
User & What does the woman use to split the logs and how does she do it? \\
Video-LLaVA~\cite{lin2023video} & The woman uses an ax to split the logs. She holds the ax in her hand and swings it down onto the log. {\color{red}She repeats this process several times}, splitting the log into smaller pieces. \\
\cmidrule{1-2}
User & What does the woman use to split the logs and how does she do it? \\
Video-ChatGPT~\cite{maaz2023video} & The woman uses a large axe to split the logs. She starts by placing the axes on the log and then swings it down with a powerful motion. The axe slices through the log with ease, and {\color{red}the woman continues to swing it} until the log is split in half. \\
\bottomrule
\end{tabular}
\end{small}
\end{center}
\vskip -0.1in
\end{table}

\begin{table}[t]
\vskip -0.1in
\caption{Video question answering comparison with Video-LLaVA~\cite{lin2023video} and Video-ChatGPT~\cite{maaz2023video}. Video examples are from Video-ChatGPT. As shown in this example, our model captures the underlying context of this video, which is a love story. The fairy tale told, which implies that love will prevail, is naturally inspired by the video without introducing an unmentioned third party (jealous sister or stepmother) as in the other models. The use of the phrase ``beautiful glass slipper'' adds a nice detail to the story. }
\label{tab:vqa_vis2}
\vskip 0.1in
\begin{center}
\begin{small}
\begin{tabular}{l p{11.5cm}}
\toprule
& \includegraphics[width=1.86cm]{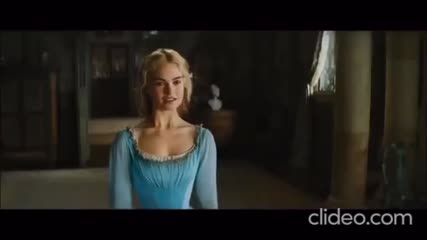} \includegraphics[width=1.86cm]{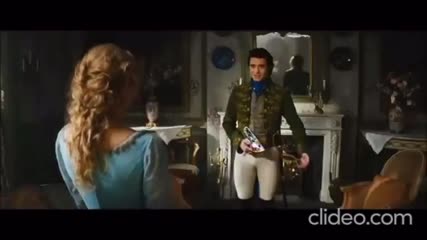} \includegraphics[width=1.86cm]{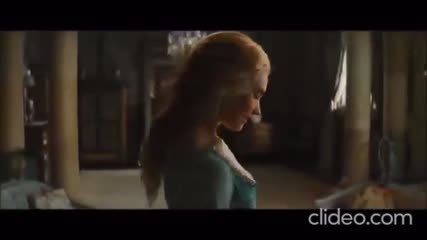} \includegraphics[width=1.86cm]{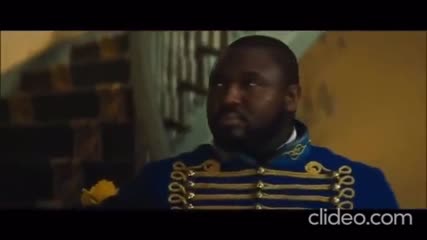} \includegraphics[width=1.86cm]{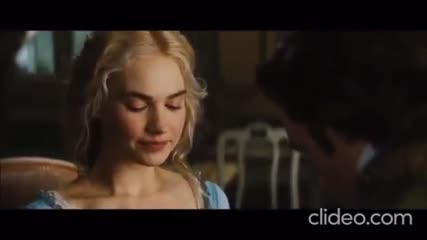} \includegraphics[width=1.86cm]{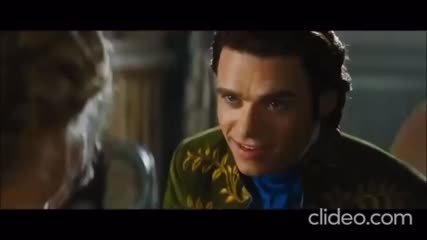} \\
User & Create a short fairy tale with a moral lesson inspired by the video. \\
\ours & Once upon a time, there was a beautiful princess who lived in a grand castle.  She was known for her kindness and compassion towards others.  One day, a handsome prince visited her castle and fell deeply in love with her.  However, the princess was hesitant to accept his love due to her fear of losing her independence and being trapped in a loveless marriage. The prince, determined to win her heart, {\color{blue}gave her a beautiful glass slipper as a gift}. The princess tried it on and danced with the prince, realizing that true love transcends all boundaries.  From that day on, the princess and the prince lived happily ever after, proving that love can conquer all. \\
\cmidrule{1-2}
User & Create a short fairy tale with a moral lesson inspired by the video. \\
Video-LLaVA~\cite{lin2023video} & Once upon a time, in a kingdom far away, there lived a beautiful princess named Rose. She was known for her kindness and beauty, and everyone loved her. One day, a handsome prince named Leo came to visit the kingdom. He was fascinated by Rose's beauty and decided to stay in the kingdom. However, Rose's jealous sister, Lily, did not like the attention Leo was giving her sister. She tried to harm Rose, but Leo saved her and they fell in love. The kingdom celebrated their love and they lived happily ever after. The moral of the story is that true love conquers all and that jealousy can lead to harmful consequences. \\
\cmidrule{1-2}
User & Create a short fairy tale with a moral lesson inspired by the video. \\
Video-ChatGPT~\cite{maaz2023video} & Sure, here's a short fairy tale with a moral lesson inspired by the video: Once upon a time, there was a beautiful princess named Cinderella who lived in a castle with her cruel stepmother and two stepsisters. She was known for her kindness and her love for animals, especially birds. One day, while she was out in the garden, she met a handsome prince who was lost in the woods. He was in need of help, and Cinderella offered to guide him back to the castle. As they walked together, the prince noticed that Cinderella's stepsisters were cruel to her and her and stepmother was always mean to her. He asked her about her life and how she managed to stay so kind and gentle despite the hardships she faced. Cinderella told him that she found happiness in the simple things in life, such as the beauty of nature and the love of animals. The prince was touched by her story and decided to help her in any way he could. He introduced her to the king and queen, who were impressed by her kindness and grace. They arranged for her to attend a ball at the castle, where she met the prince once again. They fell in love and got married, living happily ever after. The moral of the story is that even in difficult situations, one should find happiness in the simple things in life and never lose sight of their values and kindness. \\
\bottomrule
\end{tabular}
\end{small}
\end{center}
\vskip -0.1in
\end{table}

This section presents qualitative results of \ours~for image and video understanding. First, \cref{tab:iqa_ironing} showcases our performance in image question answering using the famous test example from GPT-4~\cite{openai2023gpt}. As can be seen, our model produces a reasonable answer with a good number of correct details (e.g. the type of the vehicle being SUV) and even a friendly safety warning. In comparison, one of the recent multimodal LLMs, LLaVA~\cite{liu2023visual}, produces a roughly correct answer with some inaccurate detail (mistaking the vehicle type as ``minivan or van'').

For video question answering, \cref{tab:vqa_vis1,tab:vqa_vis2} compares our method to Video-LLaVA~\cite{lin2023video} and Video-ChatGPT~\cite{maaz2023video} based on the video clips from Video-ChatGPT. In the first example of~\cref{tab:vqa_vis1} which asks to explain why a video is funny, our model yields the most concise answer among the video-language models compared, and at the same time contains a salient point that the other models failed to mention. The next example in~\cref{tab:vqa_vis1}, on the other hand, shows that our method produces fewer hallucinations than Video-LLaVA and Video-ChatGPT, as the latter two models tend to generate overly detailed action descriptions that have no basis in the video. And lastly, in the example of~\cref{tab:vqa_vis2}, our model follows the instruction prompt by producing a beautiful fairy tale with both conciseness and a moral lesson (``love can conquer all''). To summarize, our method demonstrates reasonably good multimodal understanding capabilities across different test cases, in line with the previous quantitative comparison on multiple benchmarks.

\subsection{Ablation Studies}

\textbf{Impact of Motion Tokens on Image Comprehension}. \ours~indiscriminately treat all the modalities (video, image, and text) as 1D discrete tokens fed
into LLMs. The impact of incorporating motion tokens on image comprehension is reported in \Cref{tab:supp_abla1}. As observed, including motion tokens hardly affects the understanding performance of the image, which demonstrated the effectiveness of the proposed decoupled visual-motional tokenization. \ours~is capable of modeling video, image, and text data in a unified framework.

\textbf{Impact of Weight Initialization}. We re-trained the detokenizer of \ours~from scratch without svd-img2vid-xt initialization on the WebVid-10M dataset and found that our model can still achieve comparable video generation performance. The detailed text-to-video generation results are reported in \Cref{tab:supp_abla2}. As observed, training the detokenizer from scratch had little impact on the final results.

\section{Limitations}

Our proposed model cannot generate very long videos due to its limited context window (4096) and dataset restriction. This work only used the public WebVid-10M as the video pre-training data. In WebVid, the video durations are relatively short (about 15s on average) and the video scenes barely change, which results in our model generating similar keyframes in different clips. On the other hand, a general concern is that our training cost is still too high to scale to web-scale video data, which may require further optimization through joint exploitation of spatial and temporal redundancies in video.



\end{document}